\documentclass{article} % For LaTeX2e
\usepackage[final]{colm2026_conference}

\usepackage{microtype}
\usepackage{hyperref}
\usepackage{url}
\usepackage{booktabs}
\usepackage{graphicx}
\usepackage{multirow, multicol}
\usepackage{tcolorbox}
\tcbset{
  colback=white,
  colframe=black,
  boxrule=0.5pt,
  arc=2pt,
  left=6pt,
  right=6pt,
  top=6pt,
  bottom=6pt
}
\usepackage{subcaption}
\usepackage{wrapfig}

\usepackage{lineno}
\usepackage{amsmath}

\definecolor{darkblue}{rgb}{0, 0, 0.5}
\hypersetup{colorlinks=true, citecolor=darkblue, linkcolor=darkblue, urlcolor=darkblue}

\title{Early Stopping for Large Reasoning Models \\ via Confidence Dynamics}

\author{%
  Parsa Hosseini$^{1}$\thanks{Equal Contribution. Code is available at \url{https://github.com/sudoparsa/CoDE-Stop}} \quad 
  Sumit Nawathe$^{1*}$\quad 
  Mahdi Salmani$^{2*}$\quad \\
\textbf{Meisam Razaviyayn$^{2}$\quad
  Soheil Feizi$^{1}$} \\
  $^{1}$University of Maryland \quad $^{2}$University of Southern California
}

\begin{document}

\ifcolmsubmission
\linenumbers
\fi

\maketitle

\begin{abstract}
    
Large reasoning models rely on long chain-of-thought generation to solve complex problems, but extended reasoning often incurs substantial computational cost and can even degrade performance due to overthinking. A key challenge is determining when the model should stop reasoning and produce the final answer. In this work, we study the confidence of intermediate answers during reasoning and observe two characteristic behaviors: correct reasoning trajectories often reach high-confidence answers early, while incorrect rollouts tend to produce long, unproductive reasoning traces and exhibit less reliable confidence dynamics. Motivated by these observations, we propose CoDE-Stop (Confidence Dynamics Early Stop), an early stopping method that leverages the dynamics of intermediate answer confidence to decide when to terminate reasoning, requiring no additional training and easily integrating into existing models. We evaluate CoDE-Stop on diverse reasoning and science benchmarks across multiple models. Compared to prior early stopping methods, it achieves a more favorable accuracy–compute tradeoff and reduces total token usage by 25-50\% compared to standard full-length reasoning. In addition, we provide analyses of confidence dynamics during reasoning, offering insights into how confidence changes in both correct and incorrect trajectories.

\end{abstract}

\section{Introduction}

Generating long chain-of-thought (CoT) enables large language models (LLMs) to solve complex problems, but at a substantial computational cost. As reasoning models are increasingly deployed with long generation budgets (often spanning thousands of tokens) efficient inference becomes critical. In particular, reducing unnecessary reasoning steps is essential for improving the practicality and scalability of these systems. A key challenge is determining when to terminate reasoning and produce the final answer.

Several lines of work have sought to improve the efficiency of reasoning in LLMs. Some approaches rely on training-time modifications, such as compressing CoT~\citep{arora2025traininglanguagemodelsreason} or training auxiliary models~\citep{zhang2025reasoningmodelsknowtheyre} for early stopping, but these require additional training and are not directly applicable at inference time. More recent work focuses on inference-time strategies that terminate reasoning by monitoring signals during generation, typically via intermediate answer predictions~\citep{deer, answerconv, RCPD, thinkornot}. These methods are effective at stopping trajectories that have already reached a reliable answer, but are less effective when reasoning becomes unproductive (such as when the model enters loops or exhibits unstable progress) often resulting in unnecessarily long generations.

To better understand when reasoning can be terminated, we analyze how model confidence evolves over the course of generation. We observe that reasoning trajectories exhibit distinct behaviors depending on their outcome. For trajectories that lead to correct answers, confidence typically increases over time and reaches a high value well before generation ends. In contrast, incorrect trajectories often exhibit unstable behavior, with confidence fluctuating. Motivated by these observations, we propose CoDE-Stop (Confidence Dynamics Early Stopping), an inference-time method that leverages the temporal behavior of confidence to decide when to terminate reasoning. CoDE-Stop monitors confidence over reasoning steps and combines two complementary signals: a threshold on confidence to identify when the model has reached a reliable answer, and a degeneration signal to detect unproductive reasoning characterized by unstable or non-improving confidence.

\begin{figure}
    \centering
    \includegraphics[width=\linewidth]{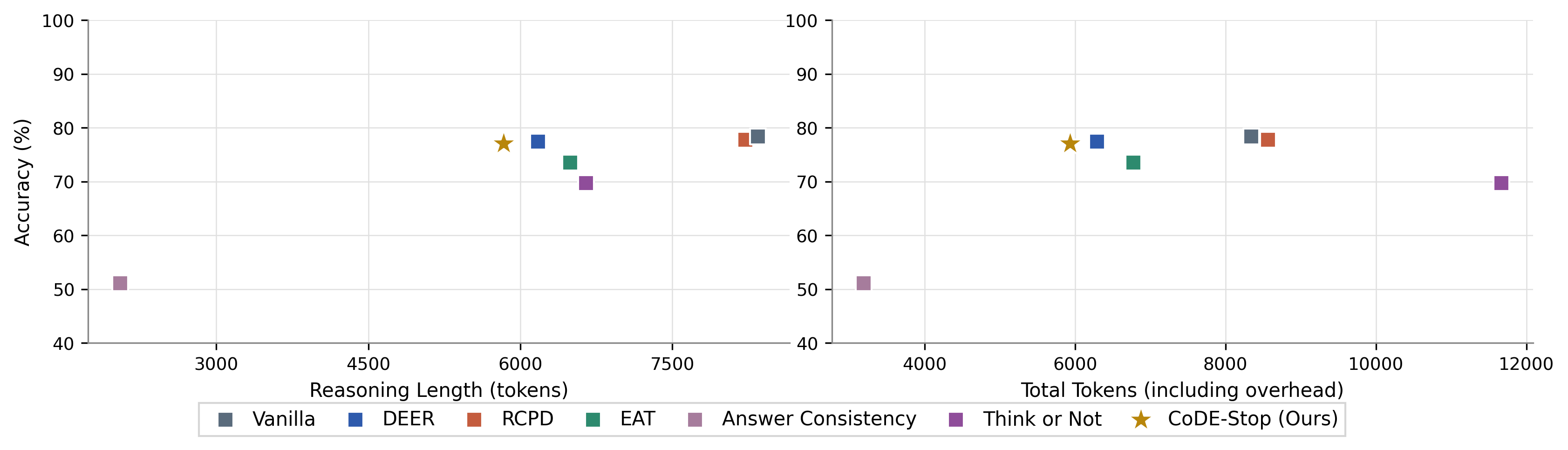}
    \caption{Accuracy vs. compute cost on Qwen3-4B averaged over 4 reasoning and science benchmarks. \textbf{Left:} reasoning length only. \textbf{Right:} total token compute including intermediate answer-generation overhead. CoDE-Stop achieves a stronger accuracy–compute tradeoff among early stopping methods.}
    \label{fig:fig1}
\end{figure}

We evaluate CoDE-Stop on various models across reasoning and science datasets, comparing against prior inference-time early stopping methods that rely on intermediate answer generation. Figure~\ref{fig:fig1} shows the tradeoff between accuracy and token usage. Whether measuring reasoning length alone or total token compute including intermediate answer overhead, CoDE-Stop achieves a more favorable accuracy–compute tradeoff, reducing token usage by 25\% while maintaining accuracy comparable to strong baselines.

Beyond overall performance, our analysis reveals a key property of confidence dynamics: the early stages of reasoning provide more informative signals for distinguishing correct and incorrect trajectories. At the same time, models tend to become increasingly confident even along incorrect trajectories, making late-stage confidence unreliable for detecting failure. These observations suggest that confidence should not be treated uniformly over time. They directly inform our design: CoDE-Stop combines a degeneration signal with non-uniform weighting over reasoning steps. Together, our results demonstrate that CoDE-Stop is a simple, training-free, and effective approach for improving the efficiency of reasoning in LLMs.

\section{Confidence Dynamics in Reasoning Models}
\label{sec:prelim}

\subsection{Problem Setup and Motivation}
%To study how reasoning evolves over time, we associate each step with a confidence score following prior work~\citep{deer,EAT}. Specifically, at certain points in the generation (e.g., when tokens such as ``\texttt{Wait}'' or ``\texttt{\textbackslash n\textbackslash n}'' appear), we prompt the model to produce an answer and measure the probability assigned to the answer tokens. We defer full details of this procedure to  Section~\ref{sec:method}. In this section, we analyze the temporal dynamics of confidence during reasoning and show that correct and incorrect trajectories exhibit distinct patterns that can be leveraged for early stopping. All plots in this section use the Qwen3-4B~\citep{qwen3technicalreport} model on the AIME~\citep{aime_aops} dataset.

We consider a reasoning trajectory generated by an LLM as a sequence of tokens \( x_1, x_2, \dots, x_T \). During generation, we identify a set of reasoning steps indexed by \( i = 1, \dots, K \), where \( T_i \) denotes the token position of the \( i \)-th step. In practice, these steps correspond to specific transition points in the generation (e.g., tokens such as ``\texttt{Wait}'' or ``\texttt{\textbackslash n\textbackslash n}''), following prior work~\citep{deer,EAT}.

We associate each reasoning step with a confidence score. At each step \( i \), we prompt ((\texttt{\textbackslash n**Final Answer**\textbackslash n\textbackslash n The final answer is \textbackslash boxed})) the model to produce an answer and compute a confidence score \( c_i \) as the average probability assigned to the generated answer tokens. This results in a sequence of confidence values \( \{c_i\}_{i=1}^K \), capturing the evolution of the model’s belief throughout the reasoning process. %We defer full details of this procedure to Section~\ref{sec:method}.

Our goal is to determine a stopping time \( t \) based on \( \{c_i\} \), such that reasoning can terminate early without sacrificing answer correctness. In this section, we analyze the temporal dynamics of confidence and show that correct and incorrect trajectories exhibit distinct patterns that can be leveraged for early stopping.

\begin{figure}
    \centering
    \includegraphics[width=\linewidth]{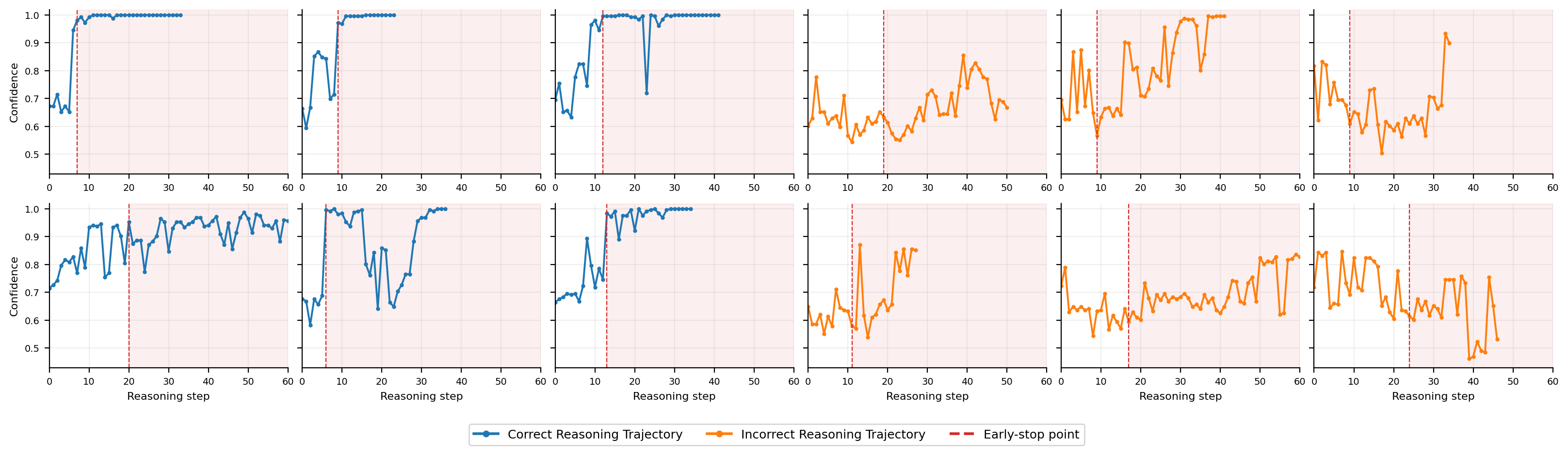}
    \caption{\textbf{Confidence dynamics across reasoning trajectories.} Correct trajectories reach high confidence early, while incorrect trajectories exhibit unstable and fluctuating confidence.}
    \label{fig:62}
\end{figure}

We first examine trajectories that lead to correct answers. Figure~\ref{fig:62} shows representative examples when prompting Qwen3-4B~\citep{qwen3technicalreport} on AIME24,25. As generation progresses, the confidence of correct trajectories increases rapidly and reaches a high value well before the end of the sequence. Despite this, the model often continues generating tokens after high confidence is already achieved, indicating that many later reasoning steps are unnecessary. This suggests that confidence provides a reliable signal for early stopping on successful trajectories, consistent with prior work such as \citet{deer}.

\begin{wrapfigure}{r}{0.48\linewidth}
    \centering
    \includegraphics[width=\linewidth]{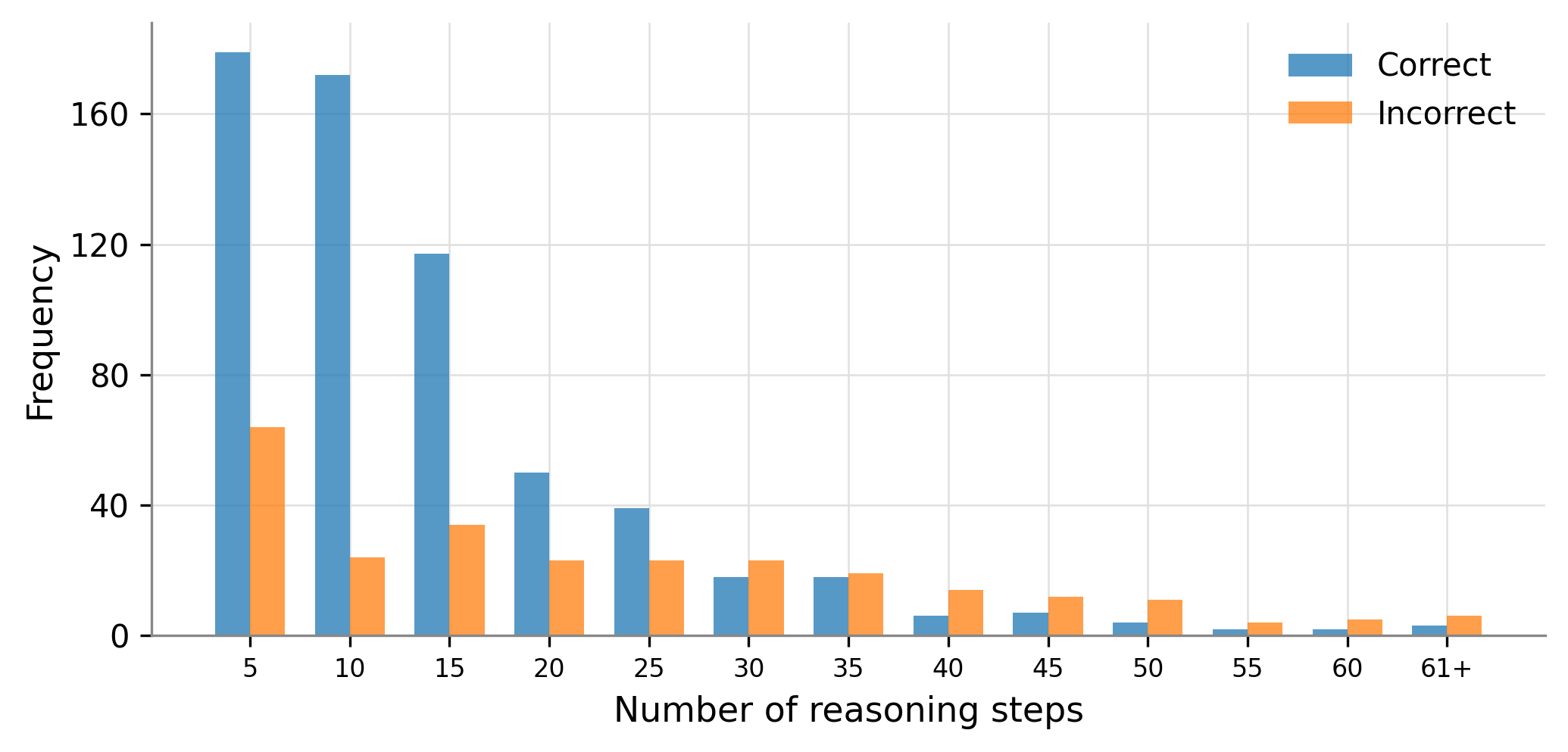}
    \caption{Incorrect trajectories are longer and exhibit a heavy-tailed distribution.}
    \label{fig:steps_freq}
\end{wrapfigure}

We next examine trajectories that lead to incorrect answers. We find that correct trajectories require on average 12K tokens, whereas incorrect trajectories extend to more than 25K tokens, doubling the total computation. Figure~\ref{fig:steps_freq} shows that a similar pattern holds in terms of reasoning steps: incorrect trajectories are substantially longer and exhibit a heavy-tailed distribution, while correct trajectories typically terminate much earlier. This indicates that a significant portion of computation is dominated by long, unproductive reasoning trajectories that do not lead to correct answers.

A key difficulty in stopping such trajectories is that confidence at a single step is not sufficient to determine whether reasoning should terminate. While high confidence indicates that the model has likely reached a final answer, low confidence is ambiguous: it may reflect exploration, after which the model can recover and arrive at a correct solution. However, incorrect trajectories often exhibit persistent instability, with confidence fluctuating over time without consistent improvement (Figure~\ref{fig:62}). This suggests that identifying unproductive reasoning requires reasoning over the full sequence of confidence values $\{c_i\}$, rather than relying on individual steps. In particular, these signals become more reliable when accumulated over multiple steps.

\begin{figure}[t]
    \centering
    \begin{minipage}[t]{0.48\linewidth}
        \centering
        \includegraphics[width=\linewidth]{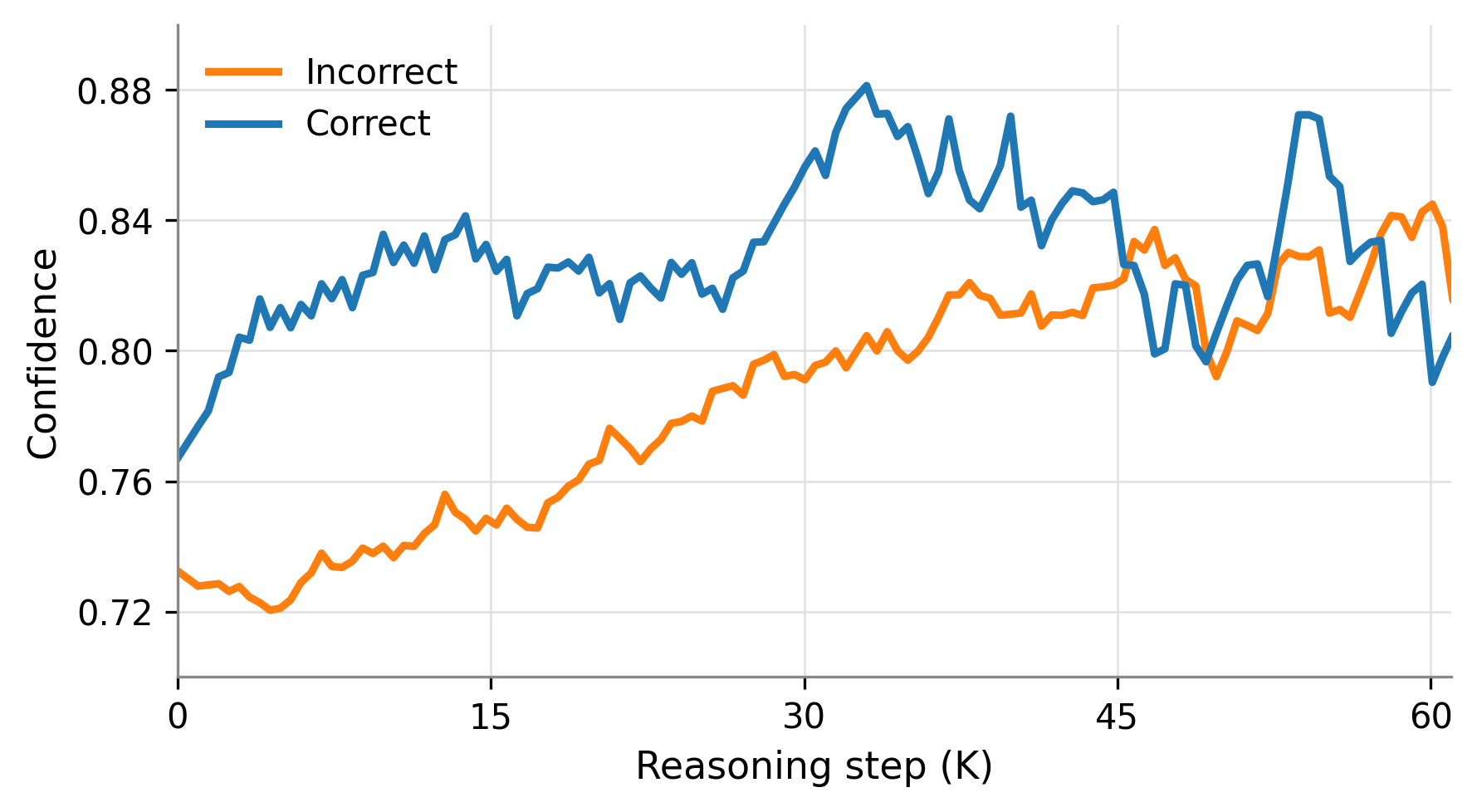}
    \end{minipage}
    \hfill
    \begin{minipage}[t]{0.48\linewidth}
        \centering
        \includegraphics[width=\linewidth]{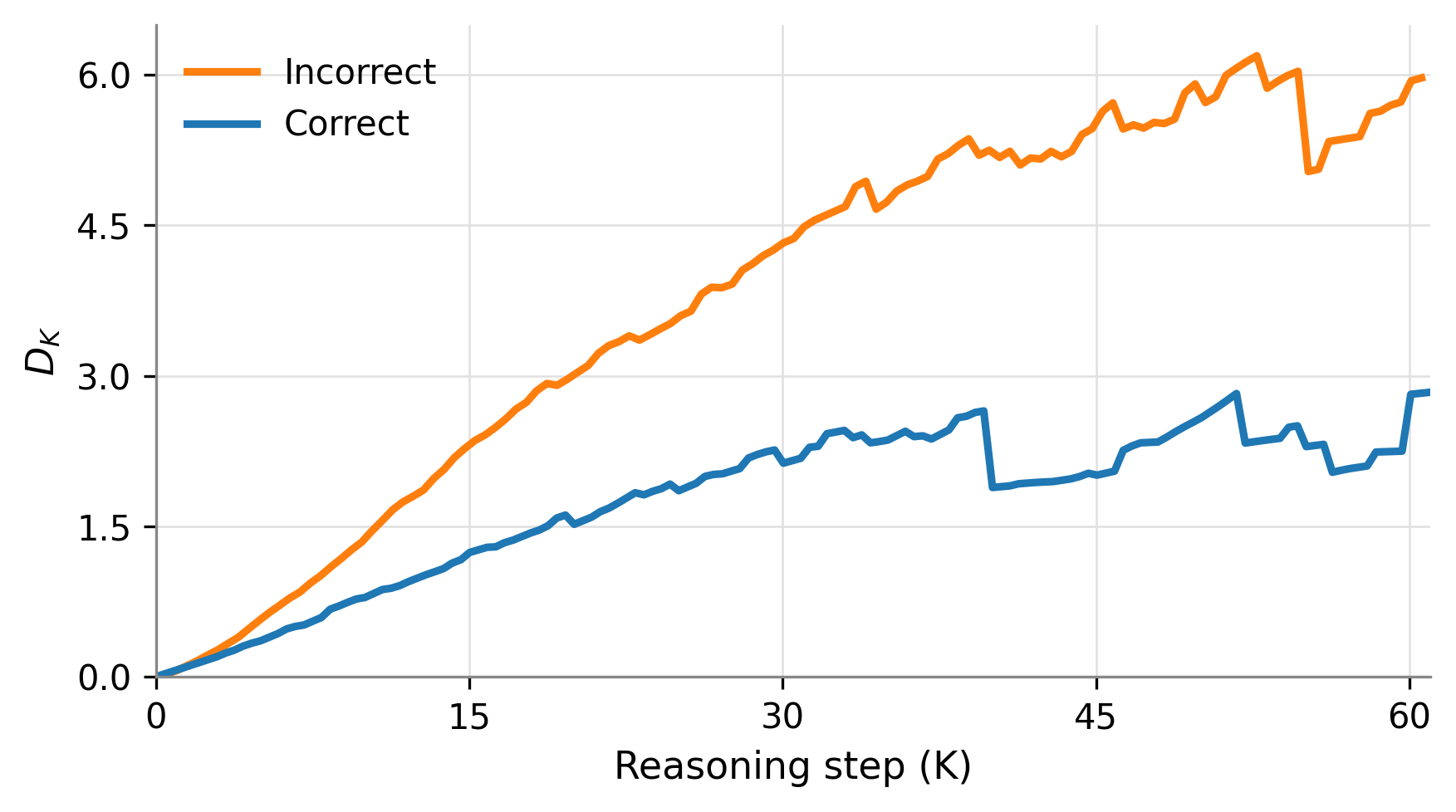}
    \end{minipage}
    \vspace{-3mm}
    \caption{Confidence and degeneration dynamics over reasoning steps. \textbf{Left}: Average confidence for correct and incorrect rollouts; confidence increases even for incorrect cases, while early steps provide better separation. \textbf{Right}: Average degeneration score $D_K$; incorrect rollouts consistently exhibit higher values, with the gap increasing over time.}
    \label{fig:confidence_step}
\end{figure}

To further understand which parts of the trajectory are most informative, we analyze confidence across reasoning steps. Figure~\ref{fig:confidence_step} (left) shows that early-stage confidence provides clearer separation between correct and incorrect trajectories, whereas later steps exhibit increased overlap. Notably, confidence tends to increase over time even for incorrect trajectories, making late-stage confidence unreliable for detecting failure. We provide the corresponding analyses for additional model--benchmark combinations in Appendix~\ref{app:extended_motivation}.

Together, these observations suggest that effective early stopping should account for the temporal dynamics of confidence. In particular, identifying unproductive reasoning requires accumulating signs of instability over time, while placing greater emphasis on earlier reasoning steps, which provide more reliable signals. To capture these properties, we introduce a degeneration score at step $k$ that accumulates instability across reasoning steps:
\begin{equation}
D_k = \sum_{i=1}^{k} w_i \, v_i,
\label{eq:deg_score}
\end{equation}
where \(v_i\) is an indicator of instability at step \(i\), and \(w_i\) assigns higher importance to earlier steps in the trajectory. Intuitively, \(v_i\) captures whether confidence exhibits unstable behavior at a given step, while \(w_i\) emphasizes early-stage signals, which are more informative for distinguishing correct and incorrect trajectories.

Figure~\ref{fig:confidence_step} (right) shows that this accumulated signal separates correct and incorrect trajectories early in the reasoning process, with incorrect trajectories consistently exhibiting higher degeneration scores. This suggests that the degeneration score provides a reliable criterion for identifying unproductive reasoning. In the next section, we provide a detailed formulation of \(v_i\) and \(w_i\), and describe how the degeneration score can be used to enable effective early stopping.

\subsection{Method}
\label{sec:method}

We propose \textbf{CoDE-Stop}, an early stopping method that leverages the temporal dynamics of confidence during reasoning. Our approach combines two complementary signals: (1) a confidence-based criterion to identify when the model has reached a reliable answer, and (2) a degeneration score that captures persistent instability in reasoning trajectories. Together, these signals enable early termination of both successful and unproductive reasoning.

At each reasoning step \( k \), we determine whether to stop generation based on these two signals. We define a ramping confidence threshold
\begin{equation}
r_k = \min\left(r_{\max}, \, r_{\min} + \frac{r_{\max} - r_{\min}}{\textit{steps}} \cdot k \right),
\end{equation}
where \( r_{\min} \), \( r_{\max} \), and \textit{steps} are constants. The parameter \textit{steps} controls how quickly the threshold increases, reaching \( r_{\max} \) after the specified number of reasoning steps and remaining constant thereafter. This reflects that higher confidence is required at later stages of reasoning.

% We combine confidence and degeneration into a unified early stopping rule. At each reasoning step \( k \), we stop generation when there is sufficient evidence that further reasoning is unnecessary. We define a ramp threshold
% \begin{equation}
% r_k = \min\left(r_{\max}, \, r_{\min} + \alpha k \right),
% \end{equation}
% where \( r_{\min} \), \( \alpha \), and \( r_{\max} \) are constants. This threshold increases with the number of reasoning steps while remaining capped.

We build on the degeneration score defined in Eq.~\ref{eq:deg_score} and specify the forms of the instability indicator \(v_i\) and weighting function \(w_i\). We implement the instability indicator as
\begin{equation}
v_i = \mathbb{1}\left( 2c_i - c_{i-1} < \delta \right),
\end{equation}
where \( \delta \) is a fixed threshold (set to 0.55 in all experiments). This captures whether confidence is low and fails to improve relative to previous step, indicating a lack of meaningful progress. The weighting function is defined as
\begin{equation}
w_i = \log\left( \frac{T_k}{T_i} \right) + 1,
\end{equation}
which assigns higher importance to earlier reasoning steps, consistent with our observation that early-stage signals are more informative. The logarithmic form provides a smooth growth over time while ensuring that the degeneration score increases monotonically with \(k\). Further analysis of these design choices is provided in Section~\ref{sec:ablation}.

We stop generation at step \( k \) if either \( c_k \ge r_k \) or \( D_k \ge \tau \), where \( D_k \) is the degeneration score accumulated up to step \( k \), and \( \tau \) is a fixed threshold. The first condition captures trajectories that have reached sufficiently high confidence, while the second condition identifies trajectories with persistent instability. Once a stopping condition is met, we prompt the model using the same answer-generation prompt to produce the final answer. Both \( r_k \) and \( \tau \) control the accuracy--compute tradeoff: lower values lead to earlier termination with reduced computation, while higher values allow longer reasoning and recover accuracy closer to the base model.

\section{Experiments}
\label{sec:experiments}
\begin{figure}
    \centering
    \includegraphics[width=\linewidth]{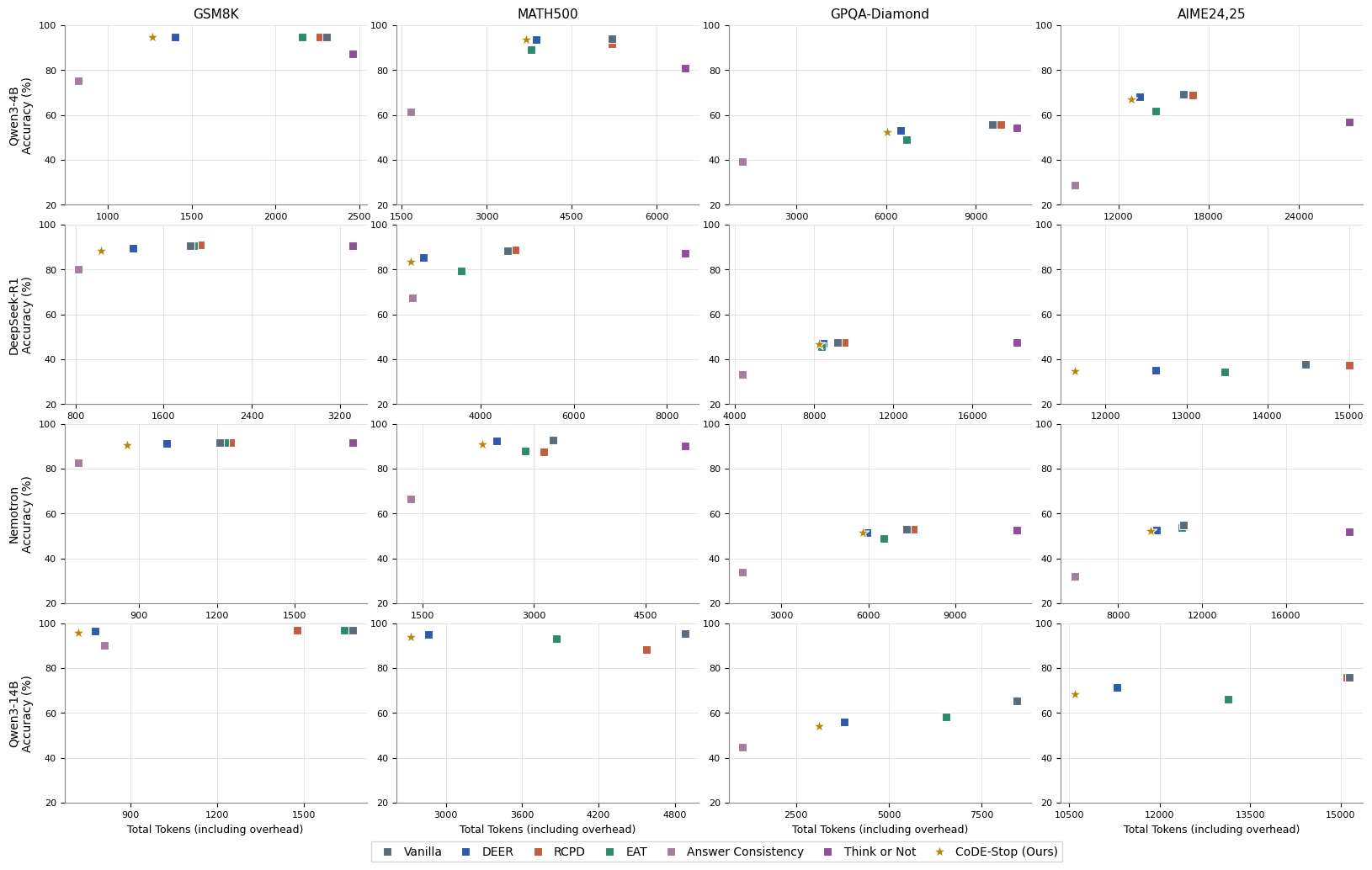}
    \caption{\textbf{Performance of CoDE-Stop against different baselines across multiple benchmarks.} CoDE-Stop consistently reduces inference cost while maintaining comparable performance to the baselines, making it Pareto optimal.}
    \label{fig:main_results}
\end{figure}

\subsection{Setup}
\textbf{Models.} We evaluate CoDE-Stop on strong and diverse open-source reasoning models: Qwen3-4B,14B~\citep{qwen3technicalreport},, DeepSeek-R1-Distill-Llama-8B~\citep{deepseekai2025deepseekr1incentivizingreasoningcapability}, and 
Llama-3.1-Nemotron-Nano-8B-v1~\citep{bercovich2025llamanemotronefficientreasoningmodels}, all accessed via the HuggingFace. For all models in our main experiments, we set the \texttt{max\_new\_tokens} to 32K. We provide the full generation and inference configurations for each model in the appendix.

\textbf{Benchmarks.} We evaluate CoDE-Stop on a diverse set of reasoning and science benchmarks: AIME 2024/2025 (60 questions)~\citep{aime_aops}, MATH500 (500)~\citep{math500}, GSM8K (1,319)~\citep{cobbe2021gsm8k}, and GPQA-Diamond (198)~\citep{rein2024gpqa}.

\textbf{Prompting Baselines.} Prior work has explored improving reasoning efficiency through prompting strategies. These approaches are orthogonal to CoDE-Stop and can be combined with our method. We evaluate four prompting-based baselines, including Vanilla prompting, Budget Forcing, Chain-of-Draft (CoD)~\citep{chainofdraft}, and NoThinking~\citep{nothinking}. We provide the exact prompts used for each method in the Appendix~\ref{sec:prompts}.

\textbf{Evaluation Setup.} For each prompt, we generate multiple reasoning trajectories per question: 15 for AIME, 2 for MATH500, 1 for GSM8K, and 5 for GPQA-Diamond, yielding about 1,000 trajectories per benchmark. Then, to reduce generation variance effect and run efficiently, we use these trajectories in all the baselines.

\textbf{Baselines.} We compare CoDE-Stop with several inference-time early stopping methods that rely on intermediate answer generation. As a reference, we also include a Vanilla setting, where the model is prompted normally without early stopping; to ensure a fair comparison, we force the model to produce a final answer when it reaches the maximum token budget, consistent with other baselines that explicitly trigger answer generation. We consider the following baselines: Think or Not~\citep{thinkornot}, DEER~\citep{deer}, EAT~\citep{EAT}, RCPD~\citep{RCPD}, and Answer Convergence~\citep{answerconv}.

% \textbf{Hyperparameters.} All methods considered in Table~\ref{tab:main_results} involve hyperparameters that control the accuracy--compute tradeoff. For baseline methods, we either use the best reported configurations from their original papers or sweep over their hyperparameters and report the configuration that achieves the best pareto frontier tradeoff between accuracy and compute.  We follow the same procedure for CoDE-Stop. All hyperparameter settings for both our method and the baselines are provided in the appendix.

\textbf{Hyperparameters.} All methods considered in Table~\ref{tab:main_results} have hyperparameters that control the accuracy--compute tradeoff. For baseline methods, we use the configurations reported in their original papers when applicable; otherwise, we sweep over a set of hyperparameter values and select the configuration that provides the best accuracy--compute tradeoff under our evaluation setup. For CoDE-Stop, we choose hyperparameters on a small held-out validation set: we tune $(\textit{steps}, r_{\min}, \tau)$ on AMC23~\citep{mathai_amc23} (40 questions with 2 rollouts each), finding that $r_{\min} = 0.9$ and 2 ramping steps work best across all models. The degeneration threshold $\tau$ is then transferred to each benchmark by scaling it with the benchmark's average reasoning length, following a scaling analysis of the degeneration score; Appendix~\ref{app:hp_transfer} provides the analysis and the full procedure. We further analyze the sensitivity of our method to its hyperparameters in Section~\ref{sec:threshold_sens}, showing that CoDE-Stop exhibits a smooth and well-behaved tradeoff, making it easy to tune in practice. To ensure transparency, all hyperparameter settings for both our method and the baselines are provided in the appendix.

\textbf{Metrics.} We report four metrics. \textbf{Acc} denotes accuracy averaged over all trajectories. \textbf{Tok} denotes the average reasoning length (in tokens) per trajectory. \textbf{CR} denotes the compression rate relative to the vanilla setting, following~\citet{deer}. \textbf{Cost} denotes the average total token usage per trajectory, including both reasoning tokens and the overhead from intermediate answer generation. %In the appendix, we also report \textbf{Tok Inc} and \textbf{Cost Inc}, which measure the same quantities as Tok and Cost but restricted to incorrect trajectories.

\begin{figure}
    \centering
    \includegraphics[width=\linewidth]{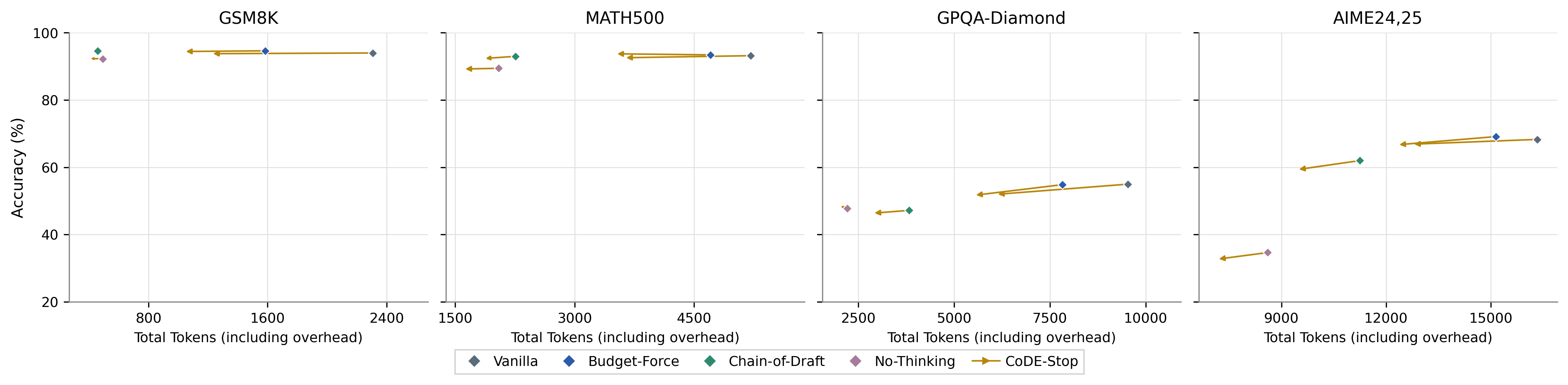}
    \caption{\textbf{Performance of CoDE-Stop with different prompting baselines.} CoDE-Stop can be combined with various prompting strategies to further improve performance.}
    \label{fig:prompts}
\end{figure}

\subsection{Results}

Figure~\ref{fig:main_results} shows the accuracy--compute tradeoff across four benchmarks and four models. The x-axis reports the total number of tokens generated (including overhead), and the y-axis shows accuracy. Each point corresponds to a method under its best-performing configuration. Across all models and datasets, CoDE-Stop consistently achieves a more favorable accuracy--compute tradeoff than baseline methods. The exact numerical results corresponding to these plots are reported in Table~\ref{tab:main_results}, and we provide some qualitative examples in Figure~\ref{fig:qual_appendix1} and \ref{fig:qual_appendix2}.

Compared to DEER, the most closely related baseline, CoDE-Stop consistently achieves a better accuracy--compute tradeoff. While DEER relies on a fixed confidence threshold and cannot terminate long, unproductive incorrect trajectories, our method additionally captures persistent instability, enabling earlier stopping of such cases.

We further evaluate CoDE-Stop under different prompting strategies on Qwen3-4B. Figure~\ref{fig:prompts} shows the accuracy--compute tradeoff for Vanilla, Budget-Force, Chain-of-Draft, and No-Thinking prompts. CoDE-Stop often improves the tradeoff across these settings by reducing token usage while maintaining or improving accuracy. Notably, even when applied to efficient prompting strategies such as Chain-of-Draft, our method provides additional gains, demonstrating that CoDE-Stop is complementary to prompt-based efficiency techniques and can further reduce unnecessary reasoning across diverse prompting paradigms.

\section{Ablation Studies}

\subsection{Length as a Proxy for Unproductive Reasoning}

We study whether trajectory length alone is sufficient to identify and stop unproductive reasoning. As shown in Figure~\ref{fig:steps_freq}, incorrect trajectories tend to be significantly longer than correct ones, suggesting that length may serve as a useful signal for early stopping.

\begin{wrapfigure}{r}{0.48\linewidth}
    \centering
    \includegraphics[width=\linewidth]{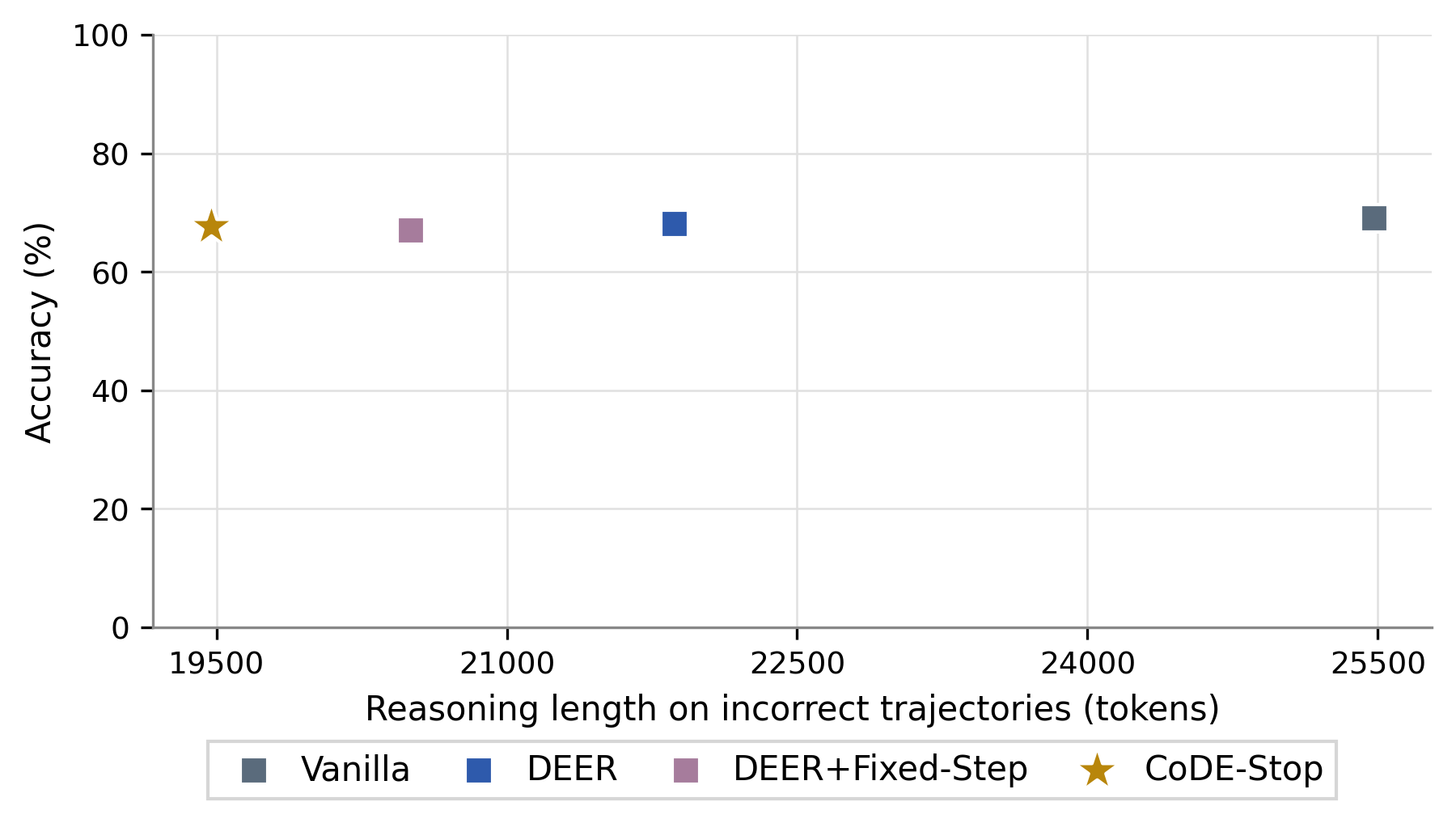}
    \caption{CoDE-Stop reduces unnecessary computation on incorrect rollouts compared to baselines at matched accuracy.}
    \label{fig:full_traj}
\end{wrapfigure}

To evaluate this, we compare three stopping strategies in Figure~\ref{fig:full_traj}, where the x-axis shows the average number of tokens on incorrect trajectories. We first consider DEER, which stops only when confidence exceeds a high threshold and therefore cannot terminate trajectories that remain uncertain yet incorrect. Next, we introduce a simple baseline, \textit{DEER+Fixed-Step}, which stops either when DEER triggers or when the number of reasoning steps exceeds a fixed threshold (we use 40). This baseline leverages trajectory length as a proxy for unproductive reasoning, without using confidence signals on incorrect trajectories.

DEER+Fixed-Step reduces unnecessary computation compared to DEER, confirming that trajectory length is indeed a useful signal for detecting unproductive reasoning. However, CoDE-Stop achieves a better accuracy--compute tradeoff, further reducing the length of incorrect trajectories at matched accuracy.  This demonstrates that while length provides a coarse signal for identifying unproductive reasoning, incorporating confidence dynamics---through our degeneration score---leads to more effective early stopping. In Appendix~\ref{app:failure_modes}, we further show that high-degeneration trajectories exhibit concrete reasoning pathologies, such as repetitive looping and unstable answer switching, beyond merely being long.

\subsection{Ablating degeneration score functions}
\label{sec:ablation}
In this section, we ablate different choices for the $v$ and $w$ functions in the degeneration score defined in Eq.~\ref{eq:deg_score}, to motivate our design choices.

We first study the degeneration function $v$, which is designed to assign high values to unconfident reasoning steps and low values to confident ones. We compare four variants: \textit{Confidence Complement} ($1 - c_i$), \textit{Confidence Drop} ($c_{i-1} - c_i$), \textit{Low Confidence} ($\mathbb{1}(c_i < \delta)$), and our \textit{Trend-aware} score ($\mathbb{1}(2c_i - c_{i-1} < \delta)$).

\begin{figure}[t]
    \centering
    \begin{minipage}[t]{0.48\linewidth}
        \centering
        \includegraphics[width=\linewidth]{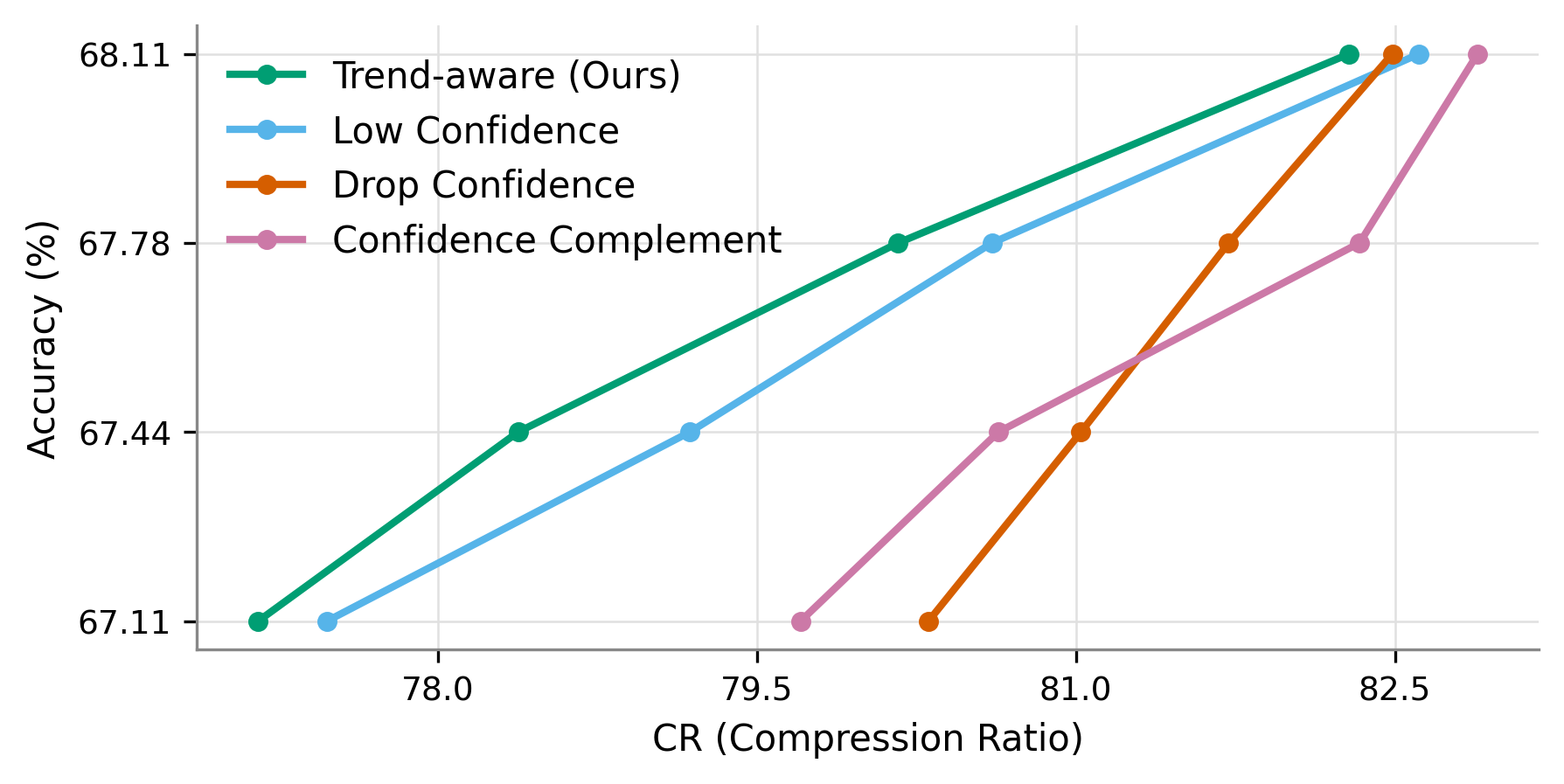}
    \end{minipage}
    \hfill
    \begin{minipage}[t]{0.48\linewidth}
        \centering
        \includegraphics[width=\linewidth]{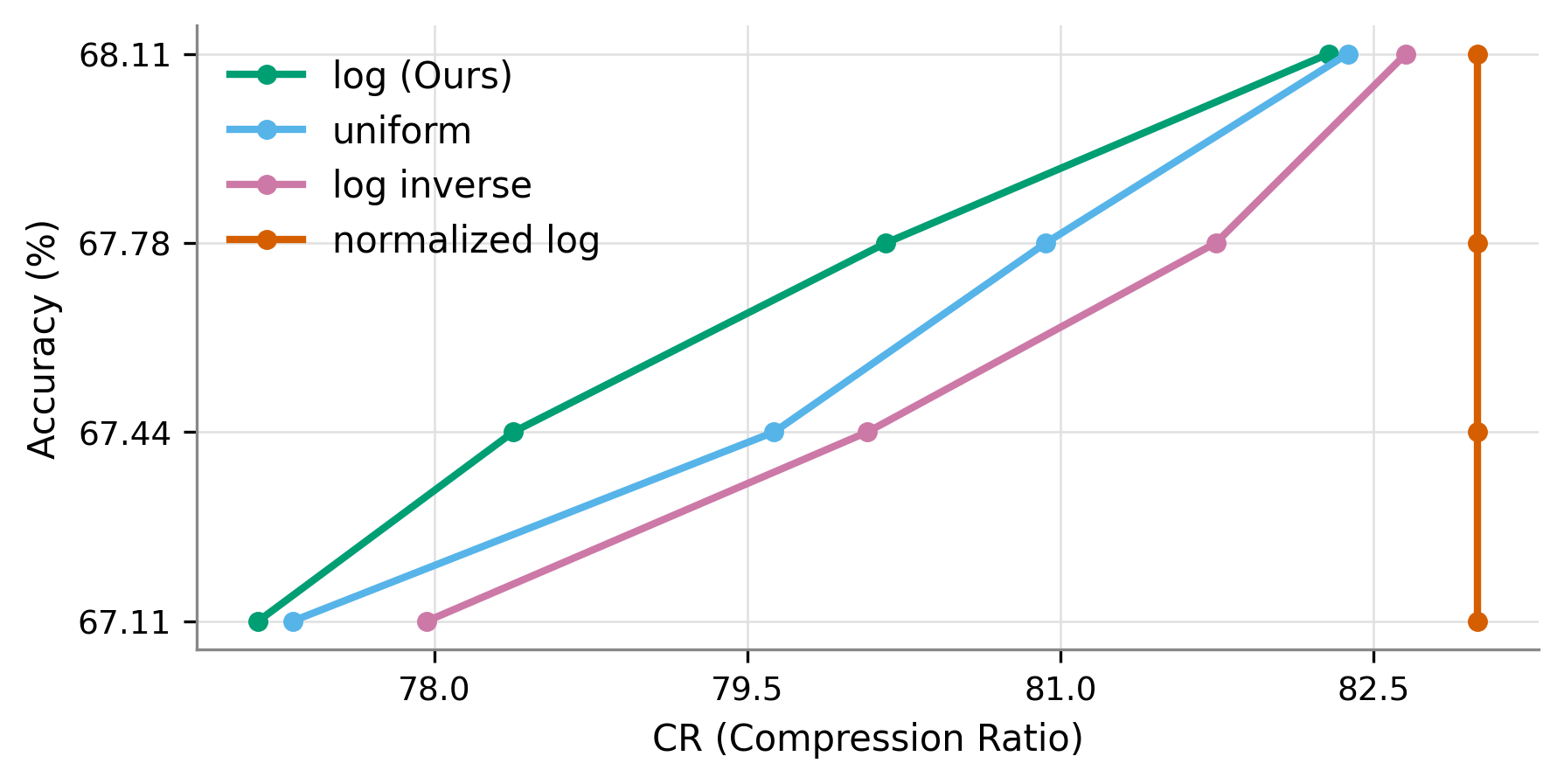}
    \end{minipage}
    \vspace{-3mm}
    \caption{\textbf{Ablation of degeneration score functions:} $v_i$ (left) and $w_i$ (right)}
    \label{fig:vw_ablation}
\end{figure}

As shown in Figure~\ref{fig:vw_ablation} (Left), our trend-aware score achieves the best accuracy–compression tradeoff across all settings. Methods based on raw confidence (Confidence Complement and Confidence Drop) are less effective due to noise and model-dependent calibration. Compared to Low Confidence, our method performs better because it not only triggers when confidence is low, but also captures sudden drops in confidence, allowing it to better identify unconfident reasoning trajectories.

We next ablate the weighting function $w$, while fixing $v$ to our trend-aware score. Let $T_k$ denote the token index corresponding to reasoning step $k$, and define $r = T_K / T_k$, where $K$ is the current reasoning step. We compare four variants: \textit{log} weighting (ours), where $w_k =\log(r)+1$, \textit{uniform} weighting, where $w_k=1$, \textit{log inverse} weighting, where $w_k = \log(1/r)+1$ and thus emphasizes later steps, and a \textit{normalized log} variant where log weights are normalized to sum to one.

As shown in Figure~\ref{fig:vw_ablation} (Right), log weighting achieves the best accuracy–compression tradeoff, followed by uniform weighting, while log inverse performs the worst. This supports our earlier observation that earlier reasoning steps are more informative for distinguishing correct and incorrect trajectories, and should therefore be weighted more heavily. Finally, we observe that normalizing the weights significantly degrades performance, resulting in near-zero compression gains. This is because trajectory length itself provides a strong signal for identifying unproductive reasoning, as discussed in the previous section, and normalization removes this signal by equalizing contributions across steps.

\subsection{Other Ablations}
\label{sec:threshold_sens}

\begin{figure}
    \centering
    \includegraphics[width=\linewidth]{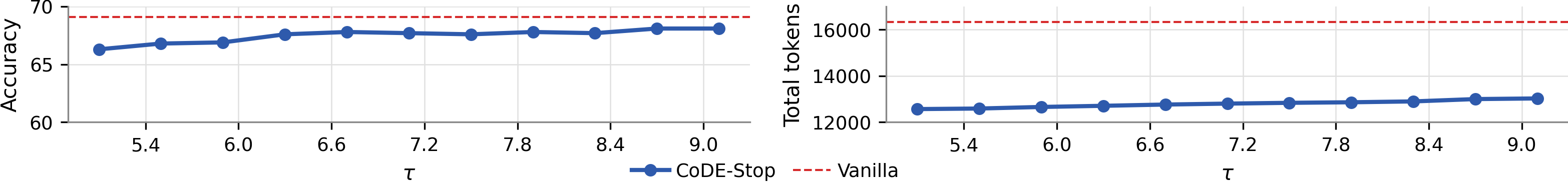}
    \caption{\textbf{Threshold sensitivity:} $\tau$ controls a smooth accuracy--compute tradeoff.}
    \label{fig:sensitivity}
\end{figure}

\textbf{Threshold Sensitivity.}  We analyze the sensitivity of CoDE-Stop to the degeneration threshold \( \tau \). Figure~\ref{fig:sensitivity} shows results on Qwen3-4B evaluated on the AIME dataset, where we vary \( \tau \) and report both accuracy and total token usage. We observe a smooth tradeoff between accuracy and compute, indicating that CoDE-Stop is robust to the choice of \( \tau \) and can be easily tuned to meet different compute budgets. Indeed, as described in Appendix~\ref{app:hp_transfer}, we choose \( \tau \) once on a small validation set and transfer it across benchmarks through a simple scaling rule based on average reasoning length.

\textbf{Reasoning Step Delimiter.} We used the token corresponding to ``Wait'' as the delimiter between reasoning steps and marking the points at which early stopping can occur. This follows prior work that reasoning models tend to include self-reflection words such as ``Wait'' often during their generation \citep{deer,EAT}. Following \citet{deer}, we additionally experiment with using ``Alternatively'' as a different delimiter. The results are found in Table~\ref{tab:alternatively}; we observe similar accuracies and efficiency gain as in our main paper, suggesting that CoDE-Stop is robust to a reasonable choice of reasoning step delimiter.

\textbf{Lower Budget.} In all experiments, we set the maximum number of new tokens to 32K. We additionally report results with a lower budget of 16K in Table~\ref{tab:B_results}. CoDE-Stop remains effective in this regime, however, the gains are more pronounced at larger budgets, where incorrect trajectories tend to grow substantially longer, increasing the opportunity for degeneration score to contribute to early stopping.

\section{Related Works}
Prior work has explored improving the efficiency of reasoning in LLMs through training methods, early stopping during inference, and probing internal model states to predict reasoning correctness. We briefly review these directions and position our approach within the literature.

\textbf{Efficient Reasoning in Large Language Models.} Several works aim to improve reasoning efficiency in large language models through approaches such as concise reasoning training, chain-of-thought compression, or latent reasoning ~\citep{arora2025traininglanguagemodelsreason,li2025compressingchainofthoughtllmsstep,munkhbat2025self,tan2025think,aytes2025sot,xu2025softcot,yi2025shorterbetter}. These approaches require additional training or modifications to the model, whereas our method operates at inference time and requires no additional training.

\textbf{Early Stopping for Reasoning Models}. Recent work has proposed several methods for early stopping in reasoning LLMs. Some approaches encourage shorter or truncated reasoning trajectories directly~\citep{chainofdraft,nothinking}. Other methods terminate reasoning based on signals observed during generation, such as answer convergence~\citep{answerconv,Dynasor}. Several works instead rely on model confidence or uncertainty signals to decide when the model is confident enough to stop reasoning~\citep{deer,thinkornot,deepconf,EAT,RCPD}, while \citet{conformalthinking} extend this idea with a dual-threshold framework that also stops unsolvable problems. In contrast, our approach models the temporal evolution of confidence during reasoning.

% Self-Certainty~\citep{selfcertainty}

\textbf{Predicting Reasoning Success in LLMs.} Recent work shows that hidden representations of LLMs encode signals about reasoning success. By probing these internal states, several methods can predict the correctness of reasoning trajectories or final answers early~\citep{zhang2025reasoningmodelsknowtheyre,cencerrado2026answerneededpredictingllm,afzal2025knowingsayingllmrepresentations,arithmeticerrorprobing}, suggesting that such signals could enable early termination of reasoning. However, these approaches typically require training probes or auxiliary predictors, whereas our method directly leverages confidence dynamics during generation without additional training.

\section{Conclusion}

In this work, we study the confidence dynamics of large reasoning models for early stopping and introduce CoDE-Stop, a method that combines a ramping confidence threshold with a degeneration score to terminate both successful and unproductive reasoning. CoDE-Stop achieves a strong accuracy--compute tradeoff by reducing unnecessary reasoning while maintaining performance. More broadly, our results show that simple heuristics over confidence---derived from model outputs---already provide effective signals for early stopping, indicating that richer information exists in model activations. Learning to extract such signals directly from model representations is a promising direction for future work.

\section*{Acknowledgments}
% This project was supported in part by a grant from an NSF CAREER AWARD 1942230, the ONR PECASE grant N00014-25-1-2378, ARO’s Early Career Program Award 310902-00001, Army Grant No. W911NF2120076, the NSF award CCF2212458, NSF Award No. 2229885 (NSF Institute for Trustworthy AI in Law and Society, TRAILS), a MURI grant 14262683,  DARPA AIQ grant HR00112590066 and an award from meta 314593-00001. 

This project was supported in part by a grant from an NSF CAREER AWARD 1942230, the ONR PECASE grant N00014-25-1-2378, ARO’s Early Career Program Award 310902-00001, Army Grant No. W911NF2120076, the NSF award CCF2212458, NSF Award No. 2229885 (NSF Institute for Trustworthy AI in Law and Society, TRAILS), a MURI grant 14262683, DARPA AIQ grant HR00112590066, an award from meta 314593-00001, the USC Office of Research and Innovation: Generative AI Research Award, the NSF NRT Program, and the USC-Meta Center for Research and Education in AI and Learning.

\bibliography{colm2026_conference}
\bibliographystyle{colm2026_conference}

\appendix

\section{Prompts}
\label{sec:prompts}

We provide the prompts used in our experiments. The \textit{Vanilla} prompt is the default prompt used in our main experiments. The \textit{Budget Force} prompt explicitly specifies a token budget (set according to the maximum number of new tokens), following prior work~\citep{deer,budgetforce}. For \textit{Chain-of-Draft (CoD)}, we follow the prompting strategy in~\citep{chainofdraft}. The \textit{No Thinking} prompt is adopted from~\citep{nothinking}.

\begin{tcolorbox}
\ttfamily\small
Vanilla = Please reason step by step, and put your final answer within \textbackslash boxed\{\}.
\end{tcolorbox}

\begin{tcolorbox}
\ttfamily\small
Budget Force =  IMPORTANT: You may think for up to [B] tokens. Ensure your response remains within the token limit. Put your final answer in \textbackslash boxed\{\}.
\end{tcolorbox}

\begin{tcolorbox}
\ttfamily\small
Chain-of-Draft = Please reason step by step, but only keep a minimal draft for each step (at most 10 words). Put your final answer within \textbackslash boxed\{\}. \\

Example 1: \\
Question: There are 15 trees in the grove. Grove workers will plant trees in the grove today. \\
After they are done, there will be 21 trees. How many trees did the grove workers plant today? \\
Answer: 21 - 15 = 6. \textbackslash boxed\{6\} \\

Example 2: \\
Question: If there are 3 cars in the parking lot and 2 more cars arrive, how many cars are in the parking lot? \\
Answer: 3 + 2 = 5. \textbackslash boxed\{5\} \\

Now answer the following question.
\end{tcolorbox}

\begin{tcolorbox}
\ttfamily\small
No Thinking = Please reason step by step, and put your final answer within \textbackslash boxed\{\}. \\
Okay, I think I have finished thinking. </think>
\end{tcolorbox}

\section{Qualitative Examples}
\label{appendix:qual_figures}
\begin{figure}
    \centering
    \includegraphics[width=\linewidth]{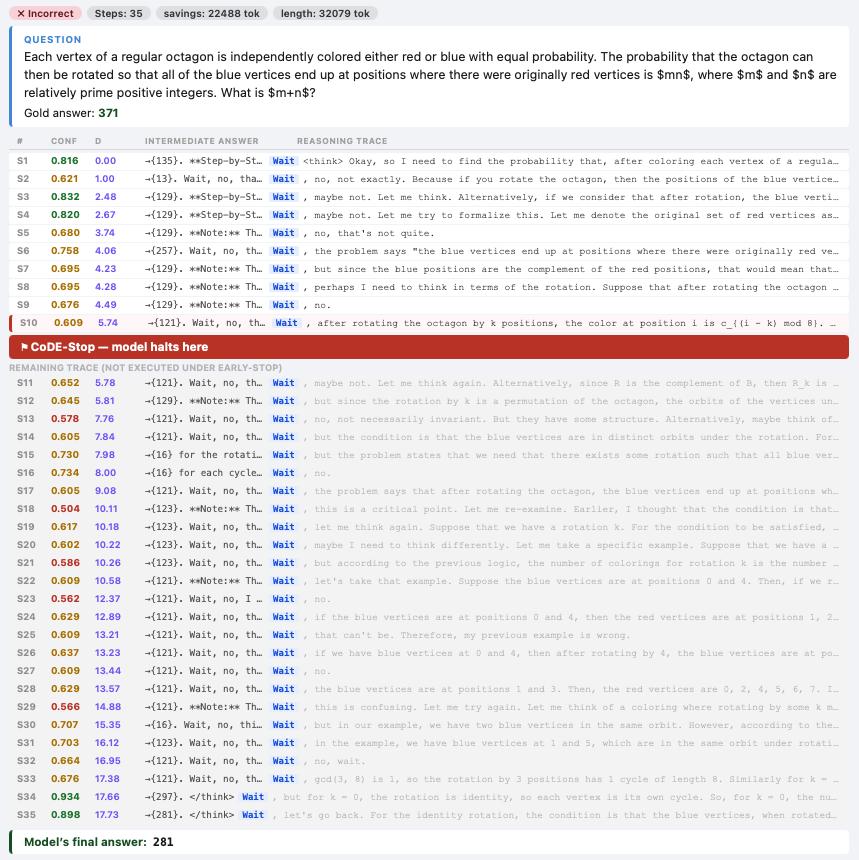}
    \caption{\textbf{Qualitative Example 1.}}
    \label{fig:qual_appendix1}
\end{figure}

We provide two qualitative examples of early stopping behavior in Figures~\ref{fig:qual_appendix1} and \ref{fig:qual_appendix2}. A qualitative example of a high-degeneration trajectory is provided in Appendix~\ref{app:failure_modes} (Figure~\ref{fig:loop_example}).

\begin{figure}
    \centering
    \includegraphics[width=\linewidth]{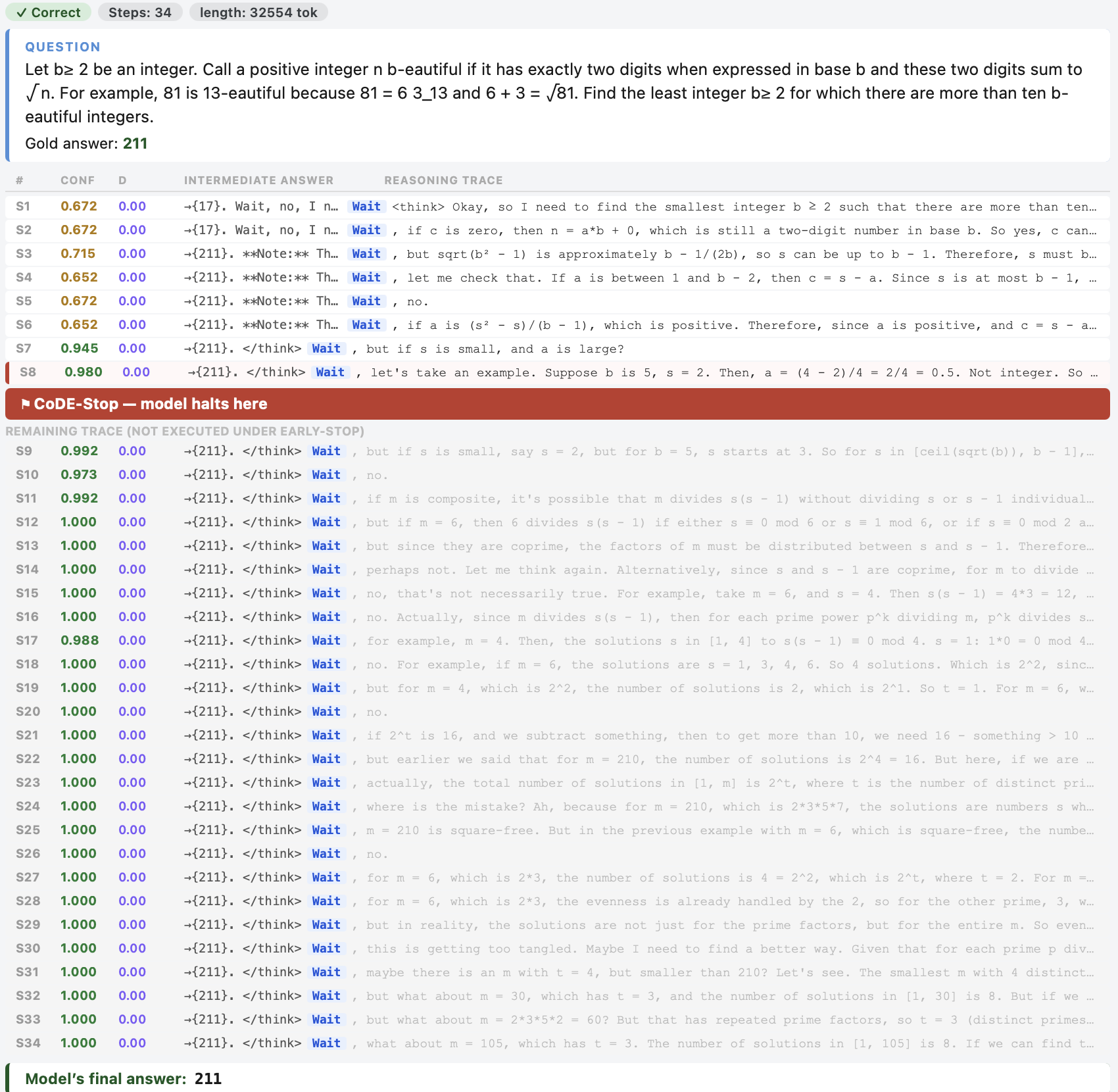}
    \caption{\textbf{Qualitative Example 2.}}
    \label{fig:qual_appendix2}
\end{figure}

\section{Results}
We provide a detailed breakdown of the performance of CoDE-Stop and various baselines in Table~\ref{tab:main_results}.

\begin{table*}[t]
\centering
\small
\setlength{\tabcolsep}{4pt}
\renewcommand{\arraystretch}{1.15}
\resizebox{\textwidth}{!}{
\begin{tabular}{llcccccccccccccccccccc}
\toprule
\multirow{3}{*}{\textbf{Model}} & \multirow{3}{*}{\textbf{Method}} 
& \multicolumn{4}{c}{\textbf{AIME24,25}} 
& \multicolumn{4}{c}{\textbf{MATH500}} 
& \multicolumn{4}{c}{\textbf{GSM8K}} 
& \multicolumn{4}{c}{\textbf{GPQA-Diamond}} 
& \multicolumn{4}{c}{\textbf{Overall}} \\
\cmidrule(lr){3-6} \cmidrule(lr){7-10} \cmidrule(lr){11-14} \cmidrule(lr){15-18} \cmidrule(lr){19-22}
& 
& \textbf{Acc $\uparrow$} & \textbf{Tok $\downarrow$} & \textbf{CR $\downarrow$} & \textbf{Cost $\downarrow$}
& \textbf{Acc $\uparrow$} & \textbf{Tok $\downarrow$} & \textbf{CR $\downarrow$} & \textbf{Cost $\downarrow$}
& \textbf{Acc $\uparrow$} & \textbf{Tok $\downarrow$} & \textbf{CR $\downarrow$} & \textbf{Cost $\downarrow$}
& \textbf{Acc $\uparrow$} & \textbf{Tok $\downarrow$} & \textbf{CR $\downarrow$} & \textbf{Cost $\downarrow$}
& \textbf{Acc $\uparrow$} & \textbf{Tok $\downarrow$} & \textbf{CR $\downarrow$} & \textbf{Cost $\downarrow$} \\
\midrule

\multirow{7}{*}{\textbf{Qwen3-4B}}
& Vanilla            & 69.1 & 16326 & 100.0 & 16326 & 94.0 & 5210 & 100.0 & 5210 & 94.8 & 2306 & 100.0 & 2306 & 55.8 & 9536 & 100.0 & 9536 & 78.4 & 8344 & 100.0 & 8344 \\
& Think or Not       & 56.7 & 14130 & 86.5 & 27343 & 80.8 & 3768 & 72.3 & 6497 & 87.4 & 1618 & 70.2 & 2462 & 54.1 & 6865 & 72.0 & 10371 & 69.8 & 6595 & 75.3 & 11668 \\
& DEER               & 68.1 & 13115 & 80.3 & 13400 & 93.7 & 3803 & 73.0 & 3878 & 94.8 & 1374 & 59.6 & 1400 & 53.0 & 6412 & 67.2 & 6496 & 77.4 & 6176 & 70.0 & 6294 \\
& EAT                & 61.7 & 13995 & 85.7 & 14489 & 89.0 & 3642 & 69.9 & 3789 & 94.6 & 2092 & 90.7 & 2161 & 49.0 & 6459 & 67.7 & 6670 & 73.6 & 6547 & 78.5 & 6777 \\
& RCPD               & 68.7 & 16325 & 100.0 & 16960 & 91.9 & 4994 & 95.9 & 5211 & 94.7 & 2140 & 92.8 & 2264 & 55.8 & 9424 & 98.8 & 9825 & 77.8 & 8221 & 96.9 & 8565 \\
& Answer Convergence & 28.9 & 5958 & 36.5 & 9073 & 61.5 & 897 & 17.2 & 1656 & 75.1 & 401 & 17.4 & 825 & 39.1 & 947 & 9.9 & 1210 & 51.1 & 2051 & 20.3 & 3191 \\
% & CoDE-Stop (Ours)   & 67.7 & 12588 & 77.1 & 12800 & 93.4 & 3575 & 68.6 & 3647 & 94.6 & 1208 & 52.4 & 1233 & 52.8 & 6064 & 63.6 & 6144 & 77.1 & 5859 & 65.4 & 5956 \\
& CoDE-Stop (Ours)   & 67.0 & 12630 & 77.4 & 12837 & 93.5 & 3624 & 69.6 & 3696 & 94.6 & 1243 & 53.9 & 1268 & 52.4 & 5961 & 62.5 & 6038 & 76.9 & 5865 & 65.9 & 5960 \\
\midrule

\multirow{7}{*}{\textbf{Qwen3-14B}}
& Vanilla            & 76.0 & 15137 & 100.0 & 15137 & 95.3 & 4878 & 100.0 & 4878 & 96.7 & 1668 & 100.0 & 1668 & 65.2 & 8447 & 100.0 & 8447 & 83.3 & 7532 & 100.0 & 7532 \\
% & Think or Not       & -- & -- & -- & -- & -- & -- & -- & -- & -- & -- & -- & -- & -- & -- & -- & -- & -- & -- & -- & -- \\
& DEER               & 71.3 & 11086 & 73.2 & 11295 & 94.8 & 2808 & 57.6 & 2862 & 96.4 & 763 & 45.7 & 778 & 56.0 & 3763 & 44.5 & 3799 & 79.6 & 4605 & 55.3 & 4684 \\
& EAT                & 66.1 & 12666 & 83.7 & 13130 & 93.0 & 3730 & 76.5 & 3871 & 96.7 & 1591 & 95.4 & 1639 & 58.3 & 6355 & 75.2 & 6533 & 78.5 & 6086 & 82.7 & 6293 \\
& RCPD               & 76.0 & 14526 & 96.0 & 15096 & 88.2 & 4379 & 89.8 & 4573 & 96.7 & 1394 & 83.6 & 1476 & 65.2 & 8098 & 95.9 & 8433 & 81.5 & 7099 & 91.3 & 7394 \\
& Answer Convergence & 41.9 & 6261 & 41.4 & 10216 & 70.9 & 920 & 18.9 & 1588 & 90.1 & 385 & 23.1 & 811 & 44.7 & 830 & 9.8 & 1060 & 61.9 & 2099 & 23.3 & 3419 \\
% & CoDE-Stop (Ours)   & 70.2 & 10720 & 70.8 & 10921 & 93.0 & 2481 & 50.9 & 2529 & 95.2 & 695 & 41.7 & 709 & 53.8 & 3128 & 37.0 & 3159 & 78.0 & 4256 & 50.1 & 4330 \\
& CoDE-Stop (Ours)   & 68.3 & 10368 & 68.5 & 10589 & 94.0 & 2671 & 54.7 & 2723 & 95.8 & 708 & 42.4 & 721 & 54.2 & 3085 & 36.5 & 3116 & 78.1 & 4208 & 50.5 & 4287 \\
\midrule

\multirow{7}{*}{\textbf{DeepSeek-R1}}
& Vanilla            & 37.6 & 14461 & 100.0 & 14461 & 88.4 & 4576 & 100.0 & 4576 & 90.7 & 1840 & 100.0 & 1840 & 47.4 & 9177 & 100.0 & 9177 & 66.0 & 7514 & 100.0 & 7514 \\
& Think or Not       & 37.0 & 14181 & 98.1 & 30495 & 87.2 & 4300 & 94.0 & 8391 & 90.6 & 1808 & 98.3 & 3319 & 47.5 & 9056 & 98.7 & 18243 & 65.6 & 7336 & 97.3 & 15112 \\
& DEER               & 35.0 & 11508 & 79.6 & 12616 & 85.2 & 2626 & 57.4 & 2762 & 89.5 & 1254 & 68.2 & 1322 & 47.0 & 8032 & 87.5 & 8468 & 64.2 & 5855 & 73.2 & 6292 \\
& EAT                & 34.3 & 12624 & 87.3 & 13472 & 79.4 & 3381 & 73.9 & 3577 & 90.7 & 1789 & 97.2 & 1870 & 45.7 & 7948 & 86.6 & 8375 & 62.5 & 6436 & 86.3 & 6824 \\
& RCPD               & 37.3 & 14478 & 100.1 & 15003 & 88.6 & 4576 & 100.0 & 4745 & 90.8 & 1840 & 100.0 & 1934 & 47.4 & 9178 & 100.0 & 9527 & 66.0 & 7518 & 100.0 & 7802 \\
& Answer Convergence & 23.4 & 6328 & 43.8 & 8934 & 67.2 & 1460 & 31.9 & 2526 & 80.2 & 518 & 28.2 & 827 & 33.3 & 2629 & 28.6 & 4394 & 51.0 & 2734 & 33.1 & 4170 \\
% & CoDE-Stop (Ours)   & 34.7 & 10649 & 73.6 & 11608 & 85.1 & 2576 & 56.3 & 2703 & 88.6 & 1056 & 57.4 & 1110 & 46.9 & 7898 & 86.1 & 8318 & 63.8 & 5545 & 68.3 & 5935 \\
& CoDE-Stop (Ours)   & 34.6 & 10662 & 73.7 & 11621 & 83.4 & 2382 & 52.0 & 2491 & 88.2 & 989 & 53.8 & 1032 & 46.8 & 7831 & 76.0 & 8247 & 63.3 & 5466 & 63.9 & 5848 \\
\midrule

\multirow{7}{*}{\textbf{Nemotron-8B}}
& Vanilla            & 55.0 & 11130 & 100.0 & 11130 & 92.8 & 3258 & 100.0 & 3258 & 91.7 & 1211 & 100.0 & 1211 & 53.2 & 7318 & 100.0 & 7318 & 73.2 & 5729 & 100.0 & 5729 \\
& Think or Not       & 52.0 & 9394 & 84.4 & 18991 & 90.3 & 2754 & 84.5 & 5034 & 91.6 & 1162 & 96.0 & 1724 & 52.8 & 6804 & 93.0 & 11142 & 71.7 & 5028 & 89.5 & 9223 \\
& DEER               & 52.7 & 9631 & 86.5 & 9847 & 92.4 & 2448 & 75.1 & 2501 & 91.1 & 984 & 81.3 & 1007 & 51.7 & 5883 & 80.4 & 5967 & 72.0 & 4736 & 80.8 & 4830 \\
& EAT                & 53.7 & 10514 & 94.5 & 11018 & 87.7 & 2765 & 84.9 & 2886 & 91.6 & 1194 & 98.6 & 1229 & 49.0 & 6334 & 86.6 & 6535 & 70.5 & 5202 & 91.1 & 5417 \\
& RCPD               & 54.9 & 10729 & 96.4 & 11169 & 87.5 & 2992 & 91.8 & 3126 & 91.7 & 1182 & 97.6 & 1253 & 53.2 & 7243 & 99.0 & 7558 & 71.8 & 5536 & 96.2 & 5776 \\
& Answer Convergence & 31.9 & 4341 & 39.0 & 5939 & 66.4 & 950 & 29.2 & 1335 & 82.6 & 421 & 34.8 & 666 & 33.9 & 1231 & 16.8 & 1656 & 53.7 & 1736 & 29.9 & 2399 \\
% & CoDE-Stop (Ours)   & 52.4 & 9282 & 83.4 & 9493 & 91.4 & 2287 & 70.2 & 2338 & 90.8 & 860 & 71.0 & 881 & 50.9 & 5634 & 77.0 & 5715 & 71.4 & 4516 & 75.4 & 4607 \\
& CoDE-Stop (Ours)   & 52.3 & 9320 & 83.7 & 9532 & 90.8 & 2250 & 69.1 & 2299 & 90.6 & 833 & 68.8 & 853 & 51.6 & 5722 & 78.2 & 5803 & 71.3 & 4531 & 75.0 & 4622 \\
\bottomrule
\end{tabular}
}
\caption{Main results comparing CoDE-Stop with baseline methods across models and benchmarks (B=32768). Acc denotes accuracy, Tok denotes average reasoning tokens, CR denotes compression rate, and Cost denotes total token compute.}
\label{tab:main_results}
\end{table*}

\begin{table*}[t]
\centering
\small
\setlength{\tabcolsep}{4pt}
\renewcommand{\arraystretch}{1.15}
\resizebox{\textwidth}{!}{
\begin{tabular}{llcccccccccccccccccccc}
\toprule
\multirow{3}{*}{\textbf{Model}} & \multirow{3}{*}{\textbf{Method}} 
& \multicolumn{4}{c}{\textbf{AIME24,25}} 
& \multicolumn{4}{c}{\textbf{MATH500}} 
& \multicolumn{4}{c}{\textbf{GSM8K}} 
& \multicolumn{4}{c}{\textbf{GPQA-Diamond}} 
& \multicolumn{4}{c}{\textbf{Overall}} \\
\cmidrule(lr){3-6} \cmidrule(lr){7-10} \cmidrule(lr){11-14} \cmidrule(lr){15-18} \cmidrule(lr){19-22}
& 
& \textbf{Acc $\uparrow$} & \textbf{Tok $\downarrow$} & \textbf{CR $\downarrow$} & \textbf{Cost $\downarrow$}
& \textbf{Acc $\uparrow$} & \textbf{Tok $\downarrow$} & \textbf{CR $\downarrow$} & \textbf{Cost $\downarrow$}
& \textbf{Acc $\uparrow$} & \textbf{Tok $\downarrow$} & \textbf{CR $\downarrow$} & \textbf{Cost $\downarrow$}
& \textbf{Acc $\uparrow$} & \textbf{Tok $\downarrow$} & \textbf{CR $\downarrow$} & \textbf{Cost $\downarrow$}
& \textbf{Acc $\uparrow$} & \textbf{Tok $\downarrow$} & \textbf{CR $\downarrow$} & \textbf{Cost $\downarrow$} \\
\midrule

\multirow{7}{*}{\textbf{Qwen3-4B}}
& Vanilla            & 62.6 & 12509 & 100.0 & 12509 & 93.4 & 4921 & 100.0 & 4921 & 95.2 & 2236 & 100.0 & 2236 & 55.9 & 9204 & 100.0 & 9204 & 76.8 & 7218 & 100.0 & 7218 \\
& Think or Not       & 62.6 & 12515 & 100.0 & 19721 & 80.3 & 3490 & 70.9 & 5702 & 88.2 & 1555 & 69.5 & 2374 & 53.0 & 6621 & 71.9 & 9271 & 71.0 & 6045 & 78.1 & 9267 \\
& DEER               & 61.9 & 10666 & 85.3 & 10756 & 93.2 & 3666 & 74.5 & 3720 & 95.2 & 1379 & 61.7 & 1405 & 53.5 & 6406 & 69.6 & 6467 & 76.0 & 5529 & 72.8 & 5587 \\
& EAT                & 59.0 & 11437 & 91.4 & 11726 & 89.8 & 3635 & 73.9 & 3767 & 94.8 & 2037 & 91.1 & 2103 & 49.9 & 6599 & 71.7 & 6785 & 73.4 & 5927 & 82.0 & 6095 \\
& RCPD               & 58.9 & 12509 & 100.0 & 13011 & 91.0 & 4718 & 95.9 & 4927 & 95.2 & 2073 & 92.7 & 2193 & 55.9 & 9105 & 98.9 & 9503 & 75.2 & 7101 & 96.9 & 7408 \\
& Answer Convergence & 25.7 & 4907 & 39.2 & 7729 & 63.1 & 940 & 19.1 & 1727 & 75.3 & 401 & 17.9 & 825 & 39.9 & 961 & 10.4 & 1221 & 51.0 & 1802 & 21.7 & 2876 \\
% & CoDE-Stop (Ours)   & 60.7 & 10384 & 83.0 & 10472 & 92.6 & 3501 & 71.1 & 3552 & 94.8 & 1158 & 51.8 & 1182 & 53.3 & 5987 & 65.0 & 6044 & 75.4 & 5258 & 67.7 & 5312 \\
& CoDE-Stop (Ours)   & 60.9 & 10541 & 84.3 & 10629 & 92.8 & 3574 & 72.6 & 3628 & 94.8 & 1237 & 55.3 & 1262 & 53.2 & 5981 & 65.0 & 6038 & 75.4 & 5333 & 69.3 & 5389 \\
\midrule

\multirow{7}{*}{\textbf{Qwen3-14B}}
& Vanilla            & 70.6 & 12096 & 100.0 & 12096 & 95.1 & 4600 & 100.0 & 4600 & 96.6 & 1824 & 100.0 & 1824 & 63.6 & 8215 & 100.0 & 8215 & 81.5 & 6684 & 100.0 & 6684 \\
& DEER               & 70.3 & 9680 & 80.0 & 9769 & 94.4 & 2744 & 59.7 & 2786 & 96.4 & 826 & 45.3 & 843 & 56.0 & 3890 & 47.4 & 3924 & 79.3 & 4285 & 0.64 & 4331 \\
& EAT                & 65.8 & 10967 & 90.7 & 11268 & 92.7 & 3630 & 78.9 & 3759 & 96.9 & 1716 & 94.1 & 1767 & 57.9 & 6282 & 76.5 & 6451 & 78.3 & 5649 & 85.0 & 5811 \\
& RCPD               & 70.9 & 11615 & 96.0 & 12081 & 88.3 & 4109 & 89.3 & 4292 & 96.6 & 1547 & 84.8 & 1638 & 63.7 & 7816 & 95.1 & 8145 & 79.9 & 6272 & 91.3 & 6539 \\
& Answer Convergence & 40.8 & 5268 & 43.6 & 8624 & 72.3 & 921 & 20.0 & 1593 & 88.2 & 388 & 21.3 & 820 & 44.6 & 851 & 10.4 & 1087 & 61.5 & 1857 & 23.8 & 3031 \\
% & CoDE-Stop (Ours)   & 69.6 & 9525 & 78.7 & 9612 & 93.2 & 2464 & 53.6 & 2503 & 94.7 & 750 & 41.1 & 766 & 53.4 & 3196 & 38.9 & 3224 & 77.3 & 3984 & 59.6 & 4026 \\
& CoDE-Stop (Ours)   & 68.7 & 9464 & 78.2 & 9551 & 94.2 & 2573 & 56.0 & 2612 & 96.0 & 781 & 42.8 & 797 & 53.0 & 3158 & 38.4 & 3187 & 78.0 & 3994 & 54.6 & 4037 \\
\midrule

\multirow{7}{*}{\textbf{DeepSeek-R1}}
& Vanilla            & 33.6 & 11651 & 100.0 & 11651 & 88.5 & 4324 & 100.0 & 4324 & 90.5 & 1812 & 100.0 & 1812 & 46.3 & 8766 & 100.0 & 8766 & 64.7 & 6638 & 100.0 & 6638 \\
& Think or Not       & 33.4 & 11510 & 98.8 & 19562 & 86.9 & 4140 & 95.7 & 7593 & 90.4 & 1775 & 98.0 & 3252 & 46.0 & 8652 & 98.7 & 16752 & 64.2 & 6519 & 97.8 & 11790 \\
& DEER               & 31.6 & 10069 & 86.4 & 10385 & 85.5 & 2615 & 60.5 & 2715 & 90.0 & 1241 & 68.5 & 1296 & 45.2 & 7689 & 87.7 & 8074 & 63.1 & 5404 & 75.8 & 5618 \\
& EAT                & 32.1 & 10776 & 92.5 & 11223 & 80.3 & 3388 & 78.4 & 3572 & 89.8 & 1733 & 95.6 & 1812 & 44.8 & 7938 & 90.6 & 8329 & 61.8 & 5959 & 89.3 & 6234 \\
& RCPD               & 33.7 & 11645 & 99.9 & 12067 & 88.4 & 4319 & 99.9 & 4476 & 90.4 & 1812 & 100.0 & 1905 & 46.3 & 8767 & 100.0 & 9101 & 64.7 & 6636 & 100.0 & 6887 \\
& Answer Convergence & 22.2 & 5748 & 49.3 & 9390 & 67.7 & 1424 & 32.9 & 2487 & 80.1 & 505 & 27.9 & 808 & 32.1 & 2679 & 30.6 & 4779 & 50.5 & 2589 & 35.2 & 4366 \\
% & CoDE-Stop (Ours)   & 31.2 & 9792 & 84.0 & 10136 & 84.2 & 2469 & 57.1 & 2557 & 89.9 & 1037 & 57.2 & 1083 & 45.1 & 7599 & 86.7 & 7974 & 62.6 & 5224 & 71.3 & 5438 \\
& CoDE-Stop (Ours)   & 31.2 & 9865 & 84.7 & 10223 & 84.2 & 2548 & 58.9 & 2643 & 90.0 & 1090 & 60.2 & 1139 & 44.8 & 7626 & 87.0 & 8009 & 62.6 & 5283 & 66.6 & 5504 \\
\midrule

\multirow{7}{*}{\textbf{Nemotron-8B}}
& Vanilla            & 53.4 & 10441 & 100.0 & 10441 & 93.2 & 3116 & 100.0 & 3116 & 91.6 & 1215 & 100.0 & 1215 & 53.3 & 7313 & 100.0 & 7313 & 72.9 & 5521 & 100.0 & 5521 \\
& Think or Not       & 50.0 & 8829 & 84.6 & 16075 & 90.9 & 2590 & 83.1 & 4641 & 91.6 & 1161 & 95.6 & 1749 & 53.1 & 6805 & 93.1 & 11022 & 71.4 & 4846 & 89.1 & 8372 \\
& DEER               & 51.9 & 9327 & 89.3 & 9476 & 92.6 & 2285 & 73.3 & 2328 & 91.1 & 972 & 80.0 & 995 & 52.1 & 5779 & 79.0 & 5860 & 71.9 & 4591 & 80.4 & 4665 \\
& EAT                & 52.8 & 9937 & 95.2 & 10329 & 88.1 & 2625 & 84.2 & 2736 & 91.1 & 1188 & 97.8 & 1223 & 49.2 & 6258 & 85.6 & 6455 & 70.3 & 5002 & 90.7 & 5186 \\
& RCPD               & 53.4 & 10080 & 96.5 & 10493 & 87.3 & 2847 & 91.4 & 2974 & 91.6 & 1186 & 97.6 & 1257 & 53.3 & 7247 & 99.1 & 7560 & 71.4 & 5340 & 96.2 & 5571 \\
& Answer Convergence & 30.2 & 4168 & 39.9 & 5706 & 67.2 & 953 & 30.6 & 1345 & 81.9 & 407 & 33.5 & 647 & 34.1 & 1181 & 16.1 & 1615 & 53.4 & 1677 & 30.0 & 2328 \\
% & CoDE-Stop (Ours)   & 51.2 & 9210 & 88.2 & 9357 & 91.5 & 2018 & 64.8 & 2056 & 90.8 & 853 & 70.2 & 874 & 51.6 & 5588 & 76.4 & 5667 & 71.3 & 4417 & 74.9 & 4488 \\
& CoDE-Stop (Ours)   & 51.0 & 9175 & 87.9 & 9321 & 92.4 & 2164 & 69.4 & 2206 & 90.4 & 848 & 70.0 & 869 & 51.2 & 5597 & 76.5 & 5677 & 71.3 & 4446 & 76.0 & 4518 \\
\bottomrule
\end{tabular}
}
\caption{Results with B=16384. Hyperparameters were tuned on AMC23 and transferred to the presented benchmarks, in the same manner as the main results.}
\label{tab:B_results}
\end{table*}

\section{Hyperparameters}

\paragraph{Hyperparameters for CoDE-Stop.} As described in Section~\ref{sec:experiments} and Appendix~\ref{app:hp_transfer}, we set the number of ramping steps to 2 and $r_{\min} = 0.9$ for all models, based on tuning on the AMC23~\citep{mathai_amc23} validation set, and we fix $r_{\max} = 0.95$. The degeneration threshold $\tau$ for each model--benchmark pair is obtained by scaling the AMC23-tuned value with the benchmark's average reasoning length (Table~\ref{tab:avg_steps}); the resulting values are listed in Table~\ref{tab:model_tv_values}.

\begin{table}[t]
\centering
\small
\setlength{\tabcolsep}{5pt}
\renewcommand{\arraystretch}{1.1}
\begin{tabular}{lccccc}
\toprule
\textbf{Model} & \textbf{AMC23} & \textbf{GSM8K} & \textbf{MATH500} & \textbf{GPQA} & \textbf{AIME} \\
\midrule
Qwen3-4B           & 9.6 & 2.5 & 5.4  & 12.6 & 14.4 \\
Qwen3-14B          & 38. & 2.0 & 5.3  & 12.5 & 15.7 \\
DeepSeek-R1-Llama-8B & 7.5 & 8.3 & 22.3 & 70.4 & 98.8 \\
Llama-3.1-8B       & 8.1 & 2.2 & 4.7  & 11.9 & 15.8 \\
\bottomrule
\end{tabular}
\caption{Average number of reasoning steps across models and benchmarks.}
\label{tab:avg_steps}
\end{table}
% \begin{table}[t]
% \centering
% \small
% \setlength{\tabcolsep}{5pt}
% \renewcommand{\arraystretch}{1.1}
% \begin{tabular}{lccccc}
% \toprule
% \textbf{Model} & \textbf{GSM8K} & \textbf{MATH500} & \textbf{GPQA} & \textbf{AIME} & \textbf{AMC23} \\
% \midrule
% Qwen3-4B           & 2.5 & 5.4  & 12.6 & 14.4 & 9.6 \\
% Qwen3-14B          & 2.0 & 5.3  & 12.5 & 15.7 & - \\
% DeepSeek-R1-Llama-8B & 8.3 & 22.3 & 70.4 & 98.8 & 38.1 \\
% Llama-3.1-8B       & 2.2 & 4.7  & 11.9 & 15.8 & - \\
% \bottomrule
% \end{tabular}
% \caption{Average number of reasoning steps across models and benchmarks.}
% \label{tab:avg_steps}
% \end{table}

\paragraph{Hyperparameters for EAT.} Following the setup in the EAT paper \cite{EAT}, we set the timescale variable, which controls the EMA smoothing in the updates, to $10$. We also set the warm-up steps, which ensure that early stopping only occurs after the warm-up stage, to $25$, consistent with the original paper. 
To determine the early stopping threshold $\delta$, we sweep values in $\{10^{-7}, 10^{-5}, 10^{-3}, 10^{-1}\}$ on a subset of AIME dataset with $20$ samples. We find that setting the threshold $\delta = 10^{-5}$ yields the best performance (Table~\ref{tab:delta_eat}).
We also evaluate different values for the number of rollouts and decoding steps used to estimate the model confidence after the \texttt{</think>} token on AIME. Increasing these values provides a less noisy estimate of the model confidence but incurs higher inference cost. We find that setting the number of rollouts to $1$ and the number of decoding steps to $2$ yields the best results.

\begin{table}[ht]
\centering
\small
\setlength{\tabcolsep}{6pt}
\renewcommand{\arraystretch}{1.15}
\begin{tabular}{c c c c c}
\toprule
& $\delta = 10^{-1}$ & $\delta = 10^{-3}$ & $\delta = 10^{-5}$ & $\delta = 10^{-7}$\\
\midrule
(Acc, Cost)
 & (20.0, 2037) & (55.0, 9425) & \textbf{(70.0, 10441)} & (70.0, 11381) \\
\bottomrule
\end{tabular}
\caption{Performance of EAT on a subset of AIME for different values of $\delta$.}
\label{tab:delta_eat}
\end{table}

\textbf{Think-or-Not.} \citet{thinkornot} is a stong baseline method that utilizes the entropy of trial answers between reasoning steps to determine an appropriate early-stopping point. However, certain features of the implementation make the method more costly than initially presented. For mathematical benchmarks (such as those utilized in our experiment), Think-or-Not generates a list of trial answers using a beam search; the entropy of the top-$K$ answers is then computed (using the token probabilities of each candidate) to estimate the entropy at the chosen point. The beam search process is computationally expensive (requiring the generation of many tokens, and a copy of the KV-cache before trial-answer generation for efficiency). Additionally, this process occurs at every sentence boundary, rather at reasoning-step boundaries as in CoDE-Stop and \citet{deer}. This adds significant overhead to this method that is not captured in the reasoning response length reductions reported (that is, Think-or-Not leads to lower reasoning lengths, but potentially high costs). Due to this high computational overhead and our limited resources, we do not evaluate this method on Qwen3-14B.

Additionally, while the manuscript \citet{thinkornot} examines different values of the entropy threshold $\alpha$ for non-mathematical benchmarks, it does not specify the threshold used for the mathematical benchmarks in their main results. In our experiments, we utilized $\alpha = 0.4$, as we found this the best balance between accuracy and efficiency. On Qwen3-4B, we also experimented with the more conservative $\alpha = 0.2$; the results are presented in Table~\ref{tab:think_or_not_alpha}.

\begin{table}[ht]
\centering
\small
\setlength{\tabcolsep}{4pt}
\begin{tabular}{lcccccccc}
\toprule
& \multicolumn{2}{c}{AIME} & \multicolumn{2}{c}{MATH500} & \multicolumn{2}{c}{GSM8K} & \multicolumn{2}{c}{GPQAD} \\
\cmidrule(lr){2-3} \cmidrule(lr){4-5} \cmidrule(lr){6-7} \cmidrule(lr){8-9}
Method & Acc. & Cost & Acc. & Cost & Acc. & Cost & Acc. & Cost \\
\midrule
vanilla & $68.70\%$ & $16329$ & $94.10\%$ & $5210$ & $94.80\%$ & $2306$ & $55.20\%$ & $9638$ \\
think\_or\_not ($\alpha=0.2$) & $68.70\%$ & $31706$ & $94.20\%$ & $9581$ & $94.80\%$ & $3846$ & $55.60\%$ & $14924$ \\
think\_or\_not ($\alpha=0.4$) & $56.70\%$ & $27343$ & $80.80\%$ & $5211$ & $87.40\%$ & $2462$ & $54.10\%$ & $10371$ \\
\bottomrule
\end{tabular}
\caption{We compare the performance of the Think-or-Not \citep{thinkornot} baseline with different entropy thresholds ($\alpha$) with Qwen3-4B. The Vanilla performance values are provided as a reference. We observe that $\alpha=0.2$ is a very conservative choice; we use $\alpha=0.4$ in the main experiments.}
\label{tab:think_or_not_alpha}
\end{table}

\section{Reasoning Step Delimiter Ablation Results}

We provide a comparison of CoDE-Stop and DEER on different delimiters for reasoning chunks in Table~\ref{tab:alternatively}.

\begin{table}[ht]
\centering
\small % Reduces font size
\setlength{\tabcolsep}{4pt} % Reduces horizontal space between columns (default is 6pt)
\begin{tabular}{llcccccccc}
\toprule
& & \multicolumn{2}{c}{AIME} & \multicolumn{2}{c}{MATH500} & \multicolumn{2}{c}{GSM8K} & \multicolumn{2}{c}{GPQAD} \\
\cmidrule(lr){3-4} \cmidrule(lr){5-6} \cmidrule(lr){7-8} \cmidrule(lr){9-10}
Delimiter & Method & Acc. & Cost & Acc. & Cost & Acc. & Cost & Acc. & Cost \\
\midrule
\multirow{2}{*}{Wait} & DEER & $68.10\%$ & $13400$ & $93.70\%$ & $3878$ & $94.80\%$ & $1400$ & $53.00\%$ & $6496$ \\
& CoDE-Stop & $67.80\%$ & $12759$ & $92.40\%$ & $3463$ & $94.50\%$ & $1300$ & $52.50\%$ & $6283$ \\
\midrule
\multirow{2}{*}{Altern.} & DEER & $68.20\%$ & $13621$ & $93.70\%$ & $3344$ & $94.70\%$ & $1173$ & $52.60\%$ & $5883$ \\
& CoDE-Stop & $67.70\%$ & $13234$ & $93.20\%$ & $3289$ & $94.50\%$ & $1133$ & $52.70\%$ & $5623$ \\
\bottomrule
\end{tabular}
\caption{Comparison of DEER and CoDE-Stop Methods with different delimiters for reasoning steps. CoDE-Stop maintains similar accuracy and cost reductions compared to this baseline for both potential delimiters, providing evidence for its robustness.}
\label{tab:alternatively}
\end{table}

\section{Scaling Behavior of the Degeneration Score and Hyperparameter Transfer}
\label{app:hp_transfer}

Recall that the degeneration score is defined as
\[
D_K = \sum_{i=1}^{K} w_i v_i,
\]
where \(K\) is the number of reasoning steps, \(v_i \in \{0,1\}\) indicates whether reasoning step \(i\) is unstable, and
\[
w_i = \log\left(\frac{T_K}{T_i}\right) + 1
\]
assigns larger weights to earlier reasoning steps.

We aim to develop a simplified intuition for how $D_K$ evolves with the number of reasoning steps $K$, in order to choose a good threshold $\tau$ for early stopping. We make a few simplifying assumptions for mathematical ease: consider a trajectory where a fraction $p$ of reasoning steps are unstable: $\mathbb{E}[v_i] = p$. Suppose that the probability of reasoning steps being incorrect is independent of its position in the sequence; that is, $v_i$ is stationary. Finally, suppose that reasoning steps are approximately the same length, so that the stopping points are equally spaced throughout generation. This gives us $T_i \approx \frac{i}{K} T_K$, so that $w_i \approx 1 + \log\frac{K}{i}$. With these assumptions, we have
\begin{align*}
    \mathbb{E}[D_K] = p \sum_{i=1}^K w_i = p \sum_{i=1}^K \left(1 + \log\frac{K}{i}\right) = p\left(K + \log \frac{K^K}{K!}\right)
\end{align*}
Using Stirling's approximation $\log(K!) \approx K \log K - K$, we have $\mathbb{E}[D_K] = 2pK$. This analysis suggests that the degeneration score scales approximately linearly with the number of reasoning steps \(K\), while \(\tau\) effectively controls the fraction of unstable reasoning steps required before stopping. Intuitively, larger reasoning budgets naturally accumulate larger degeneration scores, implying that \(\tau\) should scale proportionally with the expected reasoning length.

\begin{figure}
    \centering
    \includegraphics[width=1\linewidth]{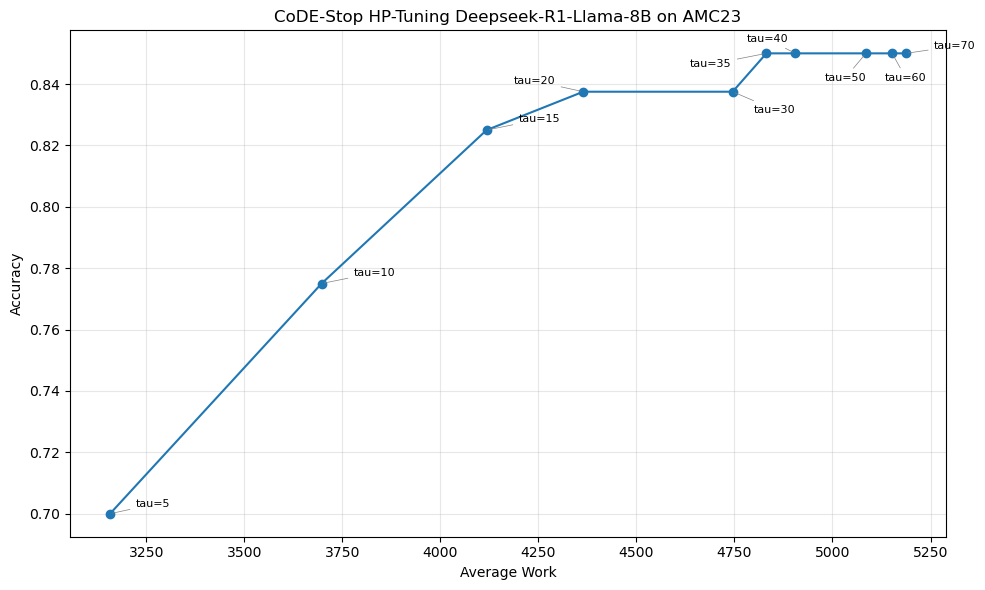}
    \caption{Hyperparameter tuning CoDE-Stop on AMC23 for Deepseek-R1-Llama-8B. For CoDE-Stop, we set $r_\text{min} = 0.9$, the number of ramping steps to be 2, and vary the degeneration threshold $\tau$. For this model, we choose $\tau = 35$ as it is the smallest threshold providing the highest accuracy in our range. Similar plots and tuning decisions were made for the other models we evaluate.}
    \label{fig:deepseek_amc23_tau_choosing}
\end{figure}

Motivated by this observation, we choose hyperparameters using a small, separate validation set. We perform inference on the AMC23~\citep{mathai_amc23} benchmark (40 questions, each with 2 rollouts for 80 total rollouts); for all models, we perform a grid search over the hyperparameters $\tau$, $r_\text{min}$, and the number of ramping steps. For all models, we find that setting $r_\text{min} = 0.9$ and the number of ramping steps to 2 gives the best performance-accuracy tradeoff on AMC23. For the threshold $\tau$, we choose the smallest threshold in our range that produces the highest accuracy; Figure~\ref{fig:deepseek_amc23_tau_choosing} provides an example using Deepseek-R1-Llama-8B (similar choices were made for all other models). 

To transfer these hyperparameters to other, larger benchmarks, we use the same $r_\text{min}$ and number of ramping steps. To obtain the threshold $\tau_t$ for the target benchmark, we transfer the tuned $\tau_s$ from the validation set (AMC23) using:
\[
\tau_t
=
\tau_s
\cdot
\frac{\bar K_t}{\bar K_s},
\]
where \(\bar K_s\) and \(\bar K_t\) denote the average number of reasoning steps on the source and target benchmarks, respectively (Table~\ref{tab:avg_steps}). To transfer $\tau$ to new datasets, one could perform a few vanilla rollouts on a small number of randomly chosen samples to estimate $\bar K_t$; this will incur little cost, and provide savings when performing full inference on the entire dataset. Scaling the tuned $\tau$ threshold for AMC23 gives us the transferred threshold values in Table~\ref{tab:model_tv_values}.

\begin{table}
\centering
\setlength{\tabcolsep}{5pt}
\renewcommand{\arraystretch}{1.1}
\begin{tabular}{lccccc}
\hline
\textbf{Model} & \textbf{AMC23} & \textbf{AIME} & \textbf{MATH500} & \textbf{GSM8K} & \textbf{GPQA-D} \\ \hline
Qwen3-4b   & 3.5 & 5.3  & 2    & 0.9 & 4.6  \\
Deepseek-R1-Llama-8B  & 35  & 90.7 & 20.5 & 7.6 & 64.7 \\
Qwen3-14b  & 3.3 & 6.9  & 2.3  & 0.9 & 5.5  \\
Nemotron-8B  & 8   & 15.6 & 4.6  & 2.2 & 11.8 \\ \hline
\end{tabular}
\caption{$\tau$ values used in hyperparameter-transfer experiments. Grid search was performed to obtain optimal $\tau$-values for AMC23. These values were scaled according to the average number of reasoning steps for each benchmark to obtain their respective thresholds.}
\label{tab:model_tv_values}
\end{table}

As a robustness check on this selection procedure, we additionally compare the transferred thresholds against thresholds tuned separately on each benchmark; the results are shown in Table~\ref{tab:transfer_results}. The transferred thresholds achieve nearly identical accuracy--compute tradeoffs, and in several cases even slightly improve them. This confirms that the degeneration threshold exhibits predictable scaling behavior across benchmarks, and that the transferred thresholds are near-optimal without any benchmark-specific tuning.

\begin{table*}[t]
\centering
\setlength{\tabcolsep}{4pt}
\small
\begin{tabular}{llcccccc}
\toprule
Model & Benchmark & Tuned Acc & Tuned Cost & Transf. Acc & Transf. Cost & $\Delta$ Acc & $\Delta$ Cost \\
\midrule

\multirow{4}{*}{Qwen3-4B}
& AIME          & 67.8 & 12759 & 67.0 & 12837 & -0.8 & +78 \\
& MATH500       & 92.4 & 3463  & 93.5 & 3696  & +1.1 & +233 \\
& GSM8K         & 94.5 & 1300  & 94.6 & 1268  & +0.1  & -32 \\
& GPQA          & 52.5 & 6283  & 52.4 & 6038  & -0.1 & -245 \\
\midrule

\multirow{4}{*}{DeepSeek-R1}
& AIME          & 34.7 & 11608 & 34.6 & 11621 & -0.1 & +13 \\
& MATH500       & 84.0 & 2657  & 83.4 & 2491  & -0.6 & -166 \\
& GSM8K         & 88.6 & 1119  & 88.2 & 1032  & -0.4 & -87 \\
& GPQA          & 46.9 & 8318  & 46.8 & 8247  & -0.1 & -71 \\
\midrule

\multirow{4}{*}{Nemotron-8B}
& AIME          & 52.4 & 9493  & 52.3 & 9532  & -0.1 & +39 \\
& MATH500       & 91.4 & 2338  & 90.8 & 2299  & -0.6 & -39 \\
& GSM8K         & 90.8 & 903   & 90.6 & 853   & -0.2 & -50 \\
& GPQA          & 51.3 & 5794  & 51.6 & 5803  & +0.3 & +9 \\
\midrule

\multirow{4}{*}{Qwen3-14B}
& AIME          & 70.2 & 10921 & 68.3 & 10589 & -1.9 & -332 \\
& MATH500       & 94.3 & 2828 & 94.0 & 2723 & -0.3 & -105 \\
& GSM8K         & 95.2 & 709 & 95.8 & 721 & +0.6 & +12 \\
& GPQA          & 53.9 & 3224 & 54.2 & 3116 & +0.3 & -108 \\
\bottomrule

\end{tabular}
\caption{Robustness check: comparison between per-benchmark tuned thresholds (Tuned) and the AMC23-based transferred thresholds used in our experiments (Transf.). The transferred thresholds achieve nearly identical accuracy--compute tradeoffs.}
\label{tab:transfer_results}
\end{table*}

\section{Degeneration Score and Reasoning Failure Modes}
\label{app:failure_modes}

\begin{figure}[t]
    \centering
    \includegraphics[width=\linewidth]{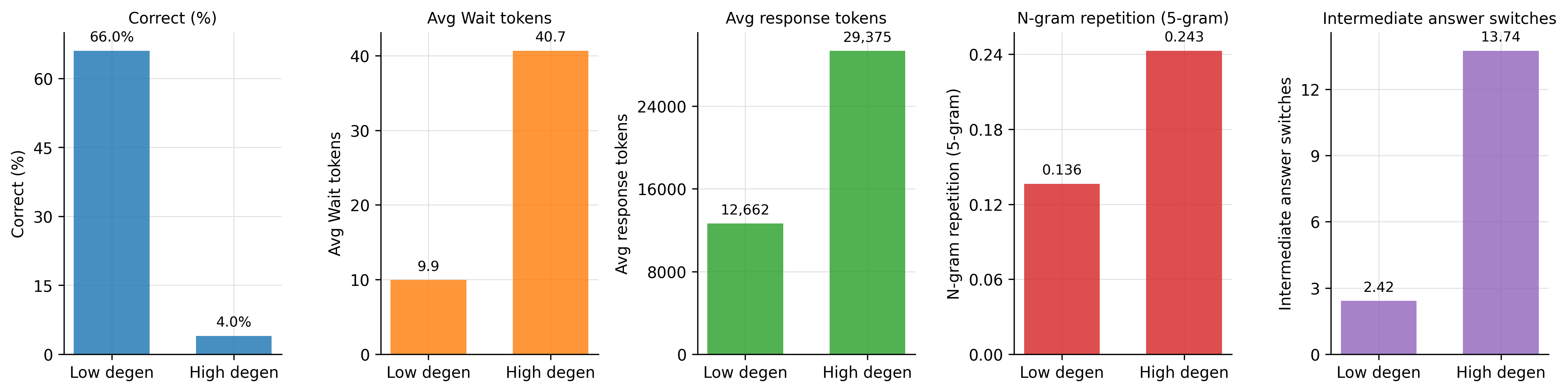}
    \caption{
    Analysis comparing high and low degeneration score trajectories.}
    \label{fig:failure_modes}
\end{figure}

To better understand what the degeneration score captures, we perform an additional analysis on the 50 highest- and 50 lowest-degeneration trajectories from AIME. Figure~\ref{fig:failure_modes} shows that high-degeneration trajectories exhibit substantially different behavior compared to low-degeneration trajectories. Specifically, high-degeneration trajectories have:
(i) significantly lower correctness,
(ii) substantially longer reasoning traces,
(iii) much higher repeated 5-gram frequency, and
(iv) dramatically more intermediate answer switches during reasoning.

These patterns suggest that high degeneration scores strongly correlate with concrete reasoning failures such as repetitive looping, unstable answer switching, and failed self-correction, rather than merely reflecting long trajectories alone.

\begin{figure}[t]
    \centering
    \includegraphics[width=\linewidth]{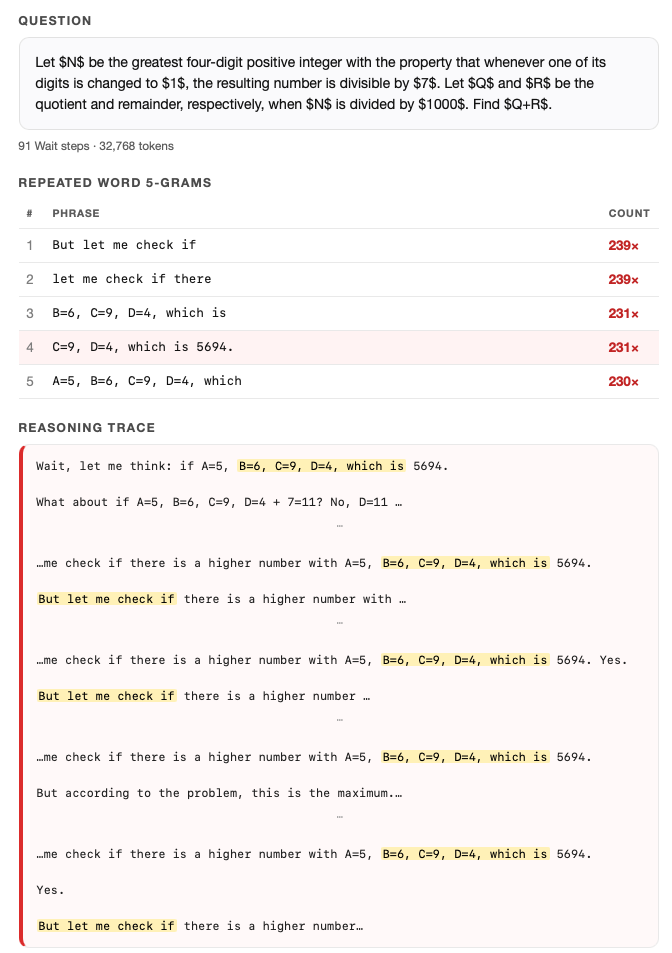}
    \caption{
    Qualitative example of a high-degeneration trajectory from AIME. The model repeatedly cycles through nearly identical reasoning patterns and revisits the same candidate solution without making meaningful progress, resulting in prolonged repetitive reasoning behavior.
    }
    \label{fig:loop_example}
\end{figure}

We additionally provide a qualitative example of a high-degeneration trajectory in Figure~\ref{fig:loop_example}. In the shown example, the model repeatedly revisits nearly identical reasoning patterns (e.g., ``C=9, D=4, which is 5694'') while cycling through the same candidate solution multiple times without making meaningful progress. The trajectory repeatedly re-enters similar reasoning states while continuing to generate tokens for a prolonged period. This behavior is precisely the type of unproductive reasoning instability that CoDE-Stop is designed to detect.

Overall, these analyses suggest that the degeneration score captures meaningful reasoning pathologies associated with unproductive long-horizon reasoning, rather than acting solely as a generic length heuristic.

\clearpage
\section{Additional Confidence Dynamics Analyses}
\label{app:extended_motivation}

In this section, we extend the motivating analyses of Section~\ref{sec:prelim} to all model--benchmark combinations, covering DeepSeek-R1-Llama-8B and Nemotron-8B on AIME, GPQA-D, MATH500, and GSM8K.

\subsection{Per-trajectory confidence dynamics}
\label{ssec:app_fig2}

\begin{figure}[h]
  \centering
  \begin{subfigure}{0.95\textwidth}
    \includegraphics[width=\linewidth]{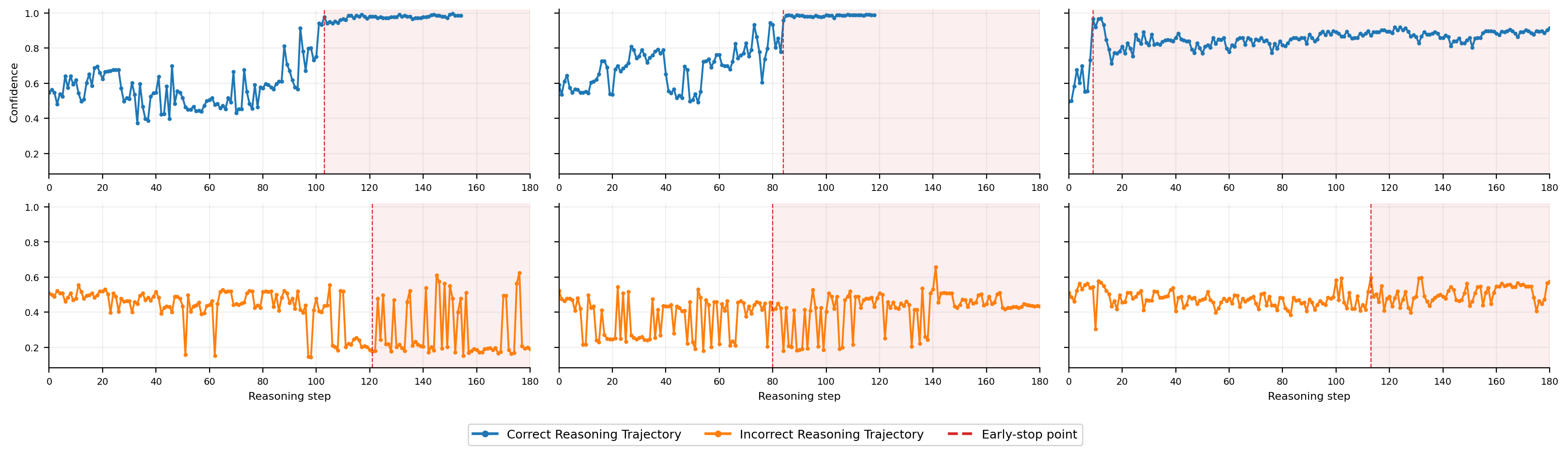}
    \caption{DeepSeek -- AIME}
  \end{subfigure}\\[2pt]
  \begin{subfigure}{0.95\textwidth}
    \includegraphics[width=\linewidth]{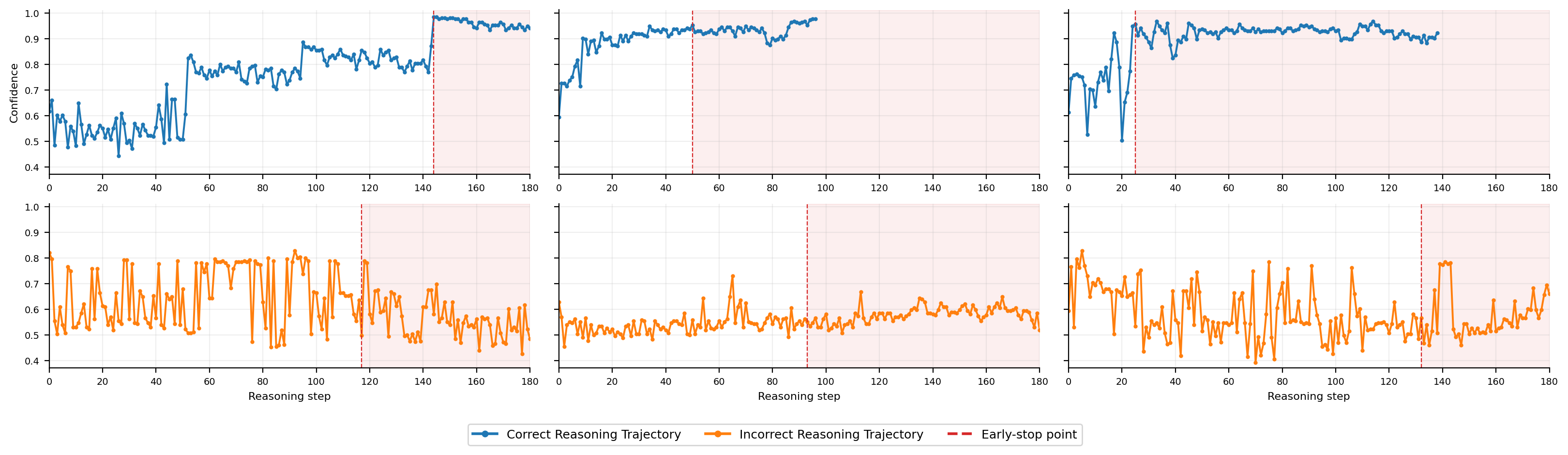}
    \caption{DeepSeek -- GPQA-D}
  \end{subfigure}\\[2pt]
  \begin{subfigure}{0.95\textwidth}
    \includegraphics[width=\linewidth]{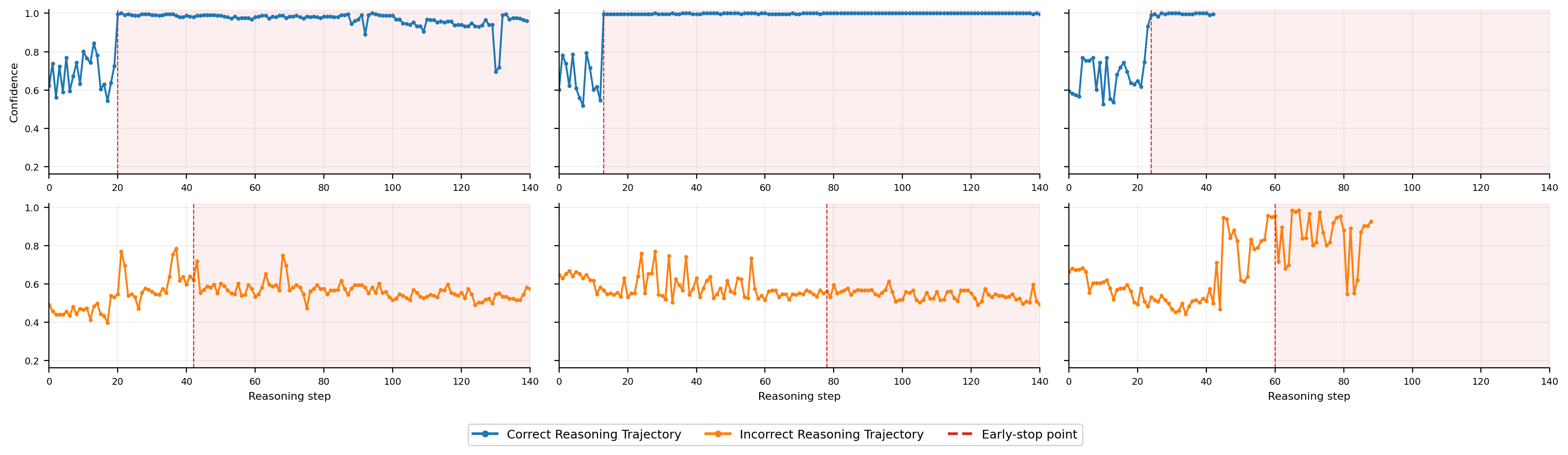}
    \caption{DeepSeek -- Math500}
  \end{subfigure}\\[2pt]
  \begin{subfigure}{0.95\textwidth}
    \includegraphics[width=\linewidth]{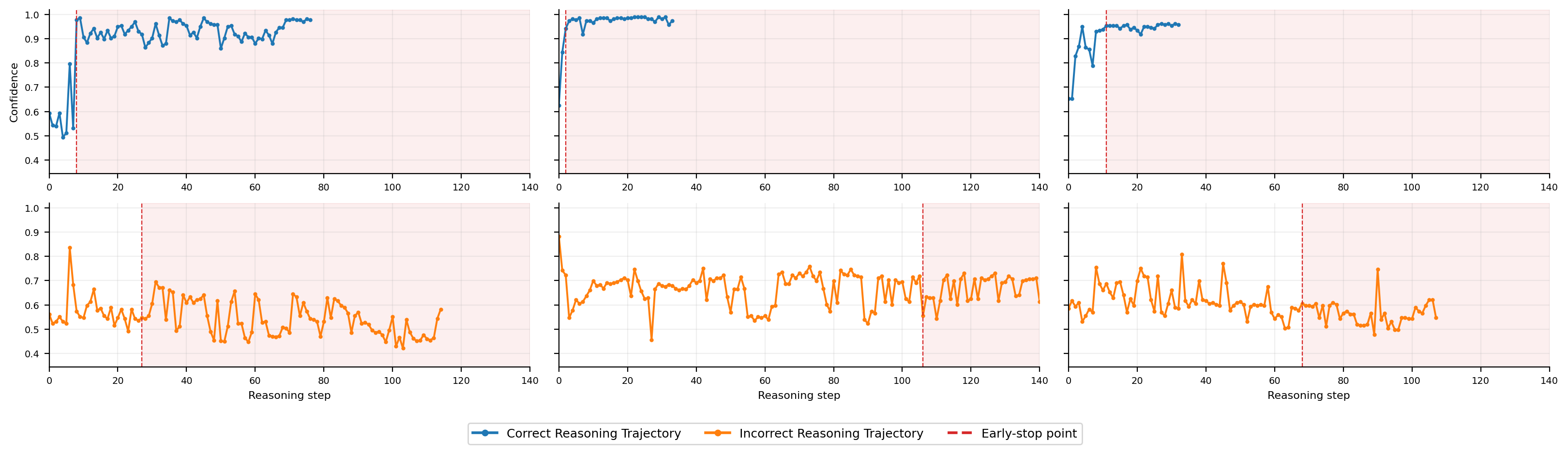}
    \caption{DeepSeek -- GSM8K}
  \end{subfigure}
  \caption{Confidence trajectories for \textbf{DeepSeek} (correct: top/blue,
  incorrect: bottom/orange; red dashed = early-stop point).}
  \label{fig:app_fig2_deepseek}
\end{figure}

\begin{figure}[h]
  \centering
  \begin{subfigure}{0.95\textwidth}
    \includegraphics[width=\linewidth]{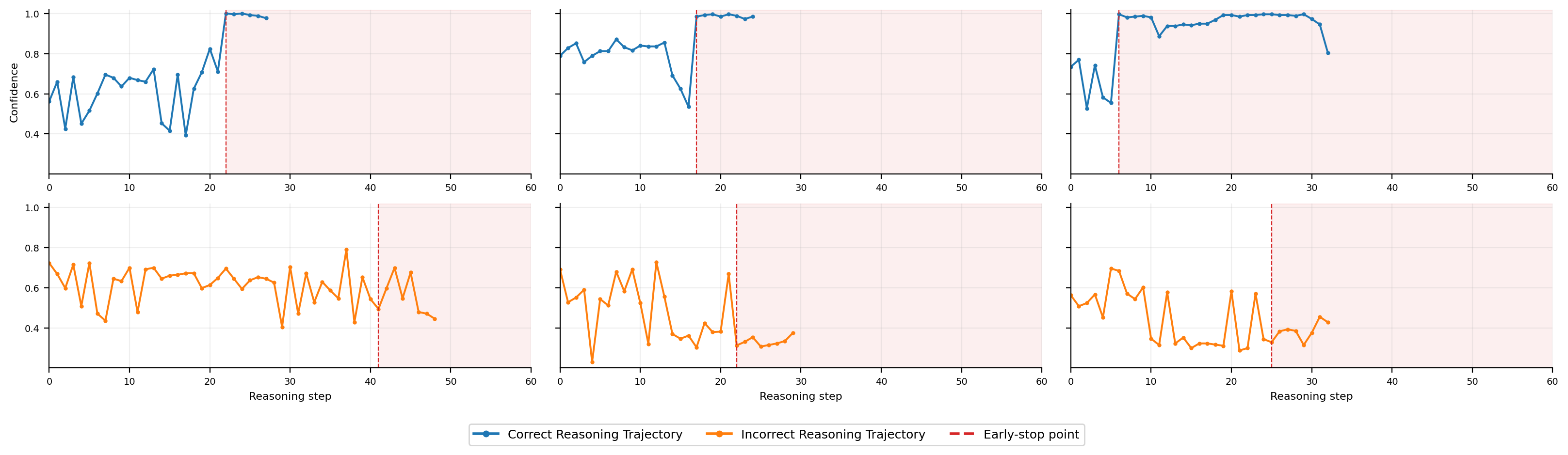}
    \caption{Nemotron -- AIME}
  \end{subfigure}\\[2pt]
  \begin{subfigure}{0.95\textwidth}
    \includegraphics[width=\linewidth]{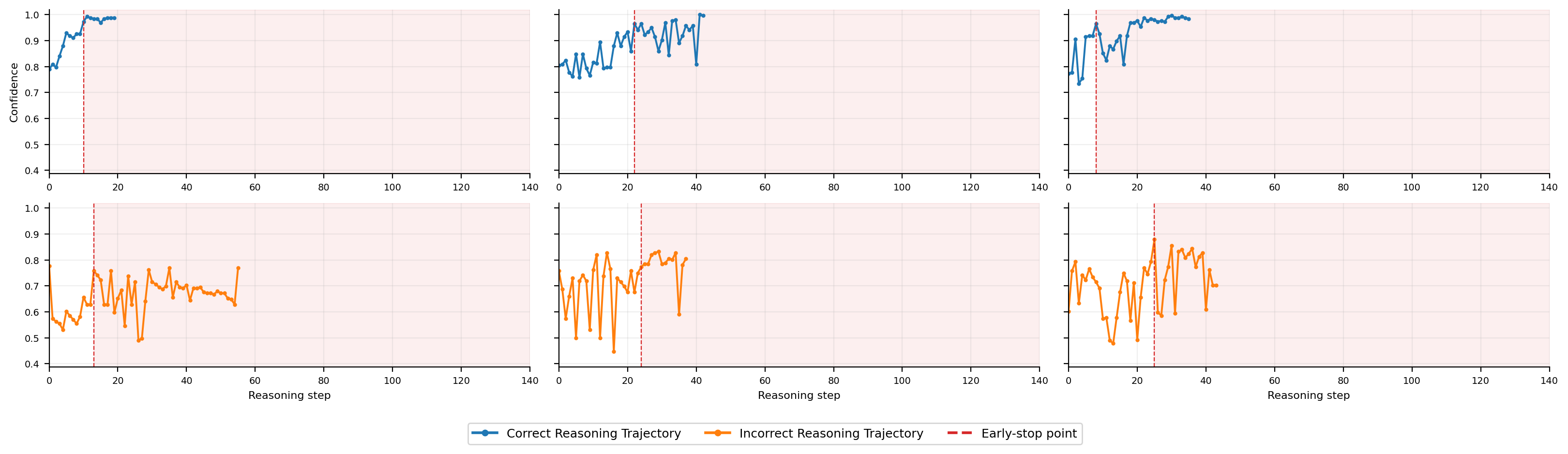}
    \caption{Nemotron -- GPQA-D}
  \end{subfigure}\\[2pt]
  \begin{subfigure}{0.95\textwidth}
    \includegraphics[width=\linewidth]{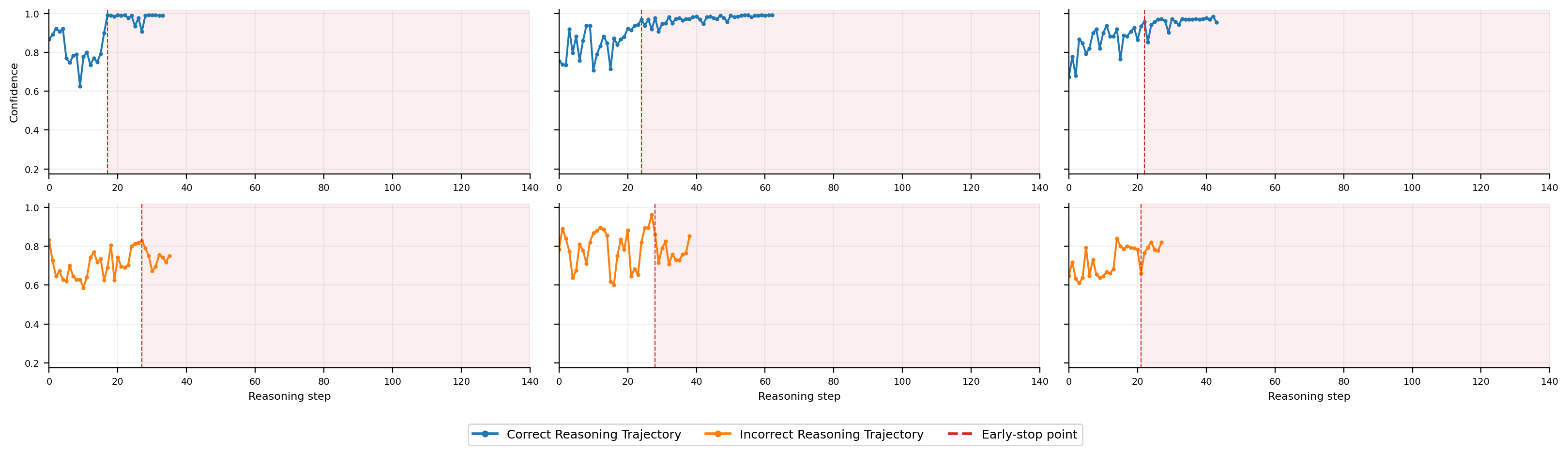}
    \caption{Nemotron -- Math500}
  \end{subfigure}\\[2pt]
  \begin{subfigure}{0.95\textwidth}
    \includegraphics[width=\linewidth]{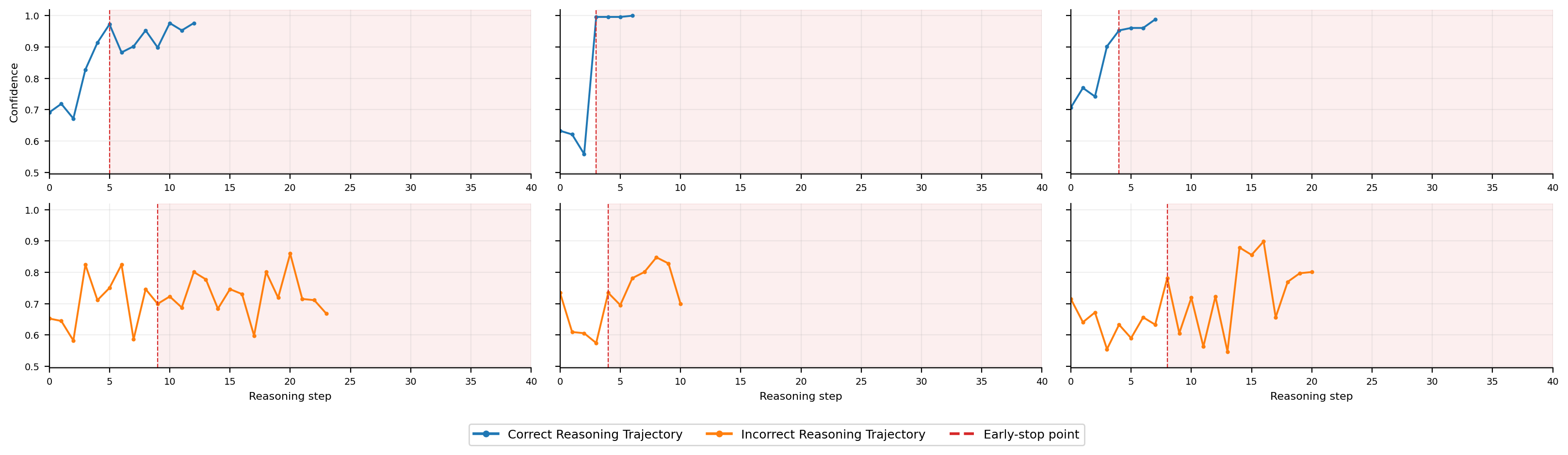}
    \caption{Nemotron -- GSM8K}
  \end{subfigure}
  \caption{Confidence trajectories for \textbf{Nemotron}. Same trend as DeepSeek:
  correct trajectories stabilize early; incorrect ones stay long and volatile.}
  \label{fig:app_fig2_nemotron}
\end{figure}

Figures~\ref{fig:app_fig2_deepseek} and~\ref{fig:app_fig2_nemotron} show the per-step confidence of representative \emph{correct} (top, blue) and \emph{incorrect} (bottom, orange) trajectories; the red dashed line marks the early-stop point and the shaded region the steps beyond it. Correct trajectories reach a high, stable confidence plateau early, whereas incorrect trajectories continue for longer and fluctuate throughout.

\clearpage
\subsection{Distribution of reasoning-step counts}
\label{ssec:app_fig3}

\begin{figure}[h]
  \centering
  \begin{subfigure}{0.48\textwidth}
    \includegraphics[width=\linewidth]{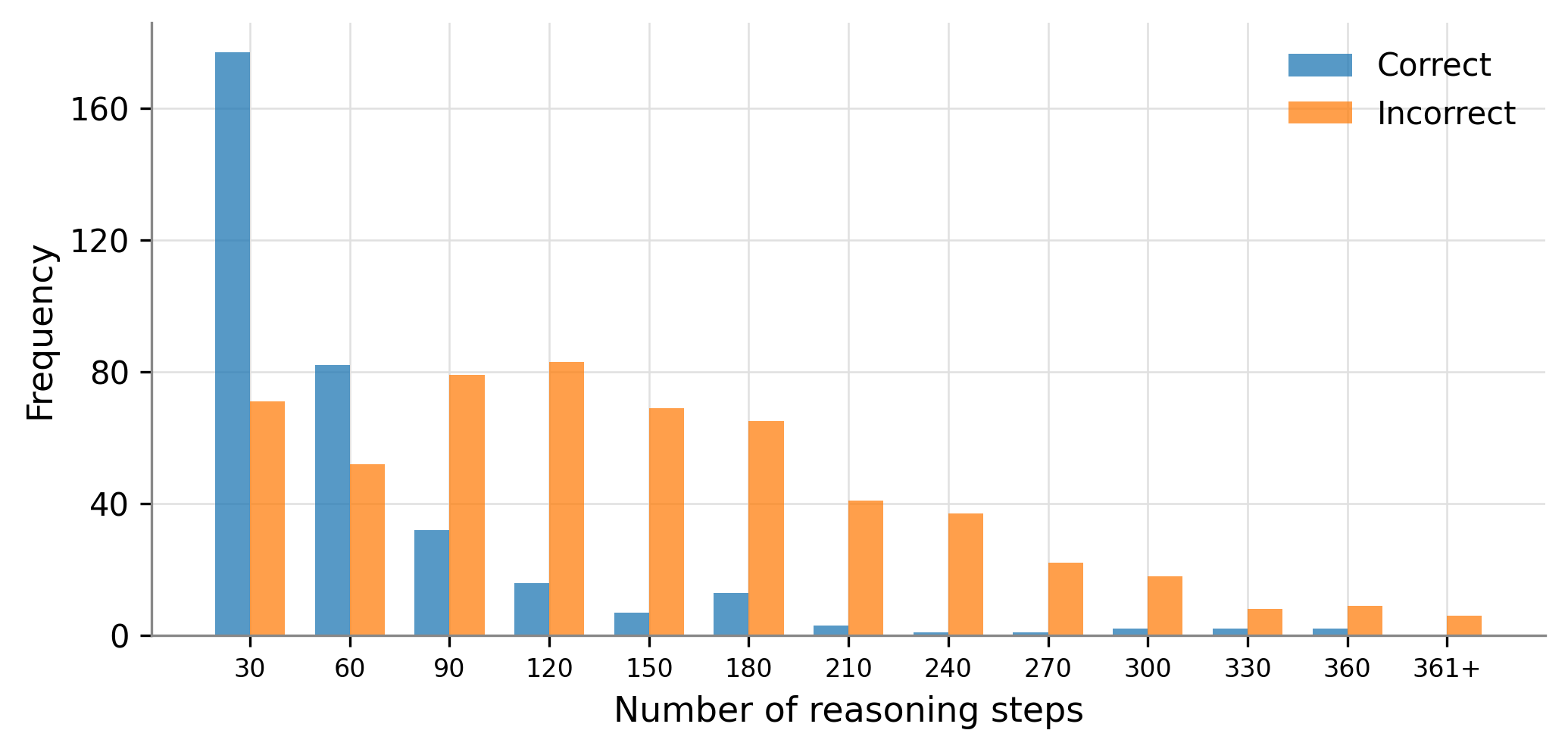}
    \caption{DeepSeek -- AIME}
  \end{subfigure}\hfill
  \begin{subfigure}{0.48\textwidth}
    \includegraphics[width=\linewidth]{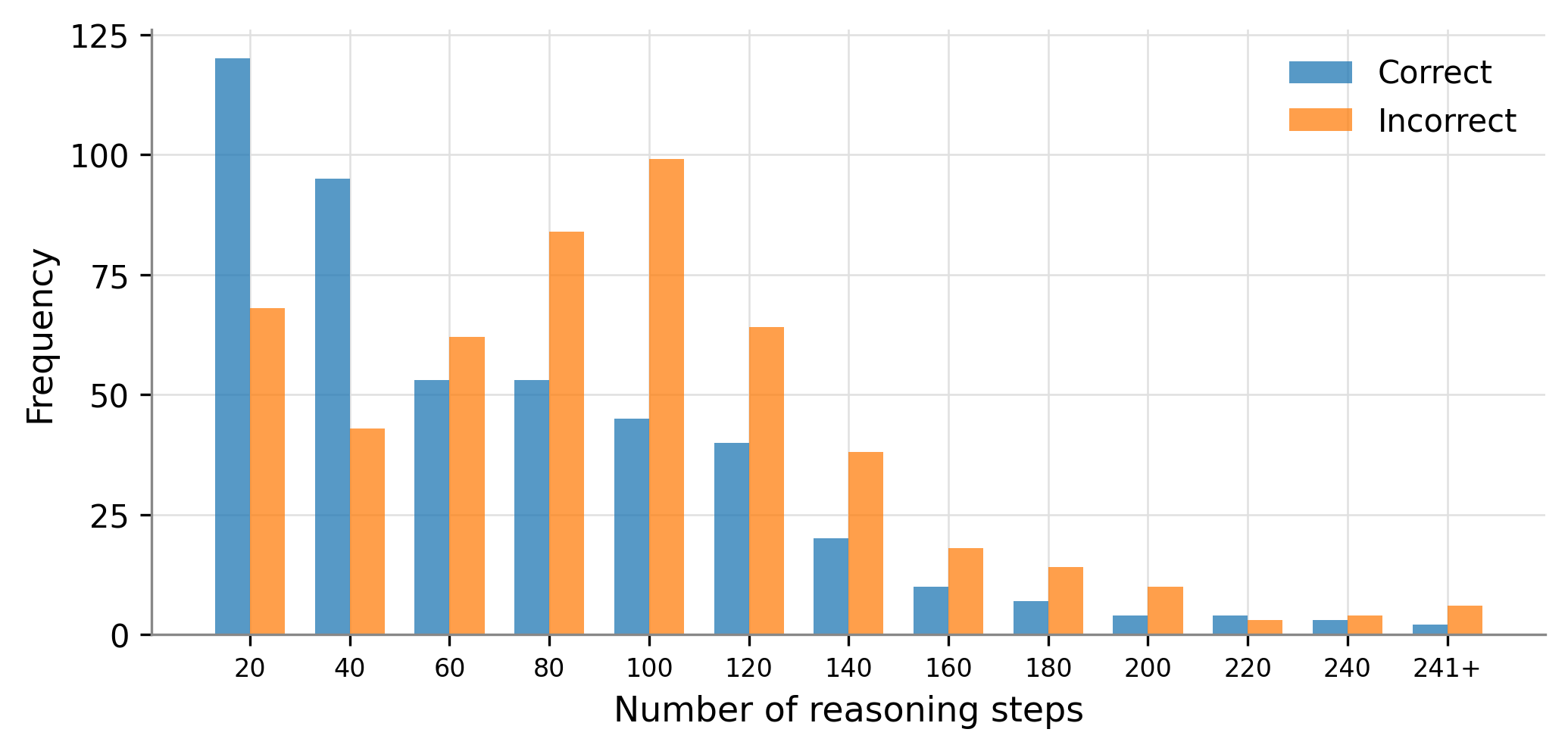}
    \caption{DeepSeek -- GPQA-D}
  \end{subfigure}\\[4pt]
  \begin{subfigure}{0.48\textwidth}
    \includegraphics[width=\linewidth]{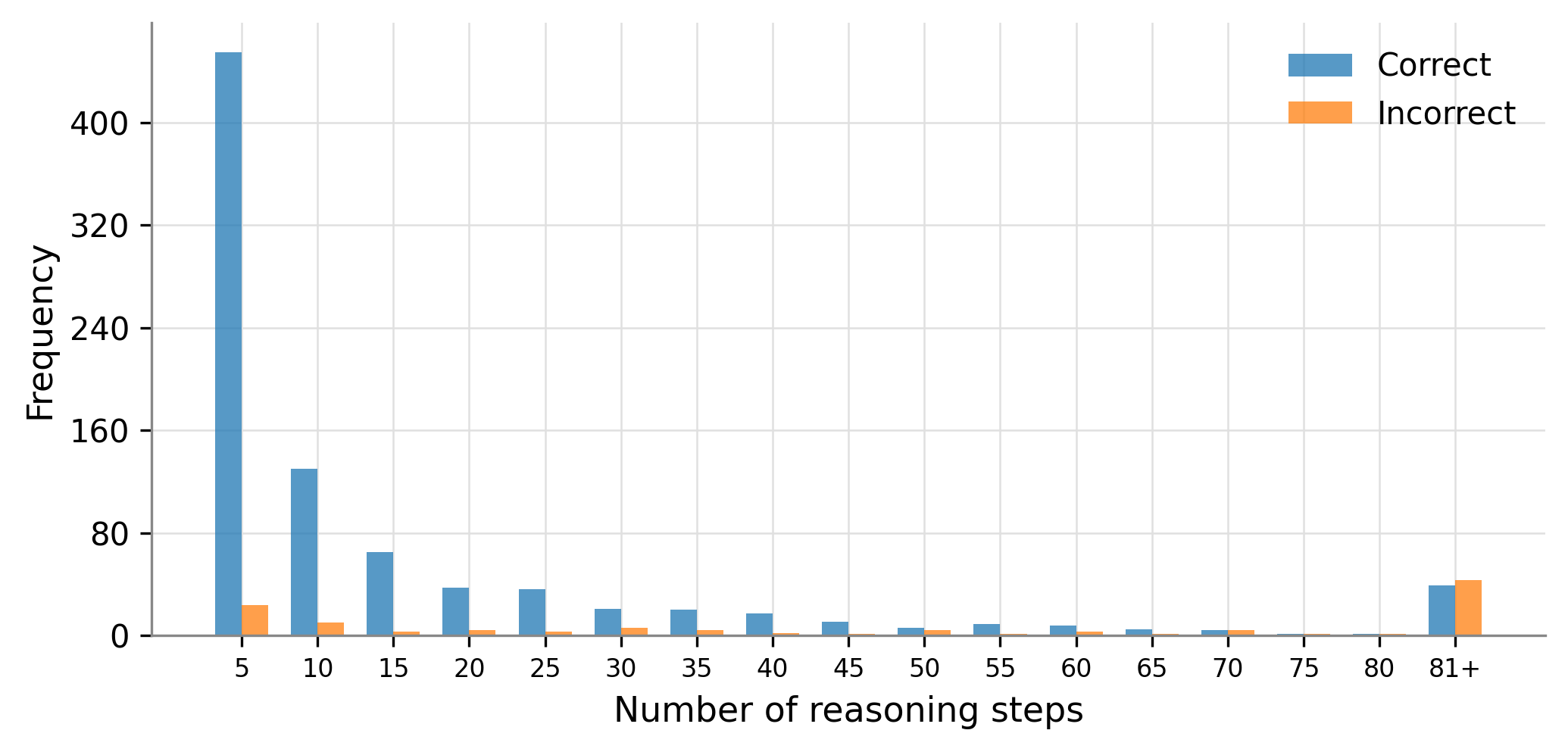}
    \caption{DeepSeek -- Math500}
  \end{subfigure}\hfill
  \begin{subfigure}{0.48\textwidth}
    \includegraphics[width=\linewidth]{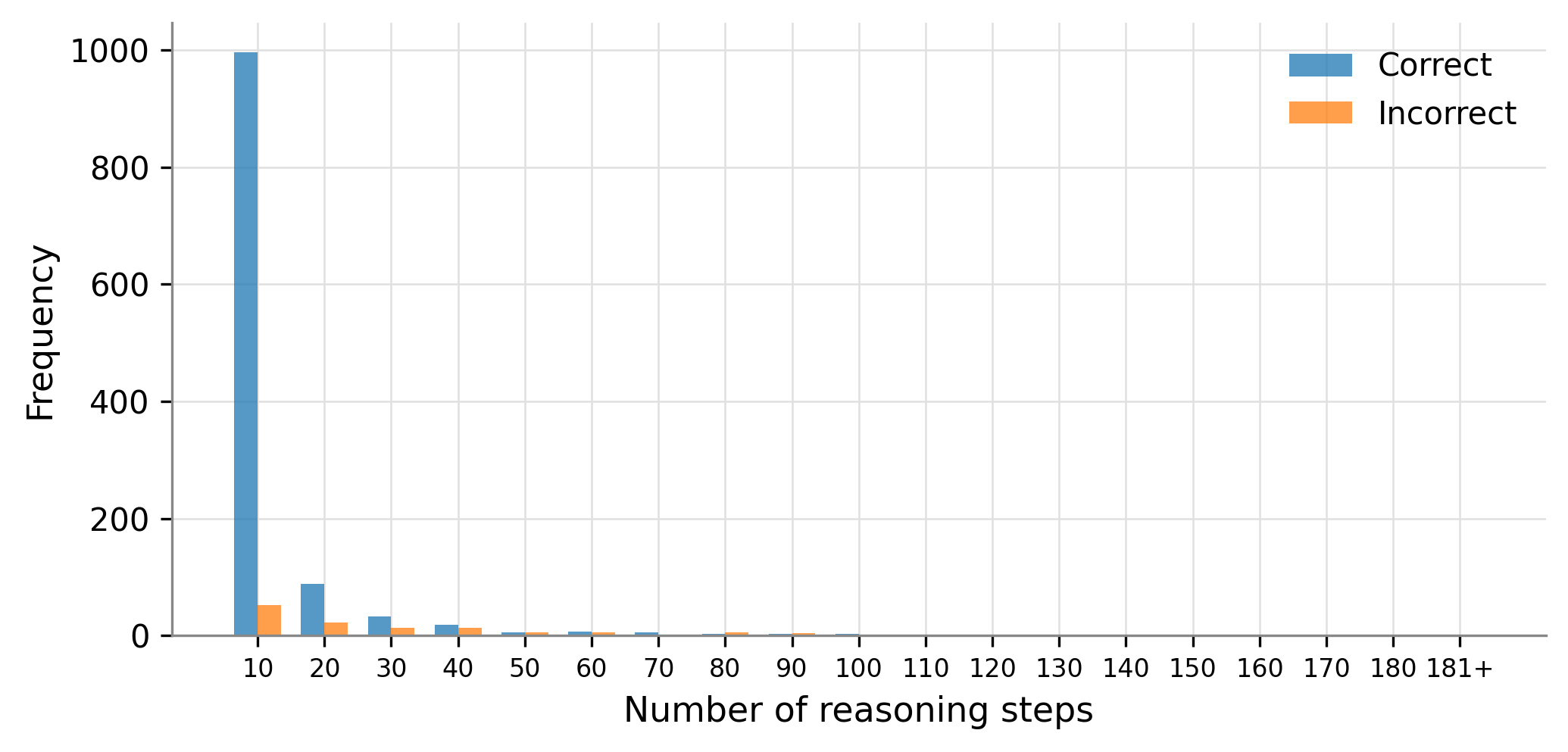}
    \caption{DeepSeek -- GSM8K}
  \end{subfigure}\\[4pt]
  \begin{subfigure}{0.48\textwidth}
    \includegraphics[width=\linewidth]{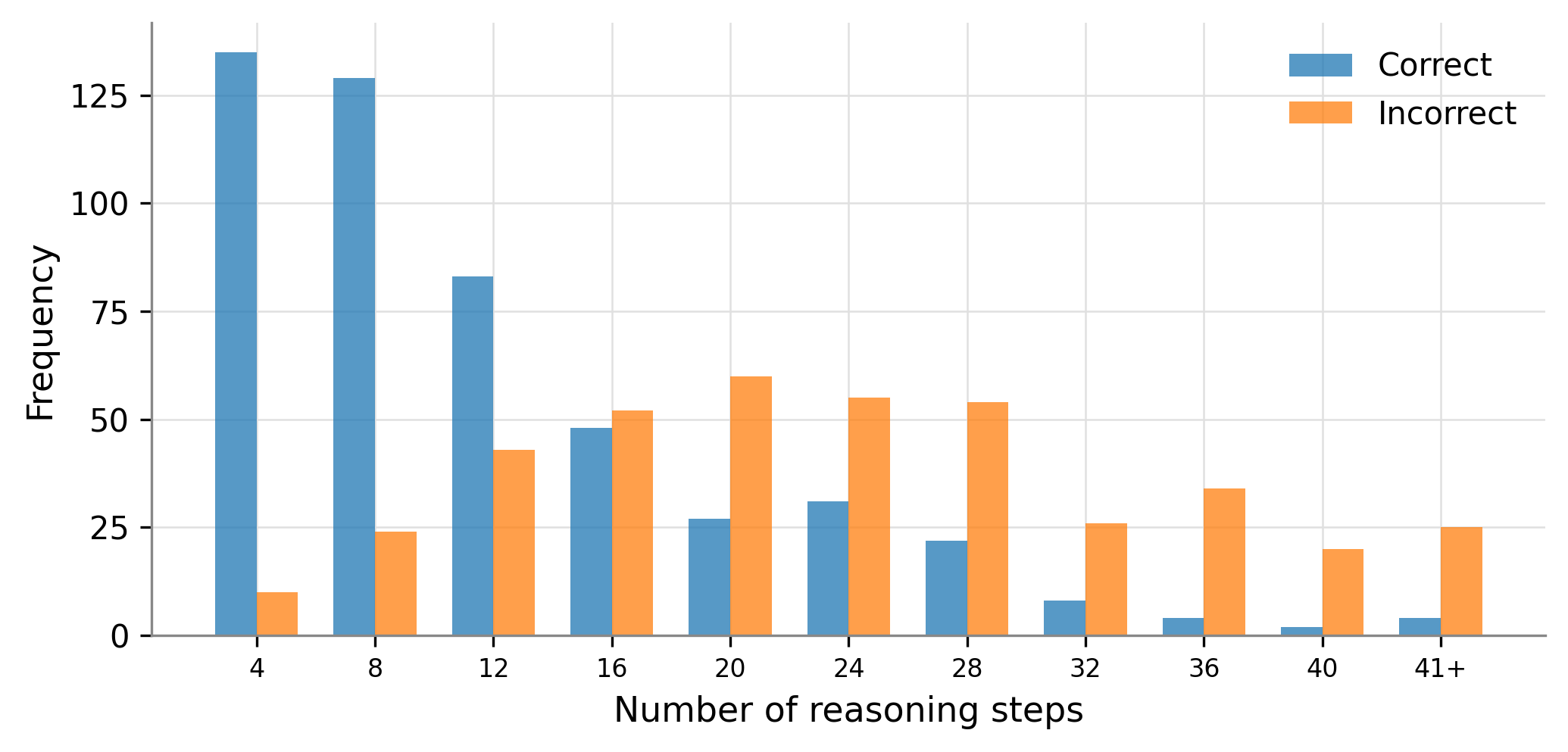}
    \caption{Nemotron -- AIME}
  \end{subfigure}\hfill
  \begin{subfigure}{0.48\textwidth}
    \includegraphics[width=\linewidth]{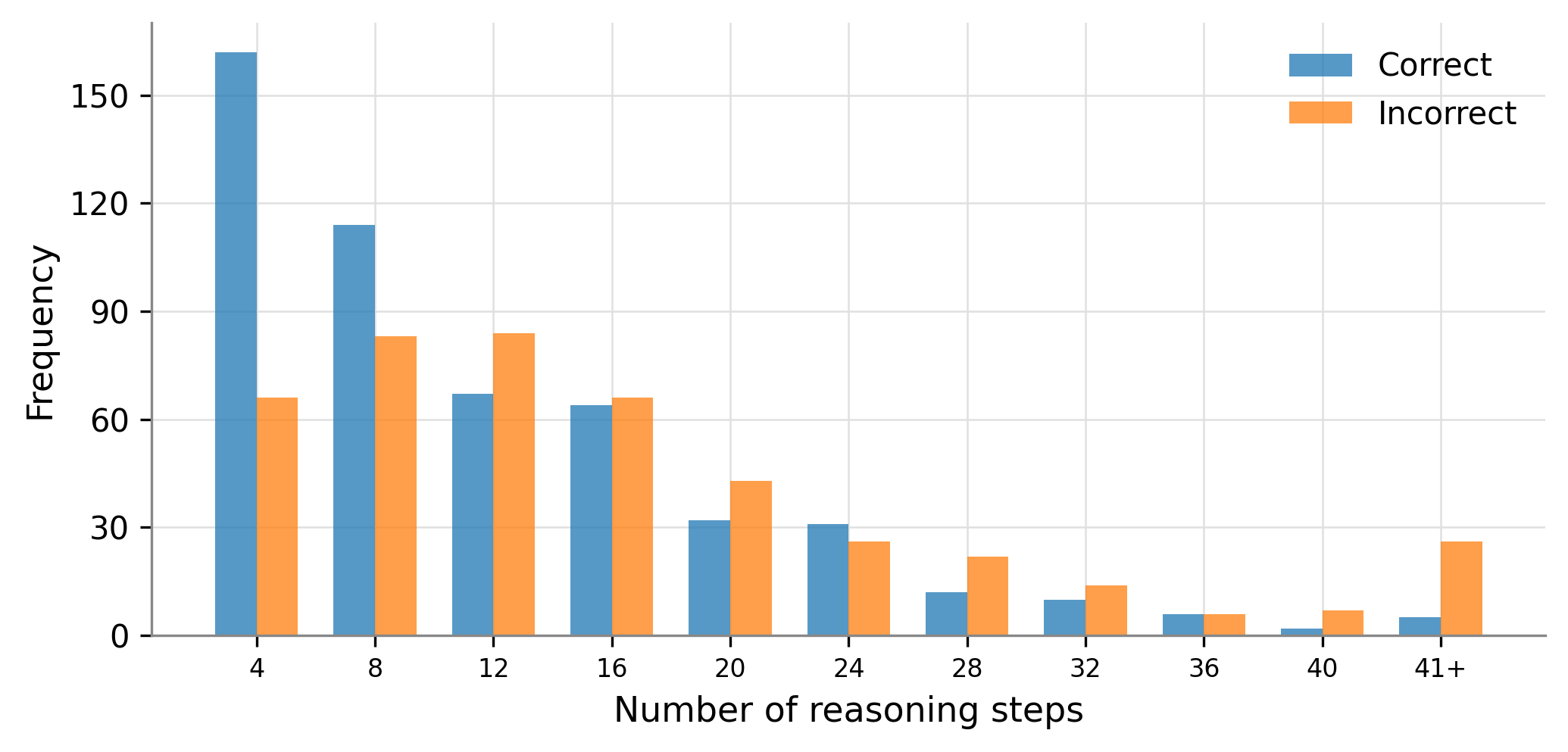}
    \caption{Nemotron -- GPQA-D}
  \end{subfigure}\\[4pt]
  \begin{subfigure}{0.48\textwidth}
    \includegraphics[width=\linewidth]{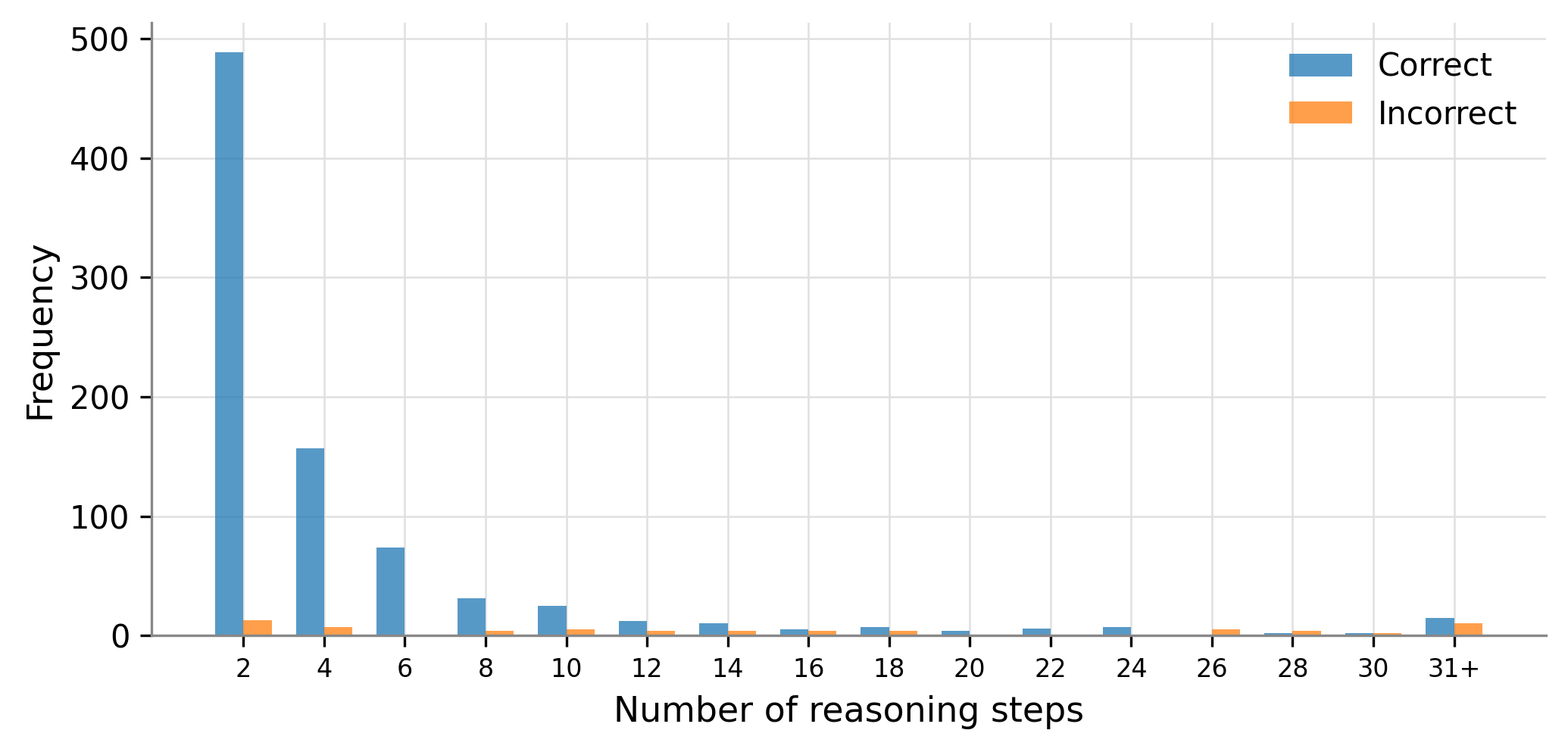}
    \caption{Nemotron -- Math500}
  \end{subfigure}\hfill
  \begin{subfigure}{0.48\textwidth}
    \includegraphics[width=\linewidth]{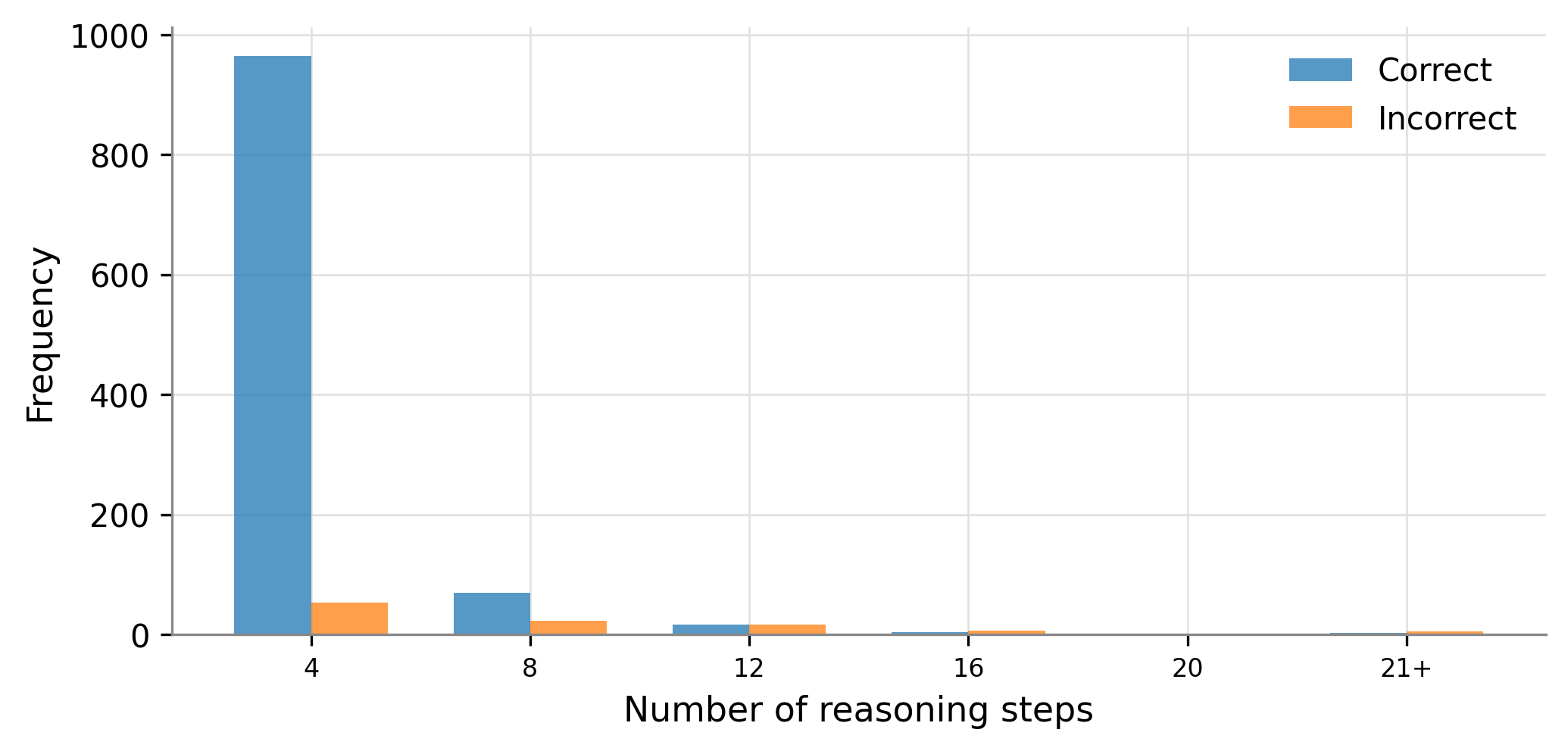}
    \caption{Nemotron -- GSM8K}
  \end{subfigure}
  \caption{Reasoning-step count distributions. Correct trajectories (blue) are
  consistently shorter than incorrect ones (orange) across all settings; GSM8K
  is the shortest, AIME the longest.}
  \label{fig:app_fig3}
\end{figure}

Figure~\ref{fig:app_fig3} shows histograms of the number of reasoning steps for correct (blue) vs.\ incorrect (orange) trajectories. Correct answers concentrate at low step counts, while incorrect answers exhibit a heavy tail toward much longer trajectories.

\clearpage
\subsection{Cumulative degeneration score $D_K$}
\label{ssec:app_fig4a}

\begin{figure}[h]
  \centering
  \begin{subfigure}{0.48\textwidth}
    \includegraphics[width=\linewidth]{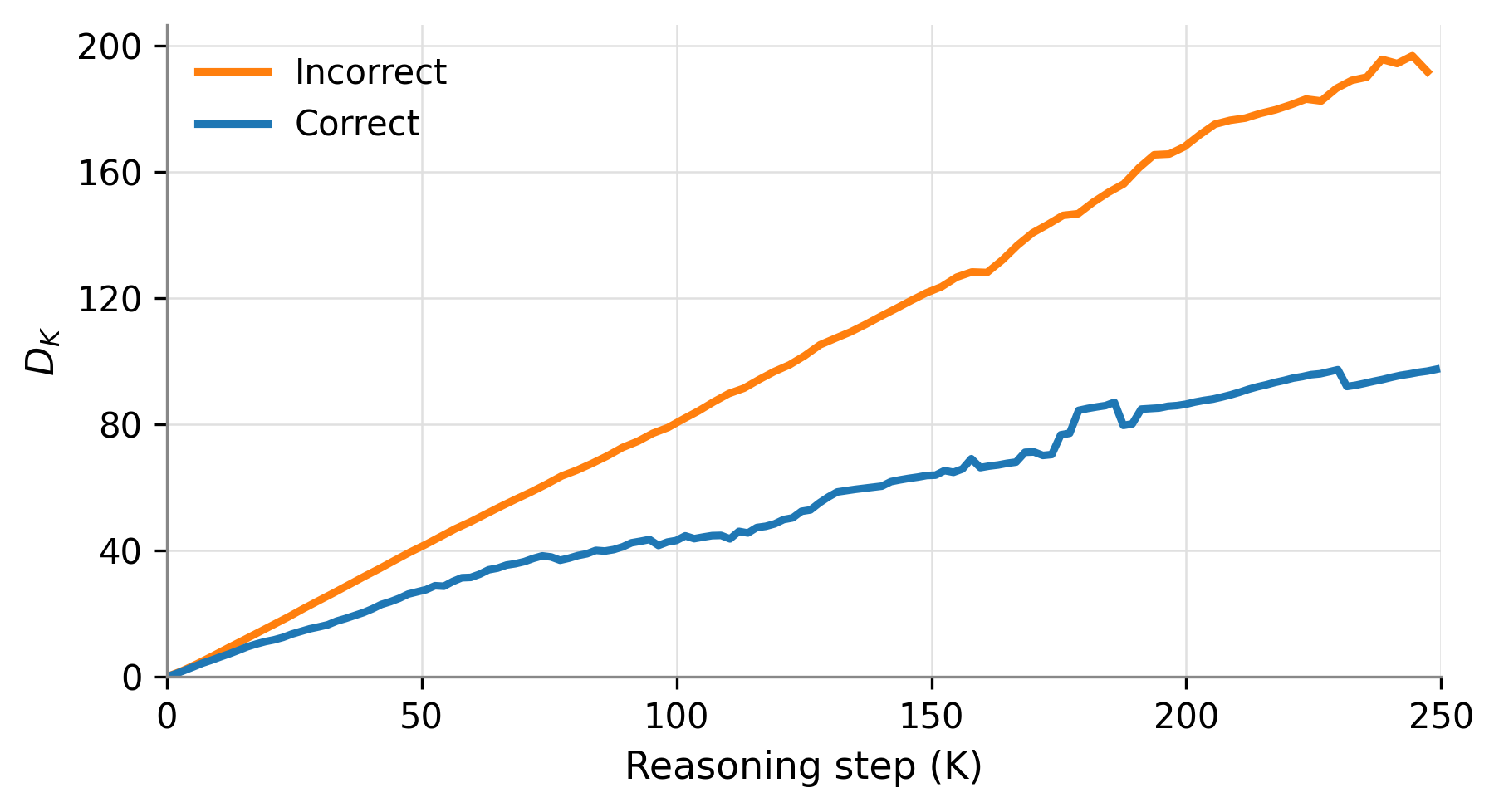}
    \caption{DeepSeek -- AIME}
  \end{subfigure}\hfill
  \begin{subfigure}{0.48\textwidth}
    \includegraphics[width=\linewidth]{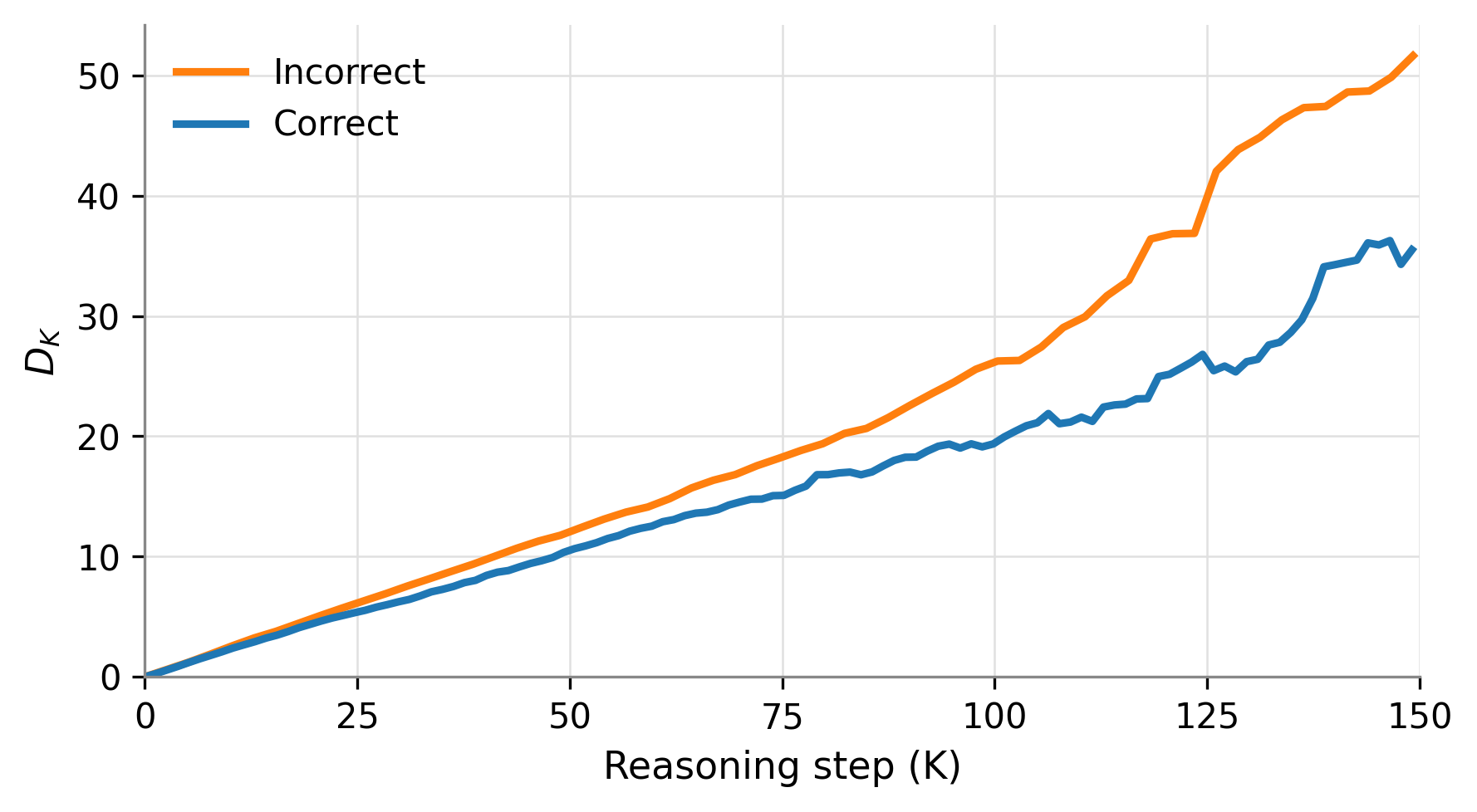}
    \caption{DeepSeek -- GPQA-D}
  \end{subfigure}\\[4pt]
  \begin{subfigure}{0.48\textwidth}
    \includegraphics[width=\linewidth]{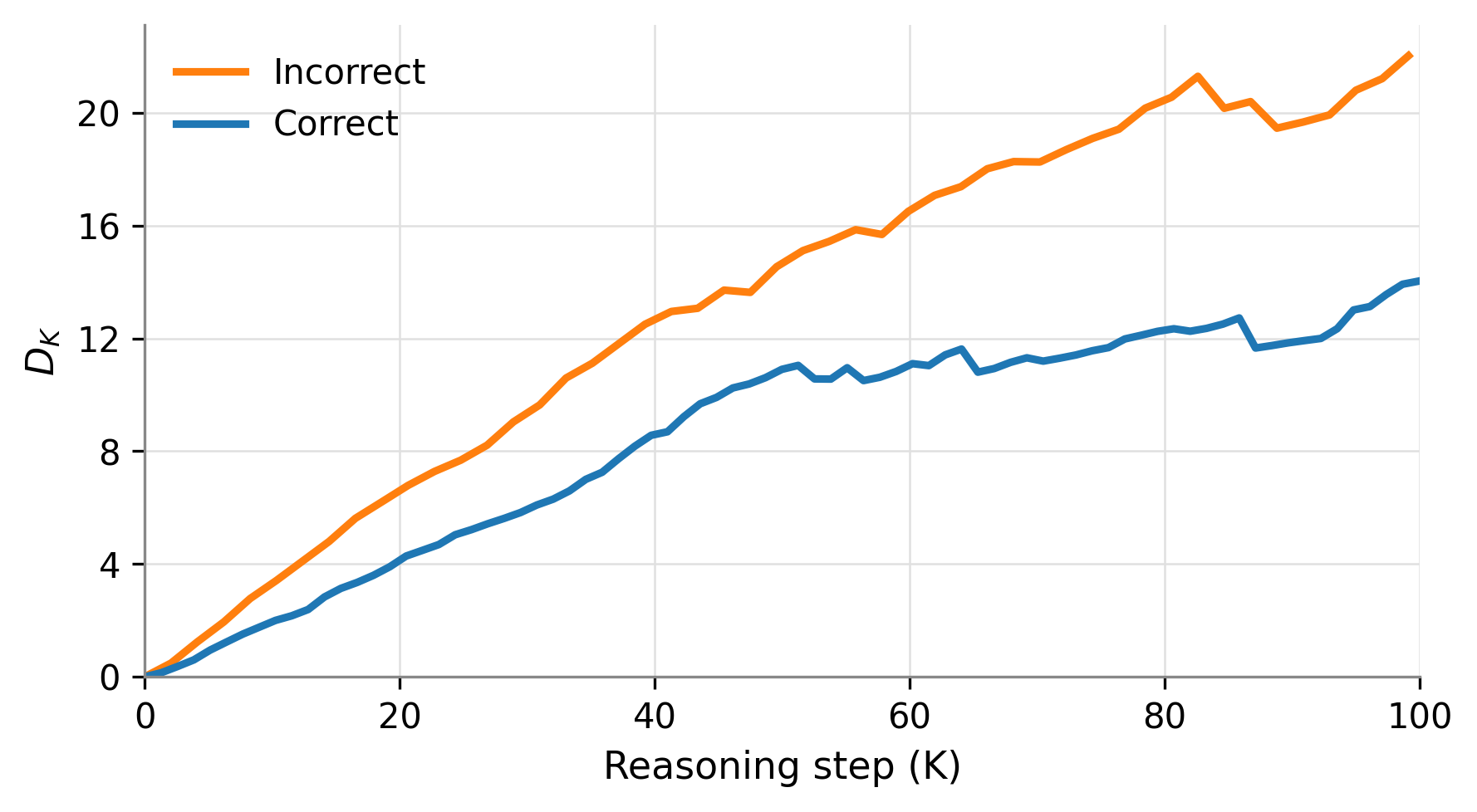}
    \caption{DeepSeek -- Math500}
  \end{subfigure}\hfill
  \begin{subfigure}{0.48\textwidth}
    \includegraphics[width=\linewidth]{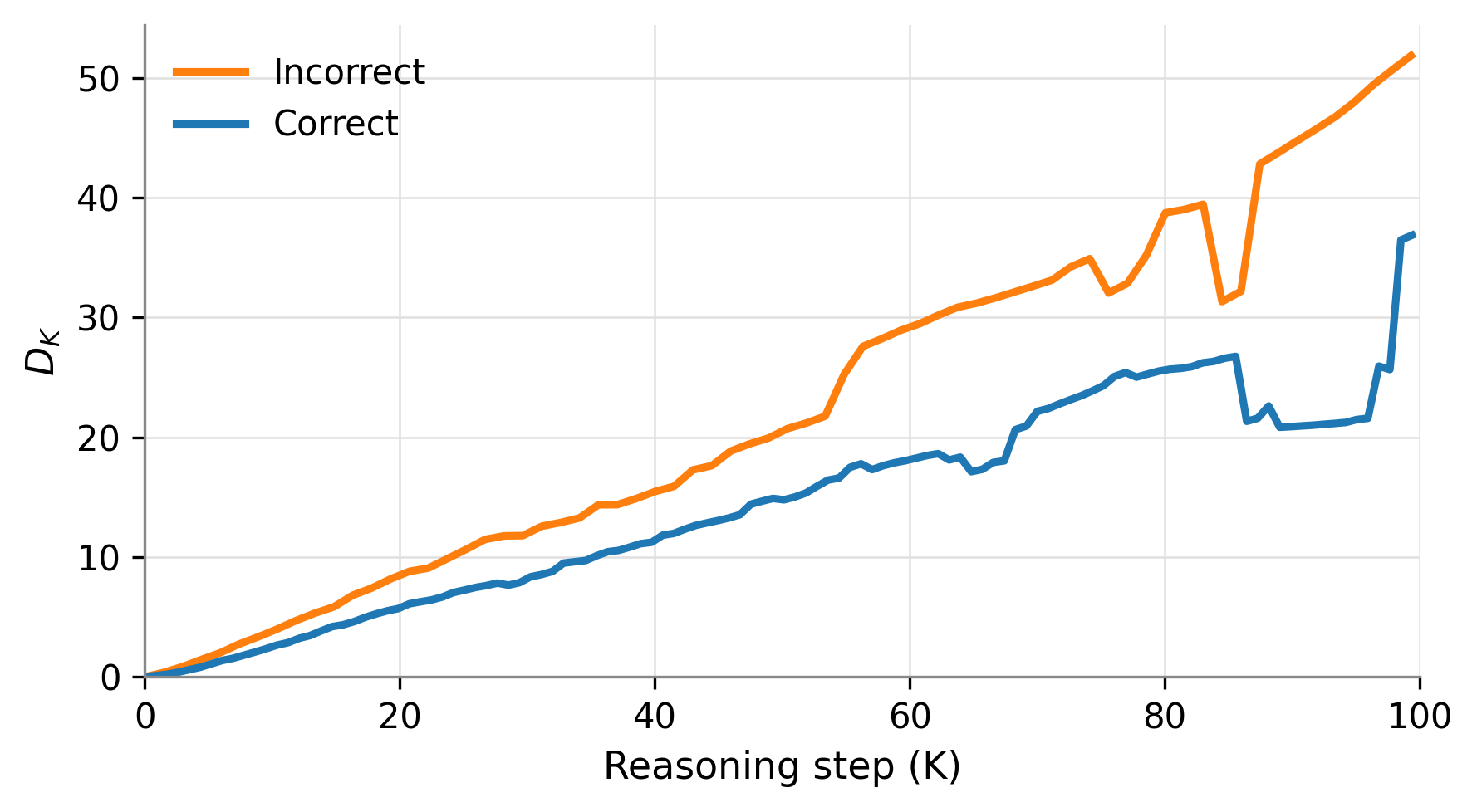}
    \caption{DeepSeek -- GSM8K}
  \end{subfigure}\\[4pt]
  \begin{subfigure}{0.48\textwidth}
    \includegraphics[width=\linewidth]{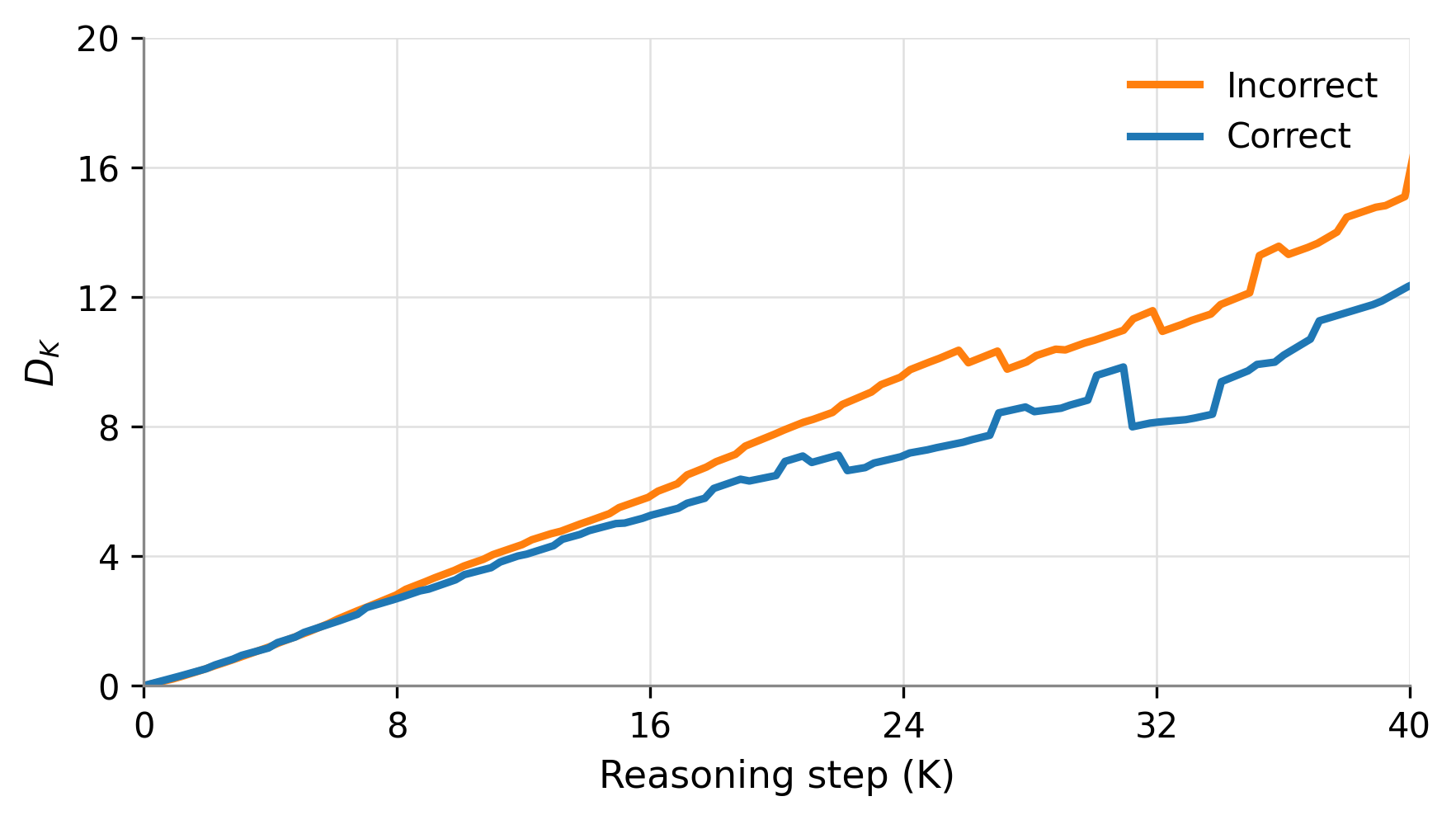}
    \caption{Nemotron -- AIME}
  \end{subfigure}\hfill
  \begin{subfigure}{0.48\textwidth}
    \includegraphics[width=\linewidth]{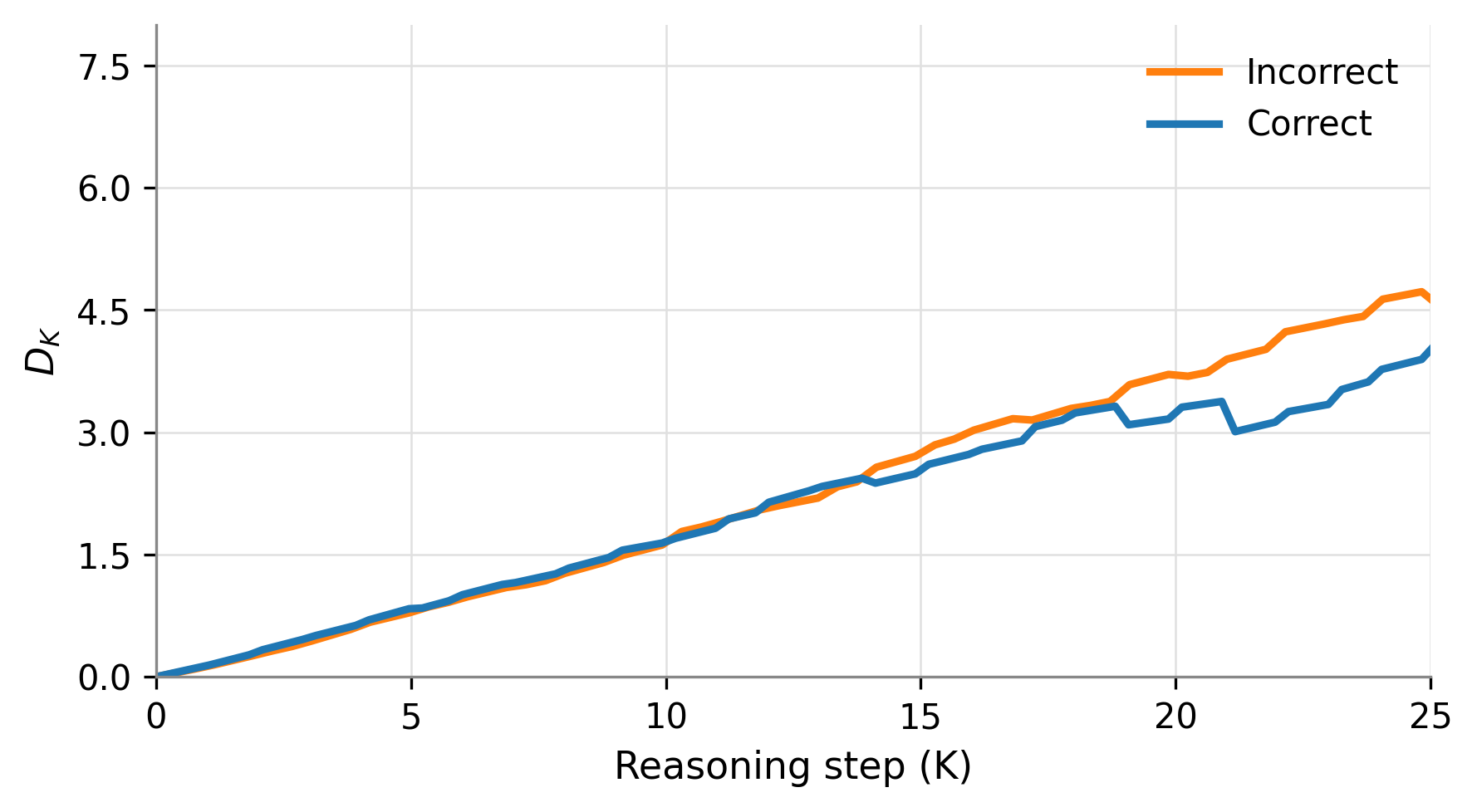}
    \caption{Nemotron -- GPQA-D}
  \end{subfigure}\\[4pt]
  \begin{subfigure}{0.48\textwidth}
    \includegraphics[width=\linewidth]{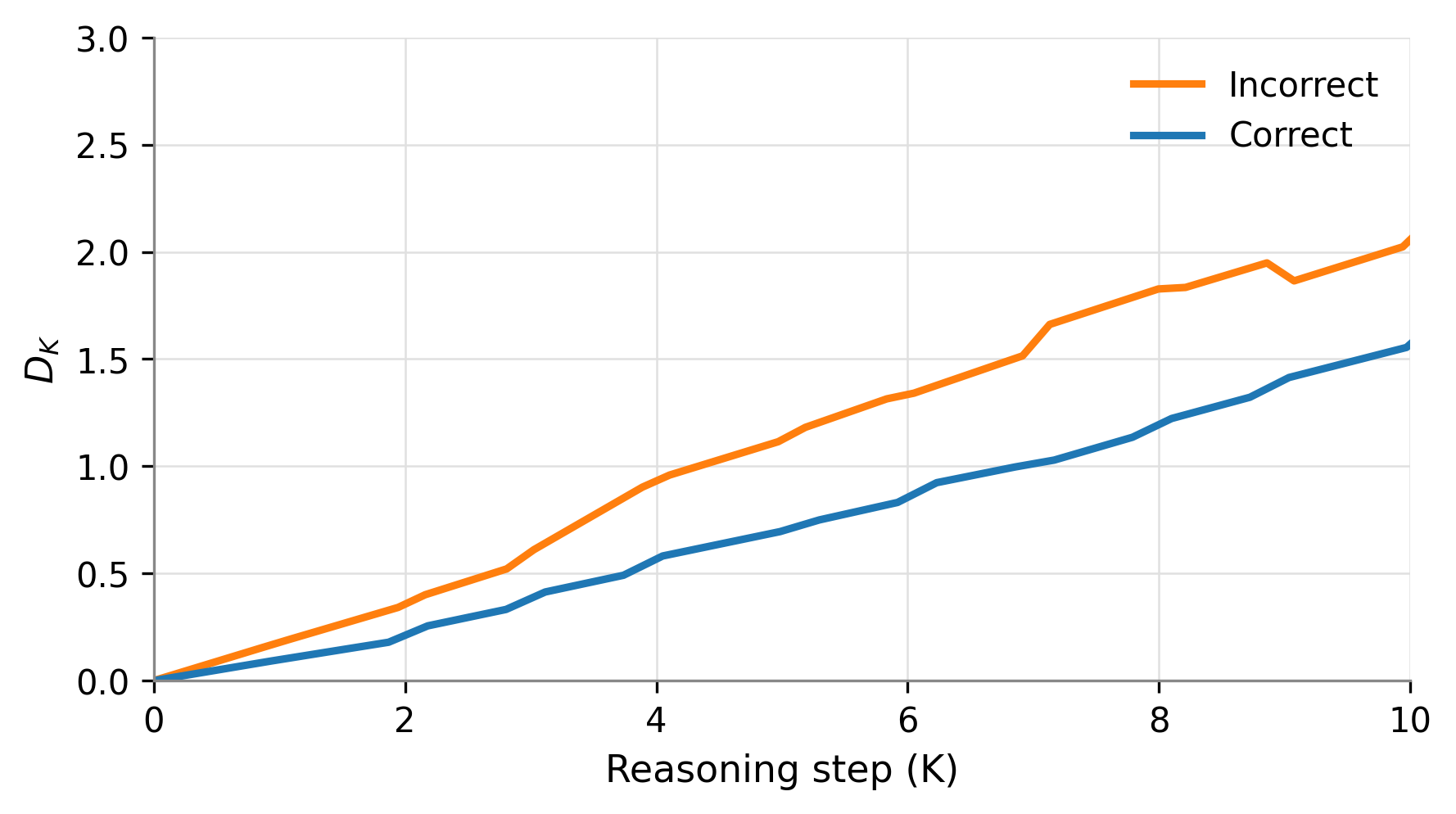}
    \caption{Nemotron -- Math500}
  \end{subfigure}\hfill
  \begin{subfigure}{0.48\textwidth}
    \includegraphics[width=\linewidth]{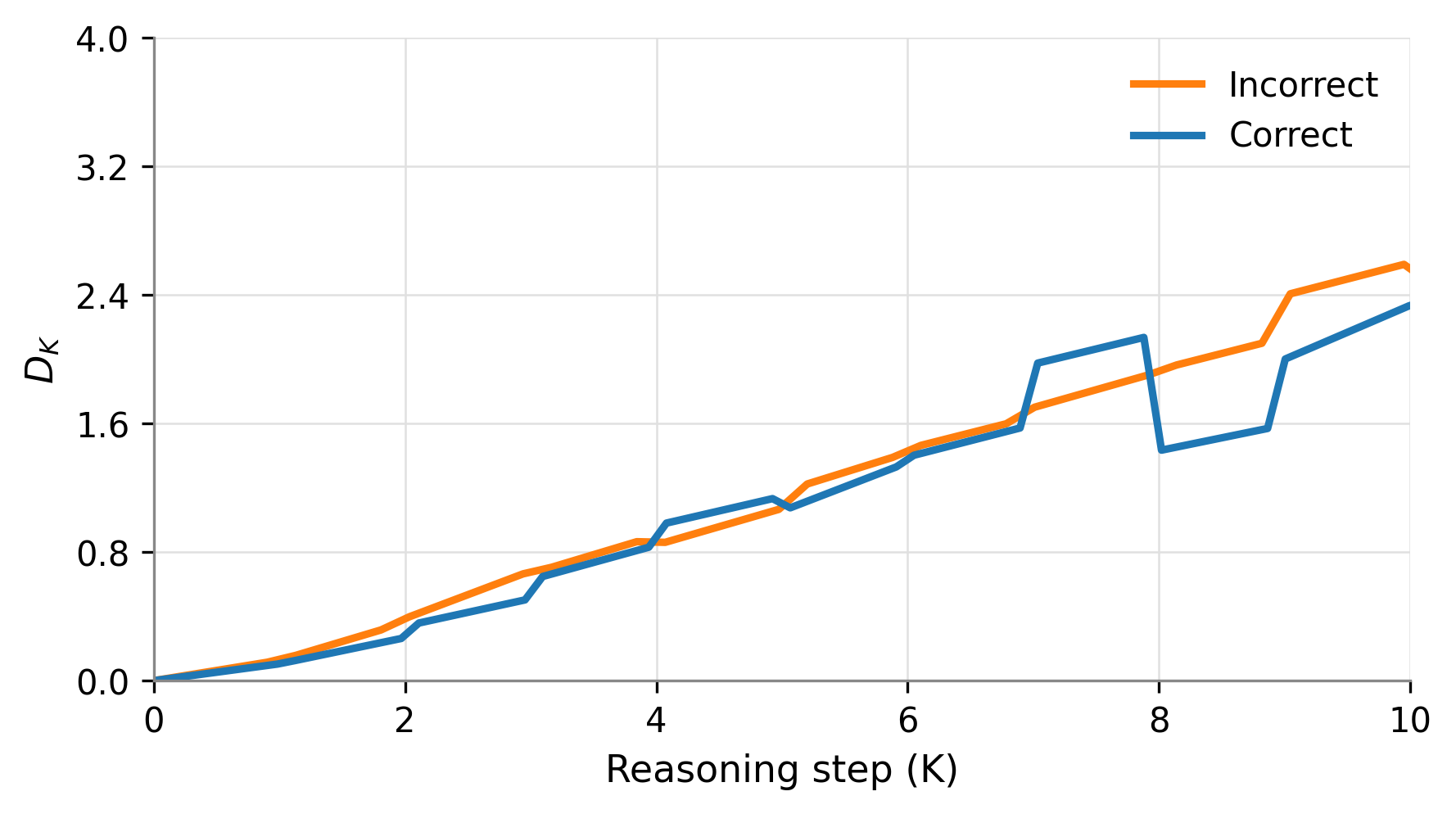}
    \caption{Nemotron -- GSM8K}
  \end{subfigure}
  \caption{Average degeneration score $D_K$ vs.\ reasoning step. Incorrect
  trajectories (orange) degenerate faster than correct ones (blue) in every
  model/benchmark combination.}
  \label{fig:app_fig4a}
\end{figure}

Figure~\ref{fig:app_fig4a} shows the average cumulative degeneration score $D_K$ as a function of the reasoning step $K$. Incorrect trajectories (orange) accumulate degeneration faster and reach substantially higher values than correct trajectories (blue), motivating the degeneration score.

\clearpage
\subsection{Average confidence dynamics}
\label{ssec:app_fig4b}

\begin{figure}[h]
  \centering
  \begin{subfigure}{0.48\textwidth}
    \includegraphics[width=\linewidth]{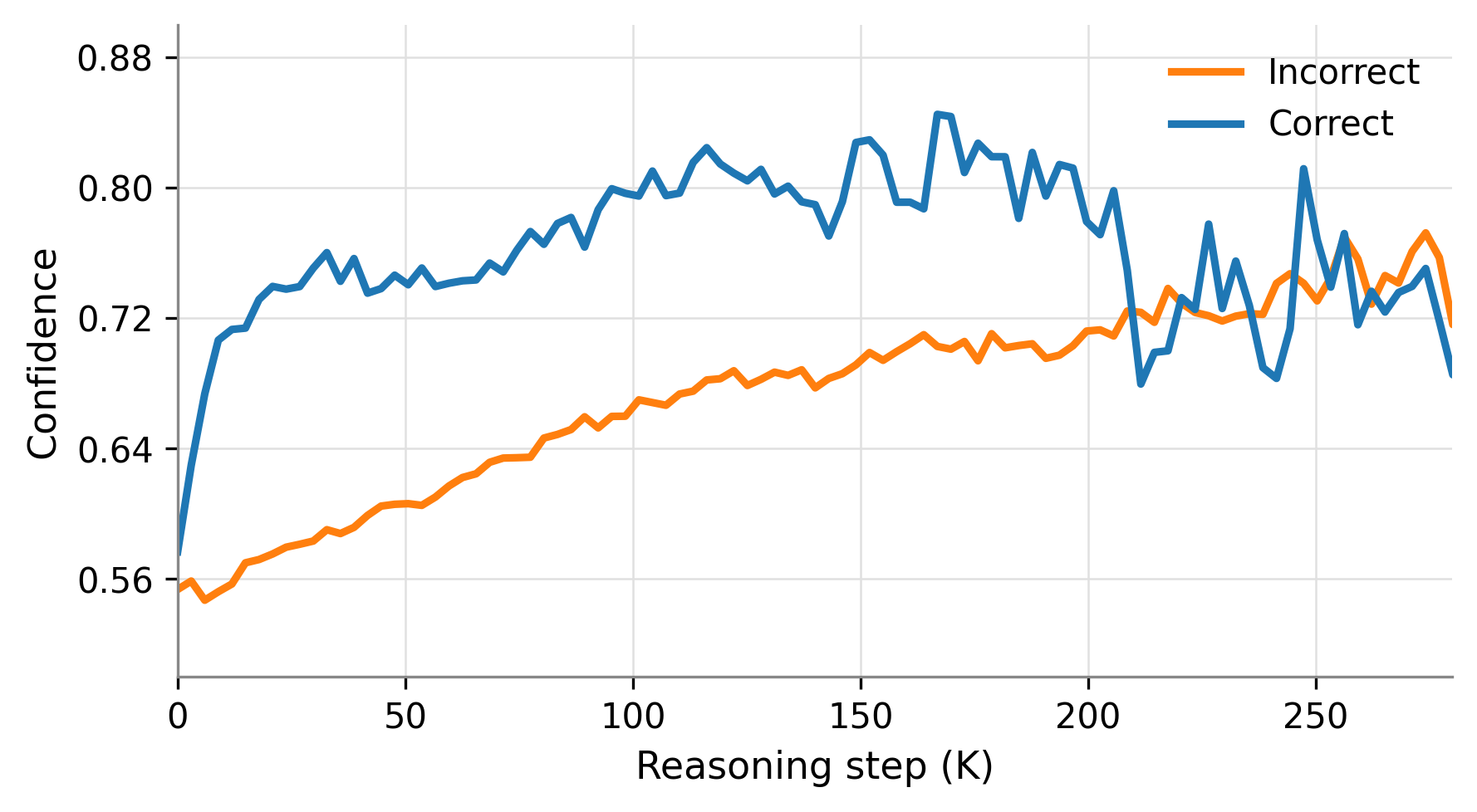}
    \caption{DeepSeek -- AIME}
  \end{subfigure}\hfill
  \begin{subfigure}{0.48\textwidth}
    \includegraphics[width=\linewidth]{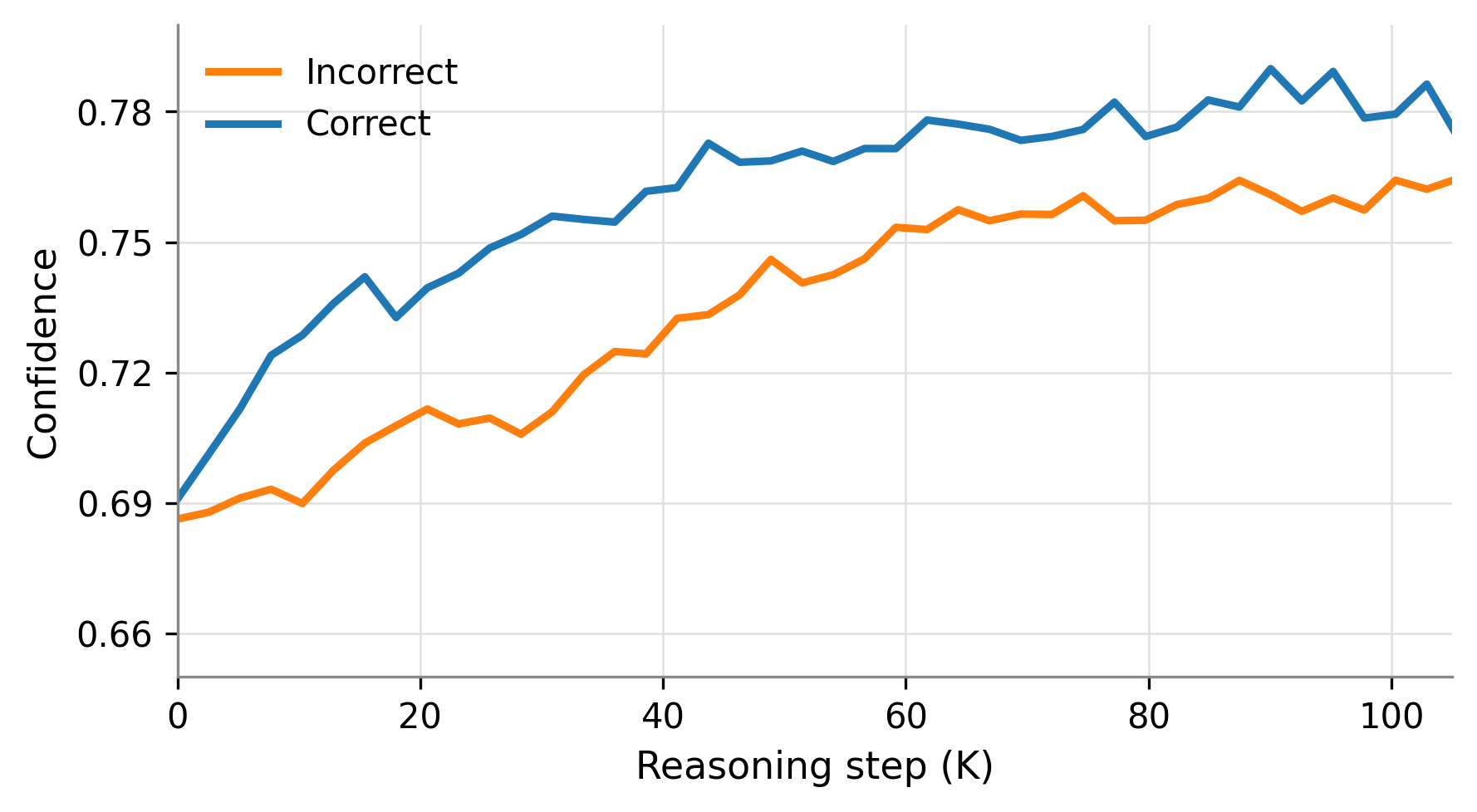}
    \caption{DeepSeek -- GPQA-D}
  \end{subfigure}\\[4pt]
  \begin{subfigure}{0.48\textwidth}
    \includegraphics[width=\linewidth]{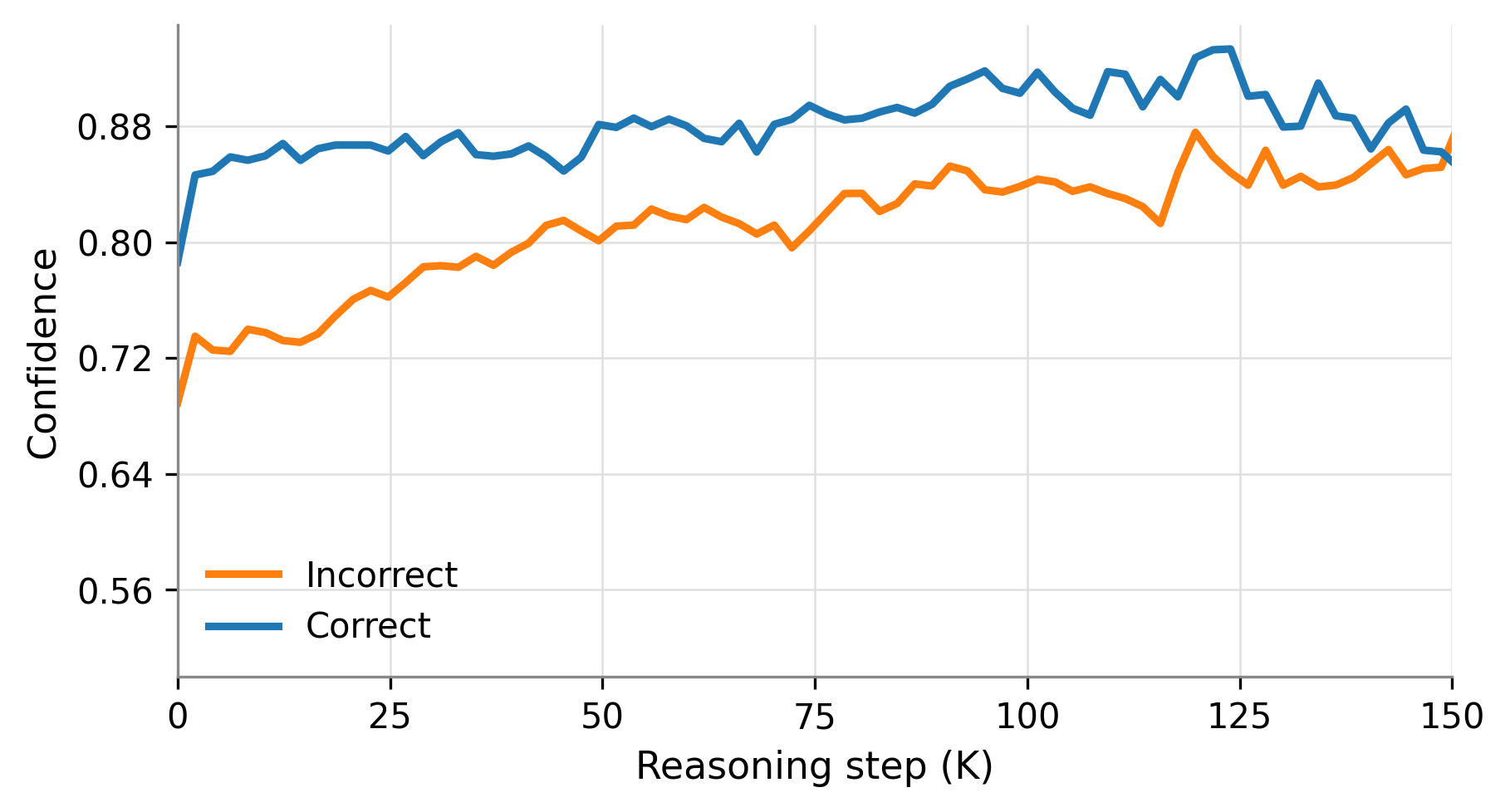}
    \caption{DeepSeek -- Math500}
  \end{subfigure}\hfill
  \begin{subfigure}{0.48\textwidth}
    \includegraphics[width=\linewidth]{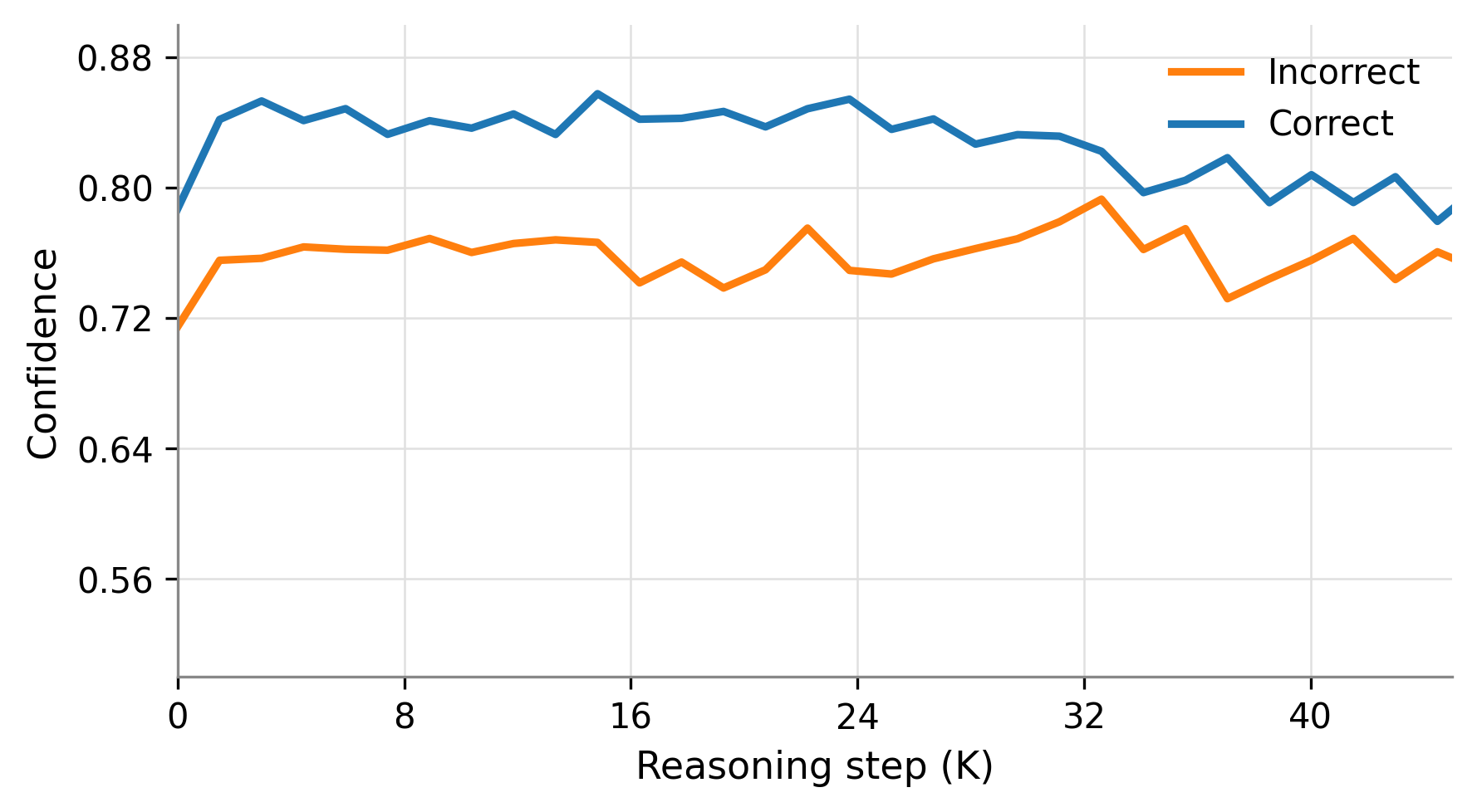}
    \caption{DeepSeek -- GSM8K}
  \end{subfigure}\\[4pt]
  \begin{subfigure}{0.48\textwidth}
    \includegraphics[width=\linewidth]{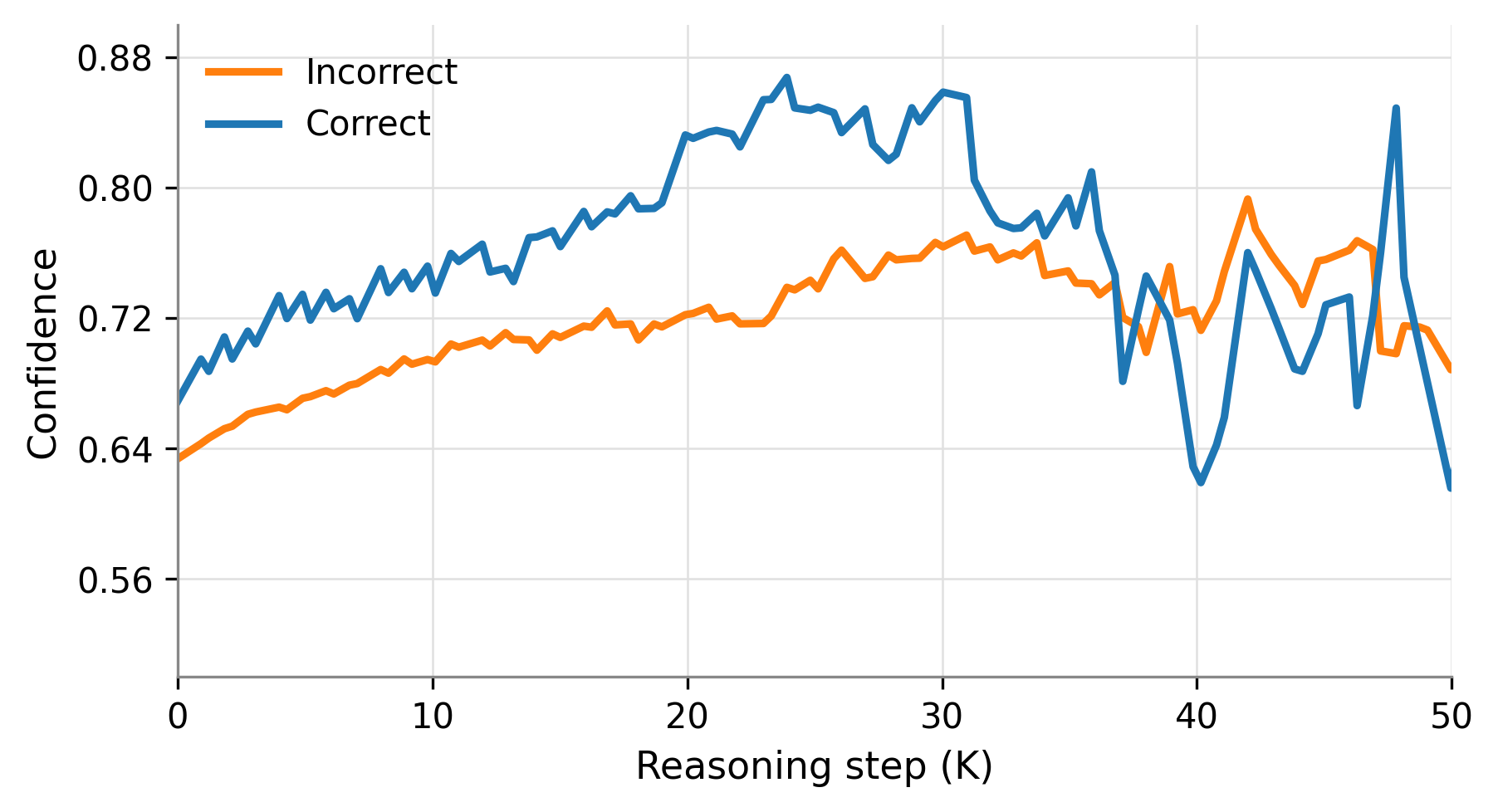}
    \caption{Nemotron -- AIME}
  \end{subfigure}\hfill
  \begin{subfigure}{0.48\textwidth}
    \includegraphics[width=\linewidth]{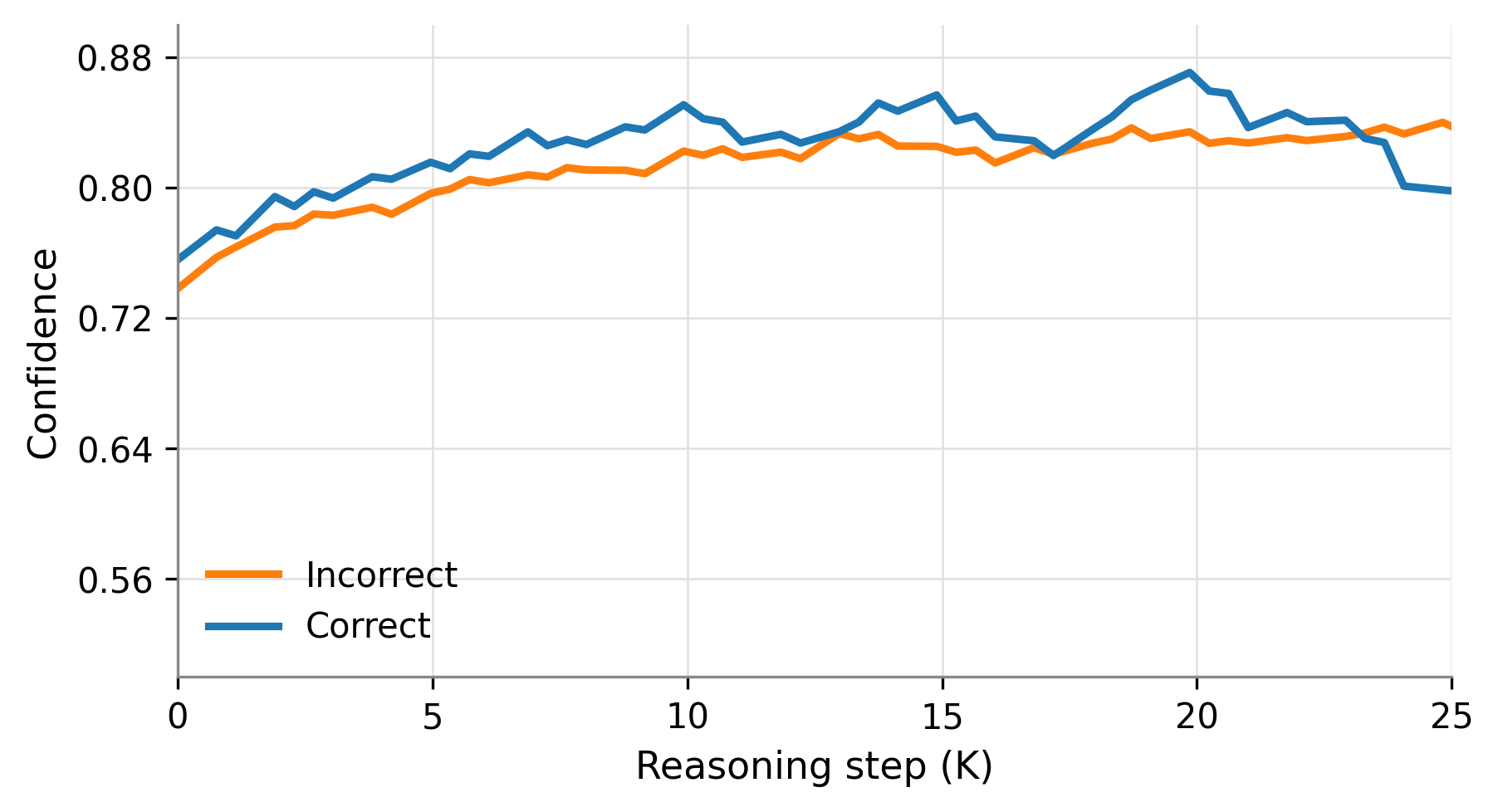}
    \caption{Nemotron -- GPQA-D}
  \end{subfigure}\\[4pt]
  \begin{subfigure}{0.48\textwidth}
    \includegraphics[width=\linewidth]{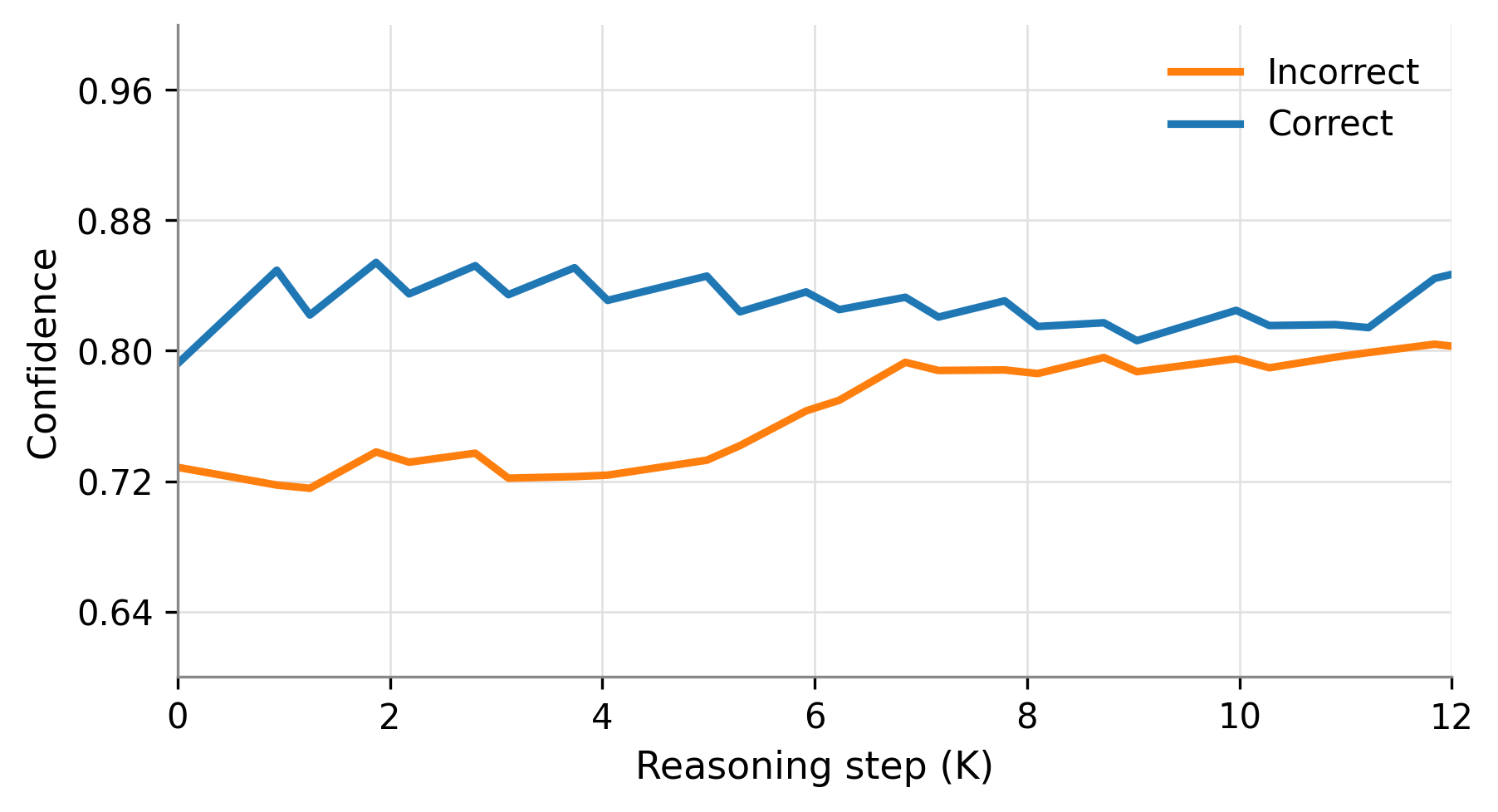}
    \caption{Nemotron -- Math500}
  \end{subfigure}\hfill
  \begin{subfigure}{0.48\textwidth}
    \includegraphics[width=\linewidth]{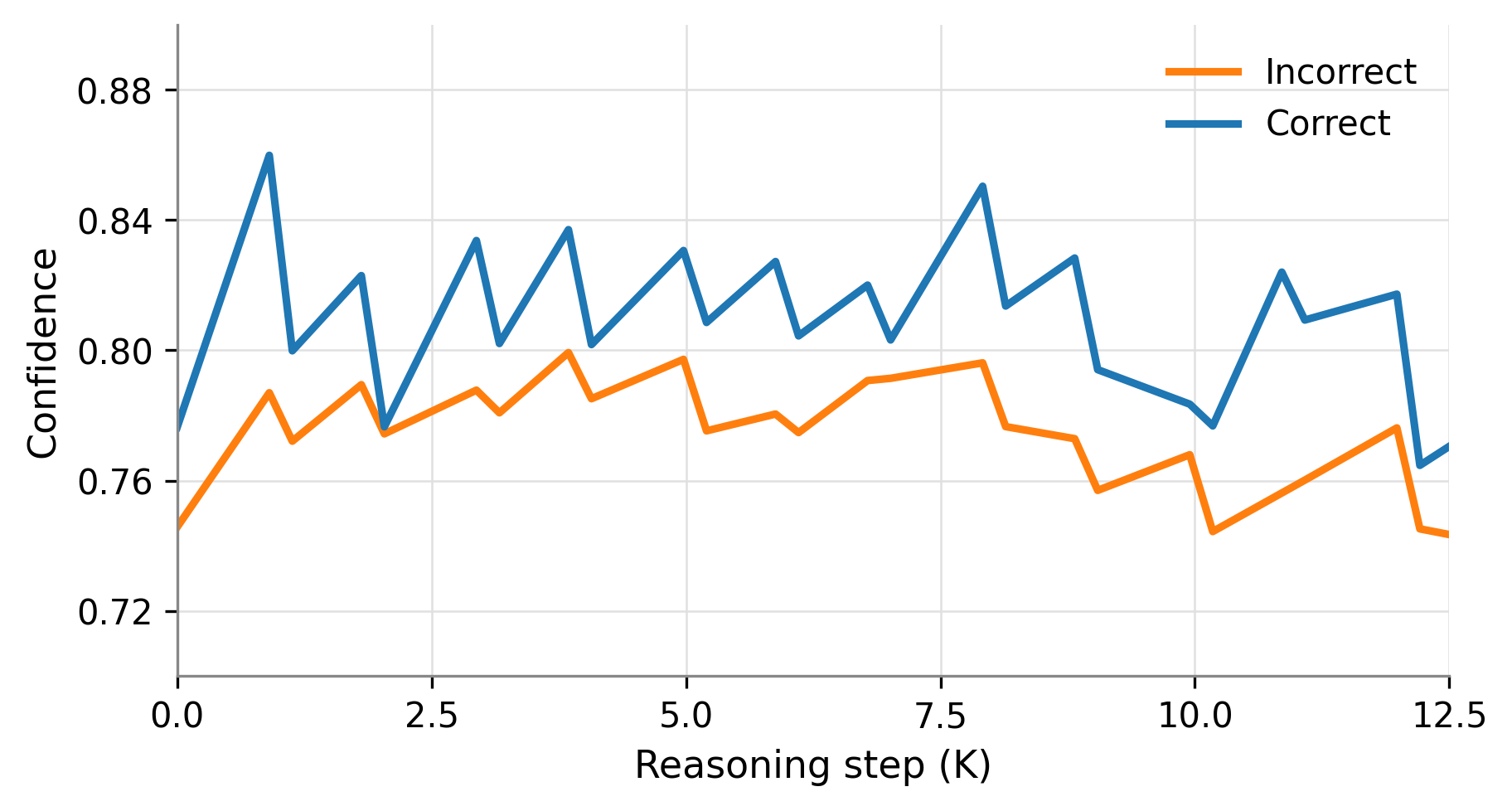}
    \caption{Nemotron -- GSM8K}
  \end{subfigure}
  \caption{Average confidence vs.\ reasoning step. Correct trajectories (blue)
  are high and well-separated early; the gap from incorrect (orange) shrinks at
  late steps, confirming early confidence is more discriminative.}
  \label{fig:app_fig4b}
\end{figure}

Figure~\ref{fig:app_fig4b} shows the average confidence as a function of the reasoning step. Correct trajectories (blue) rise quickly and remain high, so \emph{early-stage} confidence separates correct from incorrect far more sharply than \emph{late-stage} confidence, where the two curves converge, which is the key observation behind the early-weighted design.

\end{document}